# Agent-based models for the evolution of morphological alternation patterns

Aravinth Kulanthaivelu[1◑], Richard Sproat[1◑],

**1** Sakana.ai, Minato City, Tokyo, Japan

◑These authors contributed equally to this work.
* Corresponding author: aravinth@sakana.ai

## Abstract

Why is the past of English *go* the apparently unrelated *went*? Such alternations are frequent in languages. They neither aid communication nor learnability, yet they can be persistent, surviving over centuries or millennia.

We present a multi-agent simulation of the emergence of morphological stem and inflection alternations. Alternate forms arise by phonological changes or, as with *go*/*went*, from lexical alternatives associated with a subset of the population. When an agent 'hears' another agent use a novel form for a slot in the paradigm of a word (say, the past tense of *go*), they will with some probability adopt that form, possibly spreading its use to other slots in the paradigm that shared the same original form. Thus alternative forms can spread through the population and become entrenched as stem or inflectional marker alternants. Unlike many previous computational studies, our system allows for naturalistic lexical forms, realistic phonological rules, lexicons with hundreds or thousands of entries, and agent populations in the tens or hundreds. It supports several network topologies, diffusion patterns and agent adoption policies.

One issue with such simulations is evaluation: how realistic is the resulting morphology compared to those of real languages? We introduce the *AI Historical Linguist*, a novel Large Language Model-driven system that models a debate between two historical linguists. We use this to compare a set of real language morphologies, *disguised* morphologies, and experimentally evolved morphologies. The results suggest that among the factors that favor more plausible morphologies are scale-free social networks and random Bernoulli adoption of forms.

We also present three case studies modeling attested historical changes, allowing us to test what might have happened if history had been different.

All code and data are released.

# Introduction

Human language differs from communication systems of non-human animals in a number of ways. Perhaps the most salient differences relate to the size of the vocabulary, hence the variety of things that can be discussed; and the length and structure in the ways in which the items in the vocabulary can be combined (what linguists term *syntax*), and thus the complexity of the messages that can be conveyed. Two of the major questions surrounding human language are (1) how did it evolve in the first place and (2) having evolved, why did it change? It is the second question that is the topic of this paper and, within this question, more specifically how did complex morphological systems come about?

Simply put, morphology is "the problem of how a language can express in one word, what in another language must be expressed in multiple words" [1]. To take a simple example, the Japanese verb form *taberaremasen* is the negative, polite, potential or passive form of the verb *taberu* 'eat', whose English equivalent would be something like 'it cannot be eaten'. There is no straightforward way to express in one word in English what can be expressed with this single morphologically complex form in Japanese.

A common explanation for the existence of morphology is that it is a relic of older constructions that involved separate words where the boundaries between the words have eroded over time due to phonological changes. As Talmy Givón pithily put it, "today's morphology is yesterday's syntax" [2]. In addition, it has been argued [3] that morphological complexity trades off with information efficiency—in the information-theoretic sense of *information* [4, 5]—so that there is a benefit to having complex morphology, which might help explain why it evolved in many languages. While the authors of [3] are careful to stress that they do *not* make any claims about the common-language notion of *information*, nevertheless there is a sense in which the common-language notion is relevant: if nothing else, the Japanese word *taberaremasen*, communicates in one word what requires four in English. Indeed it actually communicates more than the English equivalent since it also conveys politeness to the addressee, which the English does not.

Counter to communicative efficiency is the concept of *least effort*, a concept commonly associated with the work of George Kingsley Zipf [6]. One consequence of least effort is the tendency for morphological systems to simplify over time, due primarily to phonological changes that erode distinctions. The history of English inflectional morphology has largely been a story of the gradual erosion of distinctions that were found in the ancestors of English but which, by the Middle Ages, had mostly been lost.

Still, there are some aspects of language that make little sense from either the point of view of communicative efficiency or least effort. One such case in morphology is the occurrence of morphological alternants that seem to serve no obvious function, but are nonetheless robust features of the languages that have them. Two examples of such phenomena are *stem alternations* and *inflectional paradigms*.

Paradigms will be familiar to anyone who has studied a Romance or Slavic language, and has had to memorize patterns of verb (or noun) endings. Such systems are introduced in language courses as *conjugations* (for verbs) and *declensions* (for nouns). Consider the case of two Spanish verbs *hablar* 'speak' and *comer* 'eat', whose inflection in the present indicative is as in Table 1. While the two verbs share the ending *-o* for the first person singular form, all the other endings differ between the two verbs. All of this is quite regular in that verbs that end in *-ar* invariably inflect the same way as *hablar* and those that end in *-er* invariably inflect the same way as *comer*. But what useful function does having more than one set of endings serve?

**Table 1.** Present indicative forms of the Spanish verbs *hablar* 'speak' and *comer* 'eat', exhibiting paradigm differences.

| | | | |
|---|---|---|---|
| Sg. | 1 | hablo | como |
| | 2 | hablas | comes |
| | 3 | habla | come |
| Pl. | 1 | hablamos | comemos |
| | 2 | habláis | coméis |
| | 3 | hablan | comen |

Stem alternations are less often explicitly discussed, but they will also be familiar to students of languages like Spanish. Consider Table 2, which gives the present forms of the verbs *contar* ‘count’, *sentar* ‘seat’ and *poder* ‘be able’. In addition to the paradigm differences between *-ar* and *-er* verbs already observed, these verbs also exhibit alternations in the stem, so that the singular forms and the third plural forms have one stem form, and the first and second person plural forms have another. In all cases, the alternation involves a change in the vowel from a simple (monophthong) vowel into a diphthong. The trigger for the alternation is stress: in the diphthong cases, the stress is on the stem—thus *cuénto*. In the first and second person plural, however, stress is on the affix—*contámos* and, as indicated in the orthography, *contáis*. The problem is that not all verbs undergo this alternation: observe that *comer* in Table 1 does not. So while one can predict the alternation pattern from the stress, one still has to learn whether or not the verb in question has the alternation. The stem alternations exhibited in Table 2 plausibly involve the same stem undergoing a phonological change: *pod-* and *pued-* are clearly derived from the same original root—Vulgar Latin *potēre*. But alternations need not involve the same stem at all, and cases of *suppletion* are common in languages. English provides one such example: *went* as the past tense of *go*. But for an example more parallel to the case of Spanish, one can consider the Italian verb *andare*, the present indicative of which is given in Table 3. Here we see two completely different stems, which nonetheless occupy the same positions as the two stem alternants in the Spanish case. Indeed, the trigger for the stem alternant in this case is also stress placement: in the *va-* forms (originally from the verb *vadere*) the stress is on the stem, and in the *and-* forms the stress is on the suffix. As we discuss in the context of the historical simulation presented in the “Romance verbal stem alternants” section (Supplementary Materials), stress-based stem alternations are common throughout Romance languages [7–11], though cases involving suppletion with more than one root are the minority of cases.

**Table 2.** Present indicative forms of the Spanish verbs *contar* ‘speak’, *sentar* ‘seat’ and *poder* ‘be able’, exhibiting both paradigm differences and stem alternations.

| | | | | |
|---|---|---|---|---|
| Sg. | 1 | cuento | siento | puedo |
| | 2 | cuentas | sientas | puedes |
| | 3 | cuenta | sienta | puede |
| Pl. | 1 | contamos | sentamos | podemos |
| | 2 | contáis | sentáis | podéis |
| | 3 | cuentan | sientan | pueden |

**Table 3.** Present indicative forms of the Italian verb *andare* ‘go’.

| | | |
|---|---|---|
| Sg. | 1 | vado |
| | 2 | vai |
| | 3 | va |
| Pl. | 1 | andiamo |
| | 2 | andate |
| | 3 | vanno |

Once again, why should languages exhibit such alternations? After all, they seem to serve no communicative function and serve merely to complicate the grammar for learners (both native and second-language learners). While such stem alternants can and do erode over time—see [9] for a discussion of the erosion of such cases in French—in some languages they seem to be remarkably robust. The *andare* alternation seen in Italian in Table 3 is found in a slightly different form in multiple Romance languages, suggesting that the invasion of *vadere* forms into the *andare* paradigm must have occurred close to 1500 years ago—certainly before Old French (roughly late 8th century onwards), which already shows the alternation.

The case of *andare* is typical of suppletion [12–16]. A phonologically induced alternation pattern, in this case caused by stress placement already existed in the language: if we designate the stressed stem “0” and the unstressed stem as “1”, the **stem alternation signature** for the present indicative is 000110. Then suppletive forms, here based on *vadere*, slotted in to the pattern, in this case occupying the “0” positions.

In the case of English *go*, the past tense of *wend* started replacing the original past tense, *yed*, some time during the early 15th century [17]. Indeed, both verbs *goon* and *wenden* were used more or less interchangeably with the meaning 'go' at least into the 14th century, as the following lines from the Prologue to Chaucer's *Canterbury Tales* shows:

> Thanne longen folk to *goon* on pilgrimages,
> …
> Redy to *wenden* on my pilgrymage

(Translation: then folk long to go on pilgrimages …ready to go on my pilgrimage.) In the case at hand, the old past tense form *yed*, from Old English *ēode*, was already suppletive. So the slotting in of *wend* forms would probably have seemed like substituting one arbitrary stem alternant with another.

Aronoff [13] introduced the rather odd term *morphome*, to describe alternation patterns where the alternants cannot be explained by any regular phonological change, nor do the forms in which particular alternants occur typically have any clear semantic or syntactic motivation. Rather they seem to be purely morphological in nature. Since Aronoff's term has become widely adopted in the field, we will also make use of it here. Other terms include the classical notion of *principal parts* and *rules of referral* [18]. In the latter case, the idea is that one can predict the form of a given entry by *referring* to another entry. For example, the stem of the second person plural present indicative in Spanish is the same as that of the first person plural.

In the linguistics literature on morphology, such alternations are commonly discussed assuming, at least implicitly, that there is a 'morphological component' that is part of an innate linguistic module in the brain, following the nativist theory of linguistics developed by Chomsky and others. Without getting into the merits of this broader theory, the existence of a specifically morphological module somewhere in the brains of speakers seems a dubious proposition. All languages differ in phonology, morphology and syntax, but of these three, morphology is by far the most variable. Some languages, like Vietnamese, have essentially no inflectional morphology and their derivational morphology is limited mostly to compounding. Others, of course, have exceedingly rich inflectional morphology. Languages also vary greatly in what information is encoded morphologically, and it is essentially impossible to name any kind of information that can be encoded in language that *cannot* be encoded morphologically or, contrariwise, *must* be encoded morphologically. Morphology also exhibits a wide variety of mechanisms for constructing complex morphological forms, including affixation, changes to the shape of the stem, vowel alternations, consonant alternations and tone changes. When linguists who subscribe to the nativist hypothesis propose a 'mental module' for one or another phenomenon, it is usually because they believe that this module provides useful constraints on what languages can and cannot do. If there is a morphological module, it does not seem that it does much work.

What humans *do* possess is ways of organizing knowledge. Presented with the output of random historical changes that result in seemingly pointless alternations such as the ones discussed above, speakers will try to make sense of them in one way or another. One possible outcome is that the alternations get eliminated over time. Another is that they become entrenched, but at the same time regularized so that, despite the complexity, there are some islands of certainty that speakers can make use of. If the first person indicative present stem for a verb is unpredictable, it helps if the speaker can be sure that at least the second person singular, and the third person singular and plural stems are the same. Morphomes, then are no more or less than a way of organizing arbitrary unprincipled facts that happen to have accreted over time.

In this paper we introduce a multi-agent simulation framework for modeling the adoption and spread of linguistic features, and we apply it in particular to the problem of stem and paradigm alternations. Unlike most previous work, our model allows for lexicons consisting of phonologically realistic forms, with hundreds or thousands of entries, and agent populations of several tens or even hundreds. Realistic phonological changes are modeled using the Pynini library [19, 20]. Network structure is implemented using the NetworkX library [21]. Agents 'communicate' messages involving these forms to each other, and with some probability adopt forms from other agents. We provide various models of 'compression' whereby agents can regularize their internal lexicons.

We demonstrate this system with various simulations involving synthetic languages, as well as case studies of real historical changes: see the “Case studies” section in the Supplementary Materials. One of the problems with synthetic simulations is evaluating how realistic these simulations actually are: we highlight this point in our discussion of some prior work in the next section. Ideally one could ask an expert historical linguist to rate the outcomes of the simulations for plausibility, but this is not a practical or scalable approach. To this end, we introduce the *AI Historical Linguist*, a Large Language Model-based system for assessing how plausible an evolved system is.

All of the software and needed data used in this system, along with the settings used in the reported experiments, are released at `https://github.com/SakanaAI/LanguageEvolution`.

## Prior Work

The initial emergence of language, and language evolution, also known as language change, has obvious parallels with the evolution of life, not the least of which being the gaps in evidence for important stages in the history. There is no direct evidence for how life arose on Earth, and our direct knowledge of the subsequent evolution of life is limited by what is available from the fossil record.

Similarly there is no direct evidence for how humans first developed their unique communication system. Even the date is uncertain, though [22] claim a date prior to 135Kya based on projections of the split of the Khoisan lineage. And similar to the case with biology, language change is at the mercy of the linguistic equivalent of the fossil record, namely written texts. The latter date to at most 5,000 years ago (in Mesopotamia), and are often imperfect representations of the language. Furthermore vast swathes of languages are lacking in written records, in many cases as recently as the modern era. To give one example, there is no written evidence for early Indo-European: the earliest Indo-European language to appear in written form was Hittite (17th to 13th century BCE), and its cuneiform-based writing system is a highly imperfect representation of Hittite speech. In particular, there is no direct evidence for how Proto-Indo-European evolved its complex inflectional morphology, though there is quite a bit of evidence for how that system later eroded in many branches of the language family.

Traditionally linguists avoided discussion of the origins of language. In 1866 the Société de linguistique de Paris, banned discussion of the topic (see [23]), due in part to lack of evidence, and in general historical linguists avoided the subject. In contrast, language change, for which there was sometimes direct evidence, and which in any case was amenable to study via the *comparative method* engendered a rich literature [24]. Within the last several decades there has been a number of quantitative studies of language changes and language phylogeny including [25–37].

Sociolinguists have also studied the sources of language change extensively. Influential work included that of the Milroys [38, 39], which argued that linguistic innovations are facilitated by loose-knit social networks, though subsequent work has highlighted the importance of centrally-connected members also [40]. Also broadly in this space are studies that derive properties of languages and language change from the structures of societies in which they are spoken [28, 41–43].

In addition to such empirical work, there has developed, over roughly the last three decades, a large literature on computational simulations of both language change and the emergence of language. The goal of this work is to try to fill in the gaps in our direct evidence by providing plausible mechanisms by which language *might* have emerged in the first place, and how it subsequently evolved the wide variety of language structures found in the roughly 7,000 languages in existence today. A useful repository of citations to relevant literature through 2018 can be found at the former site of the University of Illinois, Urbana-Champaign’s Language Evolution and Computation group at `https://langev.com/`. Good overviews of literature up through the early 2000’s can be found in [44, 45].

The literature is roughly equally divided between work that primarily focuses on emergence [3, 41, 46–122], and work that focuses primarily on language change [40, 123–190]. Among the latter there is often a division between models that characterize language change as primarily engendered by learners’ imperfect learning of the language transmitted to them from the previous generation; versus models that view change as primarily caused by innovations by agents who have already learned the language. See [191] for in-depth discussion.

In this work we assume the latter approach where groups of agents in a population introduce innovations, which either spread in the population or else die out. Particularly relevant to our current work are studies that relate to the topology of social networks (e.g. [192]) and modeling of the spread of innovations in a population (e.g. [188, 193, 194]). Our focus is on the evolution of morphology and morphophonological changes. Prior work in this area includes [27, 32, 33, 84, 85, 123, 149, 153, 165, 175, 187, 195–206].

One issue that comes up in language change that distinguishes it at least from models of natural selection in biological evolution is the issue of what motivates the spread of a change. While there may be cases where a change is adaptive, perhaps trading off complexity with efficiency [3], languages also contain artefacts that seem to have no adaptive significance, among these being the kinds of *morphomic* [13] alternations we discussed in the Introduction, and which will constitute the bulk of what we consider in this work. This brings up the question of whether language change can be *neutral* in that it does not confer any adaptive advantage. Prior work that discusses this topic includes [74, 75, 133, 142, 150, 164, 170].

In this paper we develop a multi-agent model that models the emergence of stem alternations and inflectional classes, and we also apply the model to simulations of historical changes in three different language groups. We largely assume that change is *neutral* as above, in that we do not directly model the adaptive advantage of agents' adoption of particular forms. Our system includes realistic phonological forms for morphemes, which may be either synthetic or based on actual reconstructed forms, as well as plausible phonological changes, and lexicons numbering in the hundreds or thousands of forms. The system can also run with tens or hundreds of agents. This distinguishes our work from a lot of prior contributions, which often use small numbers of agents, small lexica and abstract forms. For example prior work by [175, 187] on an attraction-repulsion model of the emergence of stable inflectional paradigms uses abstract forms to represent inflectional markers. While this may be fine as a way to verify the dynamics of the model, it makes it harder to evaluate the system by comparing to historically attested cases. For example, our own replication of the system in [187], using four exponents to fill eight slots for 100 lexemes produces paradigms that look as follows, where numbers in the slots denote distinct endings:

| | |
|---|---|
| Paradigm 1 | 2, 4, 2, 3, 2, 4, 3, 4 |
| Paradigm 2 | 3, 1, 4, 4, 4, 3, 4, 3 |
| Paradigm 3 | 4, 2, 1, 1, 1, 1, 1, 1 |
| Paradigm 4 | 1, 3, 3, 2, 3, 2, 2, 2 |

Such a set of paradigms falls directly out of [175, 187]'s attraction-repulsion mechanism whereby the dynamics favors picking different endings for the same slot in different paradigms. What may be hard to see from the numerical representation, though, is that the distribution of endings seems highly unnatural. This is somewhat easier to see if we replace the numbers with synthetic endings:

| | |
|---|---|
| Paradigm 1 | **-ok**, -ko, **-ok**, -em, **-ok**, -ko, -em, -ko |
| Paradigm 2 | -em, -as, -ko, -ko, -ko, -em, -ko, -em |
| Paradigm 3 | -ko, **-ok**, -as, -as, -as, -as, -as, -as |
| Paradigm 4 | -as, -em, -em, **-ok**, -em, **-ok**, **-ok**, **-ok** |

We highlight the distribution of one of these affixes, *-ok* (i.e. '2'), which shows up in the first, third and fifth slot of Paradigm 1, the second slot of Paradigm 3 and the fourth, sixth, seventh and eighth slots of Paradigm 4. This seems like a quite unnatural distribution, something that is more apparent if one uses reasonable, if synthetic, phonological forms. Note also that the model assumes an initial random assignment of exponents to slots resulting in 100 paradigms, i.e. one per lexeme in this case. This gradually reduces over time to four paradigms, as above. While the dynamics of the system make sense, it seems doubtful that any real instance of lexical paradigms evolved in such a fashion.

One of the components of our system is a novel Large Language Model-based evaluation system we call the *AI Historical Linguist*. We will show later in the paper how this can be used to evaluate the plausibility of the various synthetic systems produced by our model. For our synthesis of attested historical changes, we can appeal to how well the model replicates what is known about the changes in question.

## Paradigms and Stem Alternations

In this section we provide a formal characterization of paradigm classes and stem alternations in order to circumscribe the topic of discussion for most of the remainder of this paper.

We start by noting that the evolution of paradigm classes and stem alternations was the primary topic of the book-length study by Carstairs-McCarthy [198]. Carstairs-McCarthy's basic thesis is that stem alternations and paradigms arise because of *synonymy avoidance*, namely the tendency in natural language to avoid distinct expressions that have identical meanings. As an organizational principle in language, such a principle makes sense, and it has the consequence that when people acquire language, by default they assume that two terms that have distinct forms also have distinct meanings. To take a classic example, there may be relatively few people who are able to distinguish two common deciduous trees—say beech and elm trees—but English speakers nonetheless assume that *beech* and *elm* must denote different trees.

When two terms seem to have the same meaning, they invariably turn out to be used in complementary environments. Thus Carstairs-McCarthy gives the example of *addled* versus *rancid*, both of which have the sense of 'spoiled' when referring to food. But whereas *rancid* is typically used to refer to fats that have gone bad, e.g. *rancid butter*, *addled* is only used with *eggs*. Thus synonymy avoidance can itself be avoided by restricting apparently synonymous terms to different environments.

The relevance of synonymy avoidance to stem alternations and paradigmatic alternations is that on the face of it, having more than one stem form for a given lexeme, or several different ways of expressing *second person singular indicative present*, are instances where multiple forms express the same meaning. But as with the case of *addled*/*rancid*, morphological synonyms tend to get organized into complementary uses—paradigm classes which form equivalence classes across the vocabulary, or stem alternation patterns, so that the apparent synonyms tend not to be used in the identical environments. Thus for Carstairs-McCarthy, stem alternations and paradigms serve to separate synonymous forms into mutually exclusive bins. The alternation patterns (*signatures*) themselves serve as organizational categories to help the learner. If a lexeme has two stem alternants, for example, it helps if those alternants are always arranged so that the first variant always occupies the same slots as the first variant in other lexemes that also have two stem alternants. Thus the prevalence in Romance languages of what has been termed the N-pattern which we have seen examples of in Tables 2-3, and which we discuss in more detail in the "Romance verbal stem alternants" section (Supplementary Materials).

Formally we can define an inflected lexicon and paradigm, following the definition of [207], as follows.

> Let $\Lambda$ be a set of lexemes, let $\Sigma$ be a set of paradigm slots, where each slot corresponds to a bundle of morphosyntactic features, and let $\mathcal{F}$ be the set of possible surface forms over a fixed phonological or orthographic alphabet. An inflected lexicon is a set $\mathcal{L} \subseteq \Lambda \times \Sigma \times \mathcal{F}$, whose elements are word-types. A word-type is the triple $(\ell, s, f)$, where $\ell \in \Lambda$ is a lexeme, $s \in \Sigma$ is a slot, and $f \in \mathcal{F}$ is the corresponding surface form.

Surface forms are obtained by applying morphological changes to a lexeme. Such changes may include affixation of phonological material, stem modification, or combinations thereof. If the same morphological change is used for multiple slots within a paradigm, those slots are said to exhibit *syncretism*.

> For a lexeme $\ell \in \Lambda$, its paradigm is the set of word-types in $\mathcal{L}$ sharing that lexeme, $\pi(\ell) := (\ell, s, f) \in \mathcal{L}$. Equivalently, $\pi(\ell)$ may be regarded as the partial function from slots to surface forms induced by $\mathcal{L}$. In what follows, we assume that the relevant slot inventory is fixed within a lexical class. This induces a notion of equivalence on lexemes: two lexemes are paradigm-equivalent if their associated paradigms instantiate the same morphological realization of exponents over the same slot set. The corresponding equivalence classes may be interpreted as paradigm types.

Similarly a stem alternant pattern can be defined as follows:

A stem alternant pattern is defined relative to a lexeme $\ell$. Let $k$ be the number of slots in the paradigm of $\ell$, and let $S(\ell) = \{s_0, \ldots, s_{n-1}\}$ be the set of phonologically distinct stem alternants occurring across those slots listed in order of first appearance in the slots, with $n \leq k$. Fix an indexing of these alternants by integers $0, \ldots, n-1$. The stem alternant pattern of $\ell$ is then the vector $\boldsymbol{\sigma}(\ell) = (\sigma_1, \ldots, \sigma_k) \in \{0, \ldots, n-1\}^k$, where $\sigma_i = j$ if the form realising the $i$-th slot contains stem alternant $s_j$. Thus $\boldsymbol{\sigma}(\ell)$ records, for each slot, which stem alternant is selected.

The literature on stem alternants often distinguishes between stem alternants that are etymologically related to each other—for example the stems *pod-* and *pued-* in Spanish verb *poder* ‘be able’ (Table 2), and those that are clearly from different etymons, as with *go* and *went* in English, or the forms of *andare* in Italian (Table 3). The latter cases fall under the general category of *suppletion* whereby a totally unrelated form takes over parts of an alternation pattern. Suppletive forms are generally believed to be more restricted in their occurrence. Thus for example, [204], argues that etymologically unrelated stem alternants are far more restricted in the way they are distributed in noun (case and number) inflectional paradigms than are etymologically related forms so that, for example, non-suppletive alternations may occupy different case slots in a noun paradigm, but suppletion is restricted to number alternations. Thus it is common to find languages where the singular and plural stems are unrelated—cf. English *person*/*people*, but nouns cannot display one stem in the nominative, and a suppletive alternation in other cases, though non-suppletive alternations of this kind are allowed. Still, as the Italian *andare* example shows, comparable restrictions do not seem to apply to verbs, for reasons that are not entirely clear from McFadden’s theory that purports to explain the restriction in nouns.

## Model details

In this section we give an overview of the computational model we use to simulate language change. All agents are defined by their internal state - their lexeme-slot inventory - and their behaviour. We group these allowed behaviours into *adoption* and *compression*, dictating how agents incorporate heard utterances into their lexicon and simplify their lexicon respectively. Agents are behaviourally identical but typically differ in their internal state. The population is collectively represented by a connected graph, with each agent identified with one node only. Each agent communicates by sampling forms from their lexicon and transmitting them to their neighbors. For all agents the sampling is governed by a Zipf-Mandelbrot distribution, applied independently to slots and lexemes.

At the start of the simulation agents have lexicons which are largely identical but with some variance generated by a small set of differing forms, with agents possibly grouped into initial ‘dialects’. During the simulation these differences spread throughout the population. The model combines two distinct pressures: diffusion, which spreads forms through local interactions on the graph, and compression, which provides pressure toward simplified lexicons. Diffusion is governed by both the local agent adoption mechanism and the graph structure itself, whereas compression, our model of *analogical* processes of regularization, is controlled by the local agent compression method.

We now describe the mechanics of agent interaction per time-step. Within a communication round agents exchange utterances only with their graph neighbors. Each agent receives a queue of utterances from their neighbors. Going one-by-one through the queue, the receiving agent compares an incoming token with its current entry for the relevant lexeme-slot cell and may replace that entry according to the adoption rule. Fig 1 illustrates the default adoption pattern which we call the *Bernoulli* adoption method. With this adoption method the agent decides whether to adopt a form at random with a fixed pre-specified probability. Following adoption, all agents will (optionally) apply the compression method to their lexicons.

We leave most of the details of graph types, event mechanisms, and alternative adoption or compression policies to the Supplementary Materials. However the following details will be useful for understanding the experiments reported below:

- The default social structure graph is scale-free, but our implementation also supports *log-normal*, *Erdős–Rényi* and *complete* graphs.

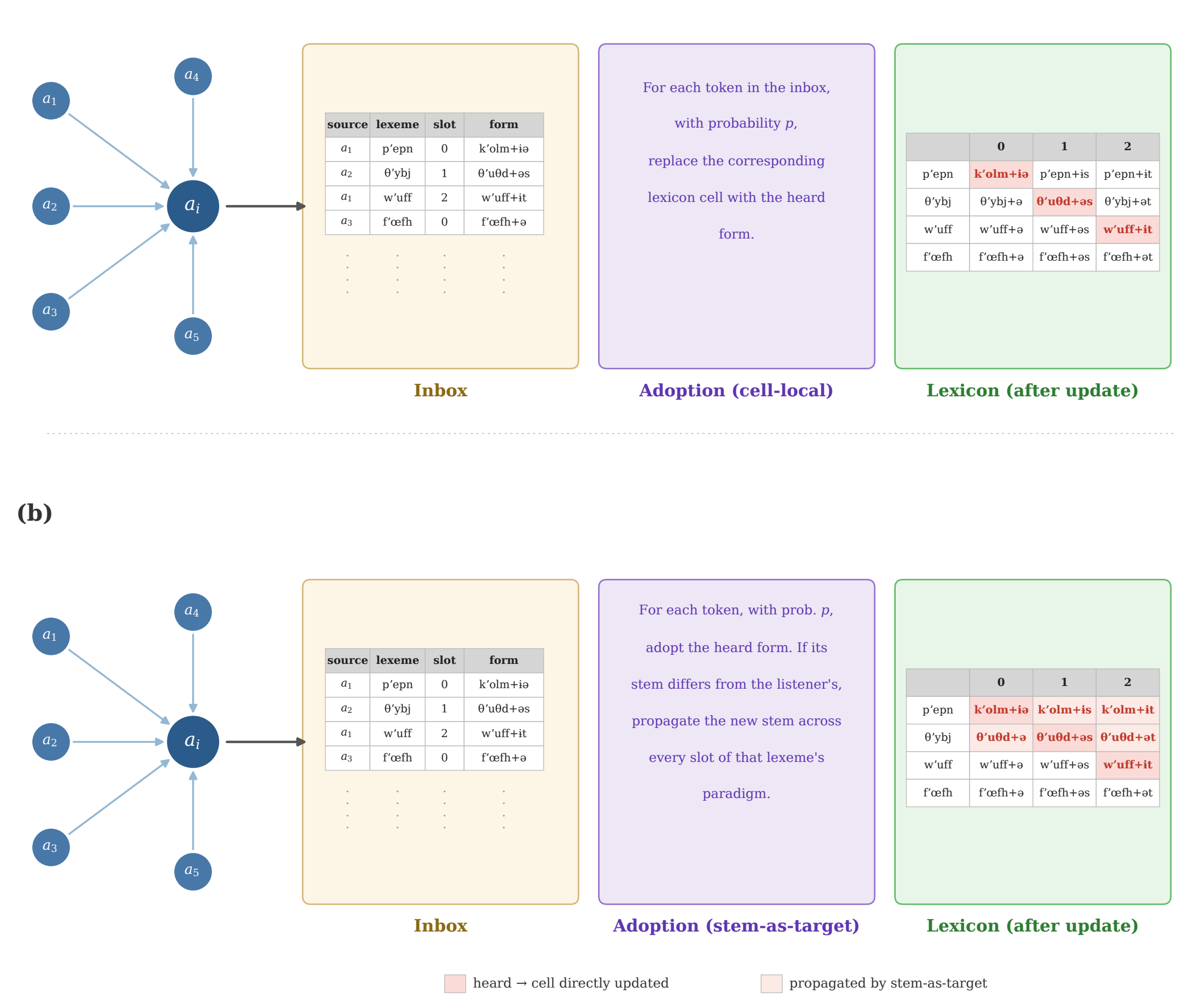

**Figure 1.** Schematic of within-step adoption. During one communication round, an agent hears forms from neighboring agents and may replace its current entry for the relevant lexeme-slot cell. In (a) the main adoption mechanism is Bernoulli adoption, optionally augmented with mild biasing terms such as novelty weighting, while (b) showcases the stem-as-target setting, which replaces not only the specific slot but all stems in remaining slots that have stem alternants that match the original stem of the specific slot.

- Initial sources of variation in the morphology are introduced by two mechanisms:
  - Phonological rules, applied to the forms for some subset of the population.
  - Alternative stems or affixes, in initial 'dialects', defined as subsets of agents.
- In addition to the Bernoulli adoption probability there is an additional control which relates to how an agent adopts a new form into its lexicon. Assume that agent $A_j$ communicates stem-affix form $s' + a'$ corresponding to a lexeme $l$ and slot $i$ to agent $A_k$, who originally had form $s + a$ for position $l_i$. Agent $A_k$ might simply decide to adopt $s' + a'$ for cell $l_i$.

  Alternatively, $A_k$ might conclude that $A_j$ is communicating a new form that should replace any instances where $A_k$ had matching input forms. Specifically, in the *stem-as-target* option, any slot for lexeme $l$ that has the stem alternant $s$ will replace $s$ with $s'$. In the non-mutually-exclusive *affix-as-target* option, any instance of an affix in the entire lexicon that has the form $a$ will replace that with $a'$. This is illustrated in Fig 1(b).

In the “Romance verbal stem alternant” section in the Supplementary Material, we will propose an additional constraint on the stem-as-target option that favors replacing stems that occupy a minority of slots in the paradigm.

- We implement the following compression methods:
  - *Class consensus* compression eliminates low-frequency stem alternant signatures by replacing them with higher frequency signatures with the same number of stem alternants.
  - *Global drift* is a gentler variant which compares each signature only to nearby signatures under a normalized Hamming distance. A stem is updated only if a nearby competitor is both sufficiently common and sufficiently dominant within that local neighborhood.
  - Finally, with *decision-tree* compression, a decision tree is first trained on the agent’s lexicon and is then used to predict forms, replacing lexical forms when the tree’s predictions differ from the current forms. This will have the effect of favoring more common patterns, at the expense of less common ones, thus regularizing the system over time. We use decision-tree compression in the implementations of Korean subject-marker allomorphy and Celtic lenition discussed in the Supplementary Materials.

  Further details of the various compression methods can be found in the Supplementary Materials.

**Linguistic interpretation of compression and adoption**
It is important at this point to explain what the compression techniques and what adoption controls like *stem-as-target* mean in linguistic terms. Compression, as we noted above, is a type of analogical regularization process. More generally, compression and adoption controls are ways of organizing, or reorganizing, lexical knowledge. First of all, *all of them assume that the agent implicitly uses the paradigm as an organizing principle that groups related forms of a single lexeme together*. *Stem-as-target* (and also *affix-as-target*), assume that the agent keeps track of the forms used within a paradigm (or in the case of *affix-as-target* slots with the same marking across paradigms), and substitutes all instances when encountering a new form. The compression techniques *class consensus* and *global drift* assume a global knowledge of stem-alternation patterns, and that the agent can reorganize forms according to statistically more robust patterns. *Decision-tree* compression involves rule learning, and thus assumes a mechanism whereby agents learn to generalize observed patterns. We take none of these assumptions to be particularly controversial.

Across all experiments we track the same core summary metrics measured at the end of the simulation.

- The first is normalized disagreement, which scales the number of clashing forms by the number of agent pairs and lexical slots, making it comparable across different population sizes.
- The next is complexity, which measures the number of distinct stem alternation signatures in the population at the end of the simulation.
- The conditional entropy of the stem alternation signature distribution measures how predictable the signature value in one slot is if you know the signature value in another slot. A low value corresponds to a highly predictable distribution of stem alternants whereas a high value means knowing the alternant pattern in one slot tells you little about another slot.
- Peak transfers measures the maximum number of forms adopted in any time step during the simulation. During a typical simulation one observes an early phase in which agents rapidly share and adopt forms, resulting in an early peak in transfers, followed by a longer tail to a steady state where fewer transfers occur.
- Finally, we measure the ‘shape’ of the final stem alternation signature distribution for a typical agent, defined to be the agent with lowest pairwise distance from all other agents. More precisely, we fit a one-parameter power law to the stem alternation signature distribution and report the measured exponent.

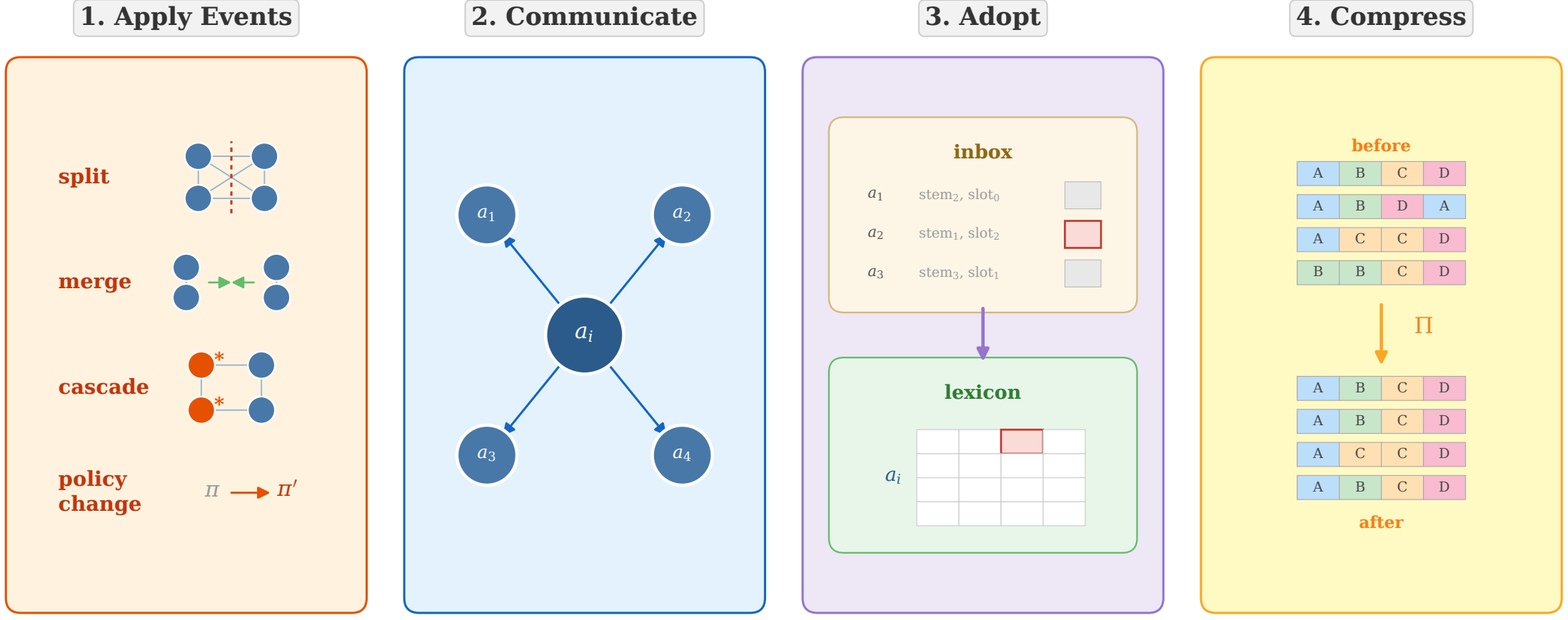

**Figure 2.** One simulation time step. Optional exogenous events are applied first, after which agents communicate on the graph and adopt heard forms. Compression is then applied synchronously, and the resulting state is recorded for later analysis.

Each time step proceeds as follows:

1. Optional exogenous events are applied. These may alter the graph or change the forms held by a subset of agents, allowing us to model processes such as population splits or targeted sound changes.
2. Every agent produces utterances sampled from its lexicon and sends them to its neighbors. (Unlike pairwise interaction models that update only a small number of speakers per tick, our scheduler forces every agent to participate in each time step.)
3. Agents adopt heard forms according to the chosen adoption policy.
4. All agents then apply compression synchronously.
5. Metrics are recorded.

A full simulation step is therefore organized into a sequence of event, communication, adoption, compression, and measurement phases. This is visualized in Fig 2.

# Research Questions

As noted above, our main focus in this paper is on the emergence of morphological stem alternations and paradigms in a multi-agent simulation. In particular we wish to answer several questions about how such morphological complexity arises. We outline these questions in the following sections.

## Lexical borrowing

The first question is whether a model that involves lexical borrowing between agents under the assumption of *neutral language change* (as discussed in the “Prior work” section) is sufficient to explain the genesis of stem-alternations and paradigms, and what the limitations are. Thus:

**RQ1.** Can stem-alternations and paradigms arise from lexical borrowing, and what assumptions do we need to make beyond neutrality in order for such alternations to arise?

### Network properties

A second question is what properties the network should have and what kinds of agent interactions most favor the language changes we are interested in. Thus:

**RQ2.** What properties of social networks and agent interaction best predict the emergence of stem-alternations and paradigms?

### Evaluation

A third issue that comes up in any model of language change is how one evaluates the system's behavior as compared to attested language change. This is relevant for both the dynamics of change, and the end result. In this paper we will focus on the end result, namely the (synthetic) morphologies that arise at the final stage of the simulation. In previous work that has simulated morphological change, there has been little attention to this detail, but it is particularly important in our view since one would like to know whether the resulting forms and their distributions look like what might have happened in real language change over hundreds of years of evolution. Ideally this would be something that a trained historical linguist might judge, but this is not a practical approach when one is running hundreds of possible simulations. In this paper we investigate using Large Language Models as proxy historical linguists in a system we dub the *AI Historical Linguist*:

**RQ3.** Can we make use of Large Language Models in the evaluation of evolved synthetic systems?

### Simulation of actual historical changes

Finally, we would like to use our system to simulate some attested historical changes, a topic we will investigate in the "Case studies" section which, for reasons of space, we defer to the Supplementary Materials:

**RQ4.** Can our system simulate actual historical changes?

## Materials and methods

### Experiments

The first empirical stage of this project is an intentionally broad but shallow set of experiments where we vary one factor of the model at a time against a fixed baseline. The goal is not to exhaust the full parameter space, but to establish the basic qualitative behavior of the model in order to identify which mechanisms have the largest effects on convergence and morphological diversity.

#### Shared Baseline

Unless otherwise stated, every experiment starts from the same language distribution, scale-free graph social network, population, and update regime. The baseline contains a population of 20 agents, each with a lexicon of 1000 lexemes and 6 inflectional paradigm slots. A shared pool of synthetic forms is constructed programmatically. More specifically, we use the Python package Pynini to construct 1000 stems from phonological templates, along with a fixed set of pre-defined affixes. To simulate dialects and introduce initial variation, we form 10 dialect groups to which each agent is randomly assigned, and substitute 50 stems from the shared pool with new and distinct stems. Finally, we apply the same phonological cascade to all agents, introducing further variation.

At each time step each agent generates 100 utterances, selected randomly from their inventory of forms, and transmits them to their neighbors. Each agent receives an ordered queue of forms from their set of neighbors and, with a fixed probability, decides whether to adopt them into their lexicon at the same corresponding lexeme and slot position. The simulation runs for 2000 time steps, with the *class consensus* compression policy applied every 10 steps, beginning at step 5. No exogenous events are applied, and there is no population turnover in the baseline configuration. We summarize these details in Table 4.

**Table 4.** Shared baseline used for the initial experiment sweep. Each experiment family modifies one aspect of this setup while leaving the remaining settings fixed.

| Component | Baseline setting |
|---|---|
| Language inventory | 1000 stems, 10 affixes, 10 dialects, shared lexeme-slot inventory |
| Initial phonology | Deaccent, IambicReversal, Fronting, Fortition, LatinVowelAssimilation, VowelDeletion, ReduceVowelAfterNonAccent |
| Population | $N = 20$, utterance length $= 100$, scale-free social graph |
| Adoption | Bernoulli, $p = 0.05$, novelty $= 0$, cell-local updating |
| Compression | class consensus, threshold $= 0.6$, interval $= 10$, start $= 5$ |
| Runtime | 2000 iterations |
| Events / turnover | none |
| Replication | 5 seeds per condition |

### Experiment Taxonomy

Each experiment we conduct takes the baseline configuration we previously defined and varies one factor, such as the probability of adoption. We group these into several dimensions of variation, each corresponding to one of the core mechanisms of the simulator. To summarize, these are:

- Adoption method; the rules which dictate whether an agent adopts a heard form.
- Bernoulli probability; the probability of adoption for the Bernoulli Adoption method.
- Birth and death; whether we allow for population turnover during the simulation.
- Compression method; the compression regimen which dictates how each agent simplifies its own lexicon.
- Compression threshold; the threshold parameter of the baseline compression policy.
- Form targeting; whether an agent replaces their entire lexeme with a new stem or the update is isolated to the heard slot.
- Novelty; a parameter which boosts the probability of an agent adopting a form if they have not heard the form in previous time steps.
- Population size.
- Topology; the structure of the social network which dictate agent interactions.
- Utterance length; the number of forms each agent transmits to their neighbors in a given time step.
- Affix inventory; whether agents begin with the fixed baseline affix inventory or a distinct randomized initial affix set.
- Dialect variation; the amount of form replacement used to seed variation across agents.

The experiments reported above form the main comparison set: in each case the population remains a single interacting system, and the manipulation changes the local conditions under which individual-level transmission proceeds. Signature-count summaries for these experiments are shown in Fig 3.

We also consider a second family of interventions which is less directly comparable to these sweeps. In these runs, the population may be split into non-interacting subpopulations, or a phonological cascade may be applied either to the whole population or to a designated subset of agents. Since these manipulations alter the global structure of the simulation, rather than only the parameters of interaction within a fixed population, we report them separately.

For the cascade experiments, the target subset is selected in one of three ways. In the random condition, the cascade is applied to a randomly sampled fraction of the population. In the degree-low and degree-high conditions, agents are ordered by connectivity degree and the cascade is applied to the specified fraction at one end of this ordering: the lowest-degree agents in the degree-low condition and the highest-degree agents in the degree-high condition. The cascade is introduced during the simulation, after which the system continues to evolve under the same transmission dynamics. The corresponding signature-count summaries are shown in Figs 4 and 5.

### Results

We summarize these results in Fig 6, which displays the measured impact for each factor that is varied, defined to be the average deviation from the corresponding baseline metric when varying along the denoted dimension.

As an immediate sense-check we note that the topology and population size have the greatest impact on the number of form transfers. Rather surprisingly we find that changing the adoption method and encouraging adoption of more novel forms does *not* seem to have a large effect on the simulation end-point. In fact, the Bernoulli adoption method appears to have the most pronounced impact on the final stem alternation signature distribution.

## Empirical Stem-Allomorphy Analysis

The model developed in this paper is intended to approximate the atomic person-to-person interactions that result in actual inflectional morphology. Given our focus on the stem alternation signature, a natural question is what the stem alternation signature distribution looks like in natural languages. In this section we explore this question empirically.

### Method

Using six datasets of inflected phonological forms with associated morphological annotation sourced from the Paralex Zenodo community, we perform a segmentation of naturally occurring forms and compute the induced stem alternation distributions. The languages we focus on are Italian, Latin, French, Hungarian, Finnish, and Estonian [208–213], covering a selection of Italic and Finno-Ugric languages. These languages do not necessarily have the same set of inflection slots and, due to data limitations, we do not always have information for the same class of lexical items. To make the calculation tractable and broadly comparable we focus on a fixed 12-cell paradigm for each language. These are described in Table 5. For Finnish and Estonian the datasets are provided with segmentation from the authors. We defer to these segmentations rather than our own methods.

**Table 5.** Each language we analyse along with the lexical class and selected slots. Every language is evaluated on a fixed 12-cell profile, and only complete paradigms are retained.

| Language | Lexeme class | Selected slots |
|---|---|---|
| Estonian | Nouns | nominative, genitive, partitive, illative, inessive, and elative in singular and plural |
| Finnish | Nouns | nominative, genitive, partitive, illative, inessive, and elative in singular and plural |
| French | Verbs | present indicative + present subjunctive, all persons and numbers |
| Hungarian | Nouns | nominative, accusative, dative, instrumental, translative, and inessive in singular and plural |
| Italian | Verbs | present indicative + present subjunctive, all persons and numbers |
| Latin | Verbs | present active indicative + present active subjunctive, all persons and numbers |

We use two segmentation methods: rule-based and LLM-based. In the former we perform segmentation by taking each phonological form and, given the cell, stripping the appropriate exponent for the form. For example, given the Italian form *parlo* and the knowledge that it is the first person present indicative of *parlare*, one knows the appropriate exponent for this cell is *-o*, and thus removes this substring from the form to get the stem *parl*. This inverts the standard procedure for generating inflected forms that learners are taught where one attaches an ending to a given stem.

This procedure is rather blunt and does not gracefully handle slight variation to an otherwise consistent underlying stem. Indeed, the rules of synchronic phonology typically applied to a concatenated stem and affix often result in the appearance of more perceived stems when performing naive exponent subtraction. This motivates the use of LLMs for segmentation. Rather than constructing all manner of rules that reflect synchronic phonology, we task an LLM, in this case Gemini-2.5-Flash, in determining the underlying stem given the language, form, and morphological information. Neither method is a perfect substitute for a detailed segmentation performed by a trained linguist, but they do serve as a *scalable* proxy.

## Results

We report coarse results of the segmentation in Table 6. Further results, in particular plots of the derived stem alternation signature distributions, can be found in the "Empirical stem allomorphy results" section in the Supplementary Materials. In all cases, one observes the recurring pattern of a small number of very common signatures followed by a longer tail of rarer types. To summarize these shapes more compactly we fit a one-parameter power-law regression to each signature distribution, $p(r) \propto r^{-\alpha}$ where $r$ is the rank of a given signature and $p(r)$ the frequency of that signature. We estimate $\alpha$ by the ordinary least squares method in log rank-frequency space. This value should be understood as a compact description of the slope rather than a strong distributional claim. Larger values of $\alpha$ correspond to steeper, more concentrated signature distributions. Table 6 reports the fitted shape parameters and regression diagnostics for each language and segmentation method.

**Table 6.** Power-law regression fits in log rank–frequency space for empirical signature distributions, by language and segmentation method. Estonian and Finnish use the segmentations provided by the dataset. Signature types counts distinct signatures in each distribution; larger $\alpha$ indicates a steeper, more concentrated rank-frequency curve.

| Language | Method | Signature types | $\alpha$ | Log-$R^2$ | Log-RMSE |
|---|---|---|---|---|---|
| Estonian | - | 56 | 2.098 | 0.922 | 0.526 |
| Finnish | - | 23 | 2.480 | 0.957 | 0.424 |
| French | Rule-based | 21 | 3.328 | 0.769 | 1.370 |
| French | LLM | 25 | 2.997 | 0.889 | 0.839 |
| Hungarian | Rule-based | 13 | 3.673 | 0.570 | 2.003 |
| Hungarian | LLM | 32 | 2.910 | 0.951 | 0.548 |
| Italian | Rule-based | 9 | 4.058 | 0.930 | 0.751 |
| Italian | LLM | 23 | 2.697 | 0.935 | 0.563 |
| Latin | Rule-based | 8 | 3.107 | 0.724 | 1.203 |
| Latin | LLM | 41 | 2.084 | 0.940 | 0.452 |

Broadly, across languages the estimated shape parameters mostly fall between 2 and 4 under these fits. Interestingly the shape parameter derived from the LLM-based method is consistently smaller than the rule-based value. Indeed, when focusing on the LLM-based segmentation the estimated shape parameters falls within the 2 to 3 range. It is at this stage ambiguous which method is more appropriate, and we leave such inquiries to further work.

The empirical distributions provide a useful point of comparison for the artificial languages generated by our simulator. Across the narrow section of languages considered here we find that stem alternation signatures are neither uniformly distributed nor collapse to a single dominant type. Instead we find a pattern in between these extremes where a small number of signatures account for much of the lexicon with a longer tail of signatures of much lower frequency. We should therefore expect a successful model to capture this same intermediate pattern of organisation. Further, the fitted rank-frequency exponents should provide a useful quantitative bridge for identifying more natural simulated outcomes.

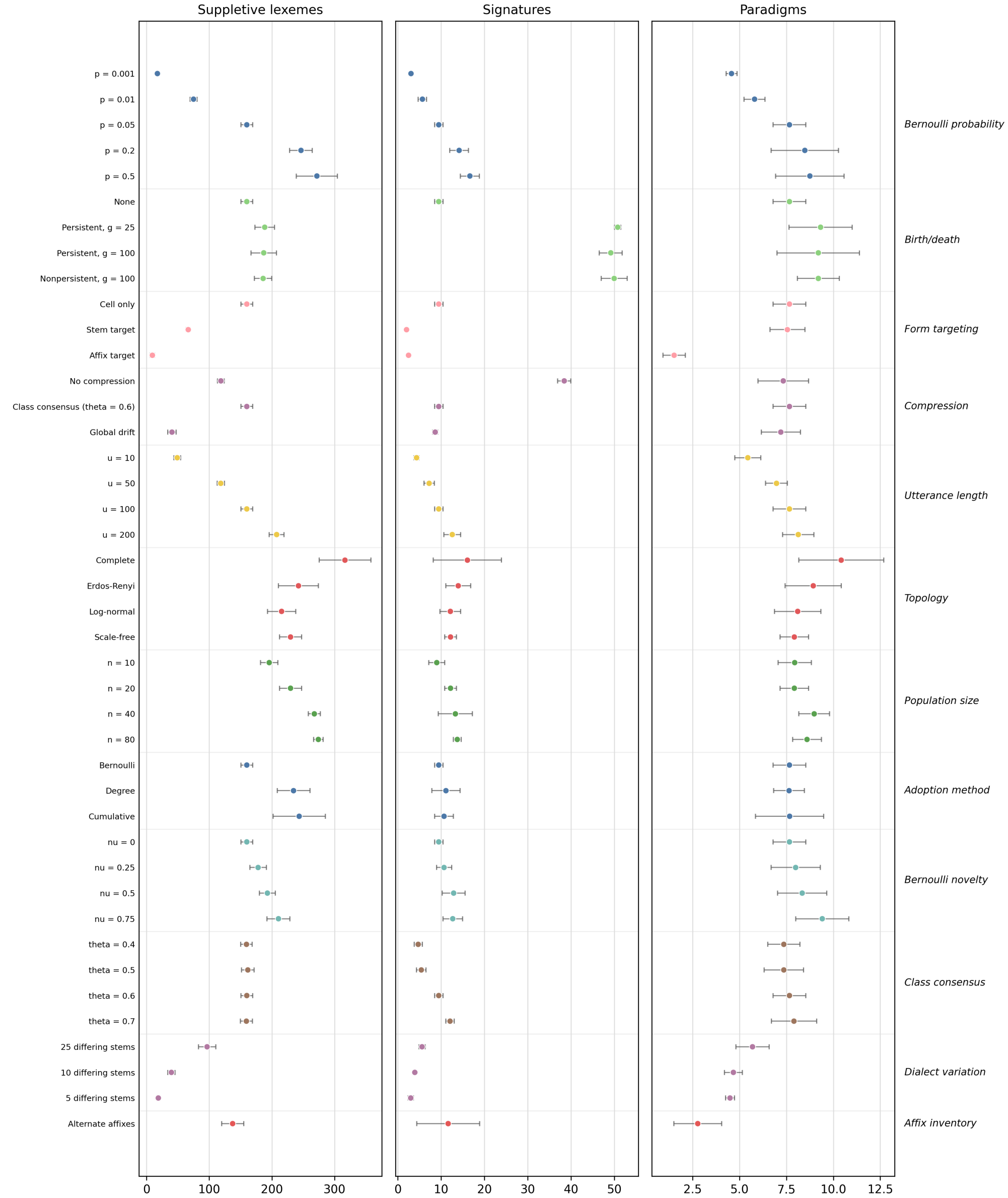

**Figure 3.** Rate of suppletion, number of stem alternation signatures, and number of paradigms for the distinct experiment configurations at the terminal state.

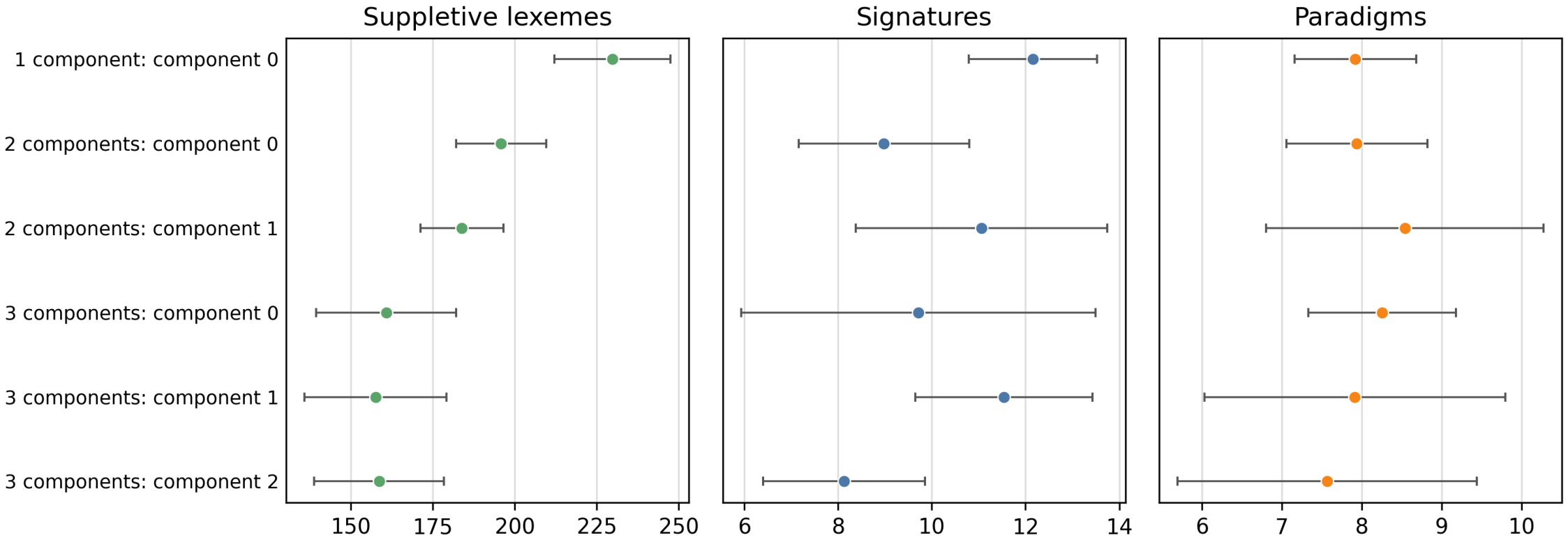

**Figure 4.** Signature-count summaries for the population-splitting intervention. The fixed population is divided into non-interacting components, changing the global interaction structure rather than only the local transmission parameters.

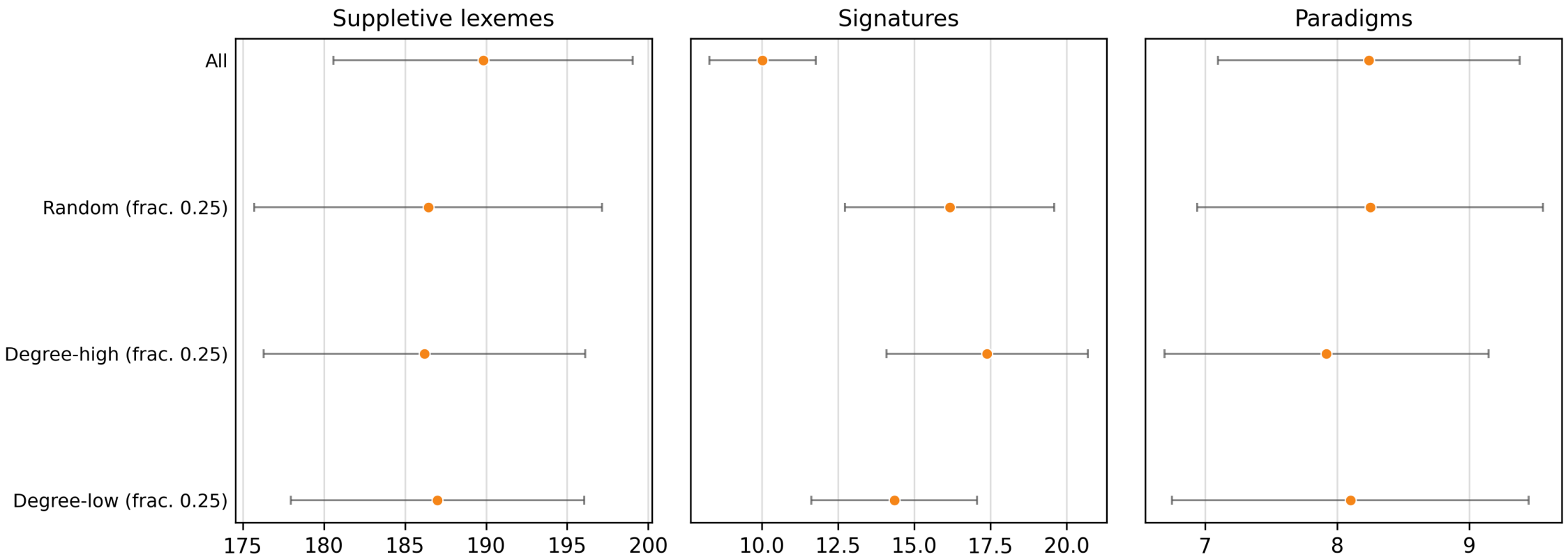

**Figure 5.** Signature-count summaries for the cascade-targeting intervention. Cascades are applied to the whole population or to subsets selected at random or by connectivity degree.

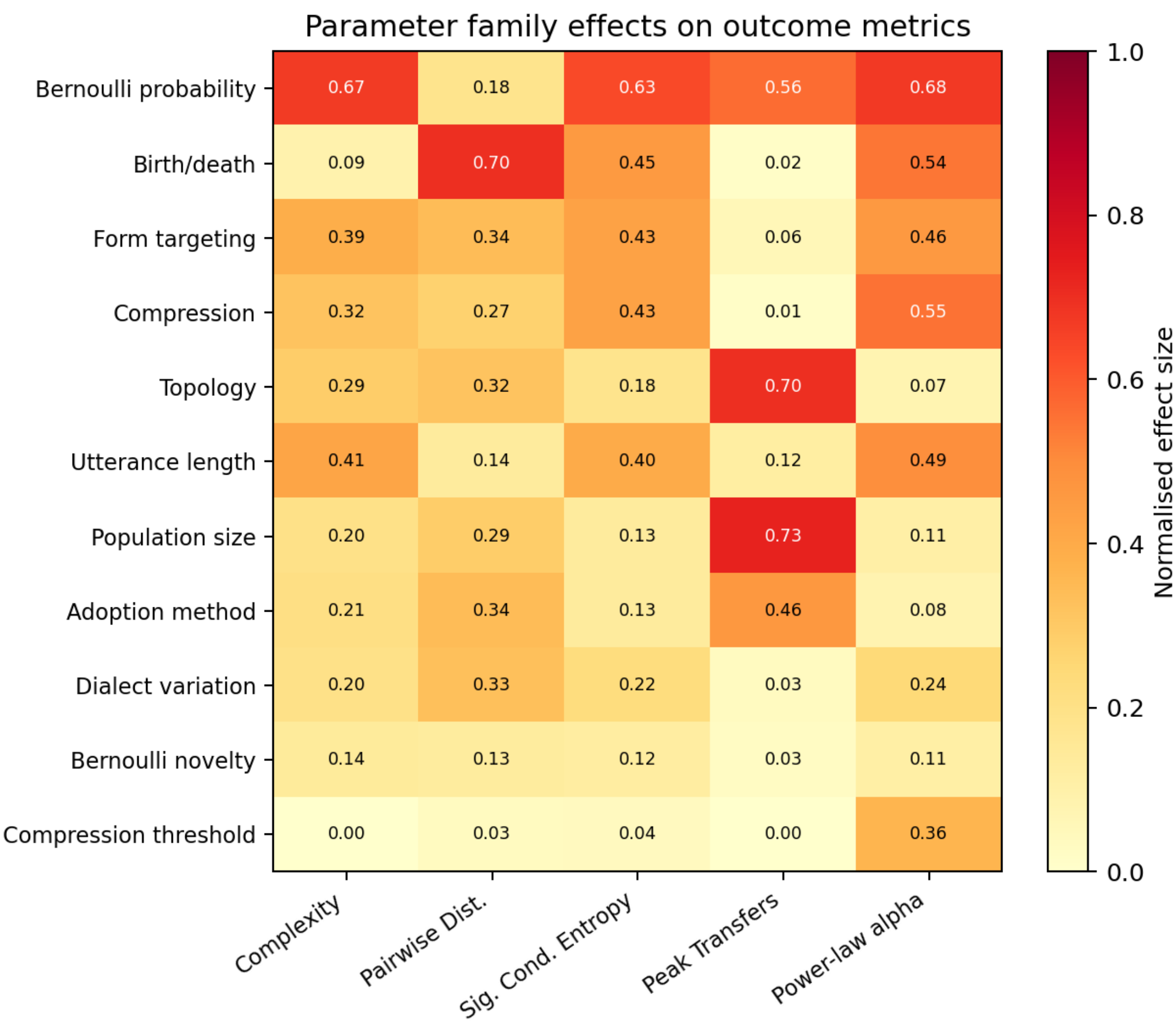

**Figure 6.** Heatmap displaying the mean normalized impact relative to the baseline of each factor for various summary statistics.

## Evaluation

### The AI Historical Linguist

One of the problems with research on computational modeling of language change is that it can be hard to evaluate the outcome. In such simulations, the plots of change may look impressive, but are they realistic? Are the final evolved systems plausible? Such problems also plague the field of artificial life, where it is often hard to know how to evaluate a system other than it "looks cool". If one is modeling an actual historical change, as in Hare and Elman's work on modeling changes in English verb morphology [123], or the three historical changes we present in the "Case studies" section in the Supplementary Materials, then at least one can see whether one's simulation is able to mimic what is known about the changes in question. But if one is running a simulation with artificial data, and one is trying to assess the end-state of the simulation, one has to judge whether the resulting system looks like a plausible instance of what one finds in natural languages. In the case of the present simulations, does the final state of the system, for example, look like a plausible morphological system, and why?

A panel of expert historical linguists with a wide knowledge of morphological patterns of the world's languages could certainly judge whether the system looks plausible: for example, the judgment might be that such-and-such a system is remarkably similar to a pattern found in Uralic languages. Or that the system looks rather implausible, because there is far too much suppletion. Needless to say such an approach would be expensive and hard to scale.

We therefore propose a novel automated method, which we call the *AI Historical Linguist*—AIHL, that engages Large Language Models in a multi-agent debate to assess the evolved morphological system, in order to come to an agreement on how plausible the system is, and to give reasons for the judgment. This work is loosely inspired by recent work such as the AI Scientist [214–216], which can sometimes help automate the discovery of new methods in AI.

The use of multi-agent debates has been investigated recently in a variety of tasks [217–223]. In such work, it has been shown that having multiple LLM agents 'discuss' a task can generally lead to improved performance over simply prompting a single agent to do the work.

The morphological system presented to the AIHL can be a real system—e.g. a portion of the verbal morphology for Latin, or a nominal paradigm for Estonian, or an artificial morphology evolved by our (or any) simulations. In the case of real morphological systems, there is of course the immediate problem that the LLM will have seen the data for that language before, and will recognize the system as being legitimate for that reason. In order to address that issue, we also introduce a system to disguise the phonology of the system so that, while the morphological patterns are the same, the words will not be recognizable. We discuss this in more detail below.

The data presented to the AIHL consists of a summary of the morphological system under consideration, including the number of lexemes represented in the sample, the total number of paradigms and sample inflected forms from each of the paradigms, along with details of the stem alternation patterns and the affixes. If there are many paradigms, as can happen in the case of Uralic languages, for example, a sample of paradigms is randomly selected.

The 'debate' involves two 'historical linguists', one being the *assessor* and the other the *critic*. The assessor is prompted with instructions and a sample of the target morphology. A fragment of a sample prompt with a simulated language is shown in Fig 7. Note in particular that the assessor is asked to consider among other things the number of paradigm classes; the patterns of stem alternation including incidence of suppletion; affix patterns including syncretism; the overall regularity of the system; and whether in general the patterns resemble those found in natural language. The assessor is then asked to rank the system on a 1–4 scale from very implausible to highly plausible for a naturally evolved system. There is also one "critical instruction" regarding possible morphological misanalysis, to which we return below.

In the case highlighted in Fig 7, the assessor (Claude-Sonnet-4-20250514 in this case) found, among other things:

- **Number and distribution of paradigm classes**: The system has 9 paradigm classes for 1000 lexemes, which is a reasonable number. However, the distribution appears uneven, with some

```
You are an expert historical linguist analyzing inflectional morphological paradigms from a language.

Your task is to assess whether the morphology presented below looks like a system that might have evolved naturally in a real human language.

Consider the following aspects in your analysis:
1. Number and distribution of paradigm classes
2. Patterns of stem alternation (suppletion, ablaut, etc.)
3. Affix syncretism (same affixes used across different slots)
4. Complexity and regularity of the system
5. Whether the patterns resemble attested natural language phenomena

CRITICAL INSTRUCTION - Morphological Misanalysis:
If you notice morphological segmentation errors in the data (e.g., "laudam+us" instead of "laud+amus"), you may comment on this misanalysis, but do NOT lower the plausibility score because of it. Such errors are
the fault of the annotator who created the data, not the morphology itself. Your plausibility rating should reflect whether the ACTUAL morphological patterns are plausible, not whether the annotations are accurate.

=== DATASET STATISTICS ===
Total lexemes in dataset: 1000
Total paradigms in dataset: 9

Showing all 9 paradigms


Here are sample lexemes from the language:


=== Paradigm 0 ===

Lexeme: b'itlev
Stem alternant signature: 000000
Forms: b'itlɨv+ə, b'itlɨv+əs, b'itlɨv+ət, b"itlɨv+'emʉs, b"itlɨv+'etɨs, b'itlɨv+ʉnt
Affixes: ə, əs, ət, 'emʉs, 'etɨs, ʉnt

…

=== Paradigm 1 ===

Lexeme: g'eðha
Stem alternant signature: 000000
Forms: g'eðh+ə, g'eðh+əs, g'eðh+ət, g"eðh+'amʉs, g"eðh+'atɨs, g'eðh+ənt
Affixes: ə, əs, ət, 'amʉs, 'atɨs, ənt
Stem alternants: g'eðh, g'eðh, g'eðh, g"eðh, g"eðh, g'eðh

…

…

Provide your analysis following this format:

⟨REASONS⟩
[Your detailed reasoning about why this system does or does not look like a naturally evolved morphological system. Be specific and cite examples from the data.] ⟨/REASONS⟩

⟨SCORE⟩
[A single number from 1 to 4, where:
1 = very implausible as a naturally evolved system
2 = somewhat implausible
3 = somewhat plausible
4 = very plausible as a naturally evolved system]
⟨/SCORE⟩
```

**Figure 7.** Sample prompt for the AI Historical Linguist with an artificially evolved language.

paradigms having only one example shown.

- **Patterns of stem alternation**: The stem alternations present serious issues:
  - There's a consistent pattern of accent shift (single to double accent marks: ' to ") in positions 4-5 across most paradigms, which is phonologically plausible
  - However, the extreme suppletion is highly problematic. In Paradigm 3, we see *l'omvɨŋ* alternating with *n'yððɨjj* – these stems share no phonological material whatsoever
  - Similarly, in Paradigm 6, *j'obfɨð* alternates with *f'ejg*, and in Paradigm 7, *p'opkʉn* alternates with *ŋ'uθk*
  - Such radical suppletion affecting multiple stem positions within single paradigms is extremely rare cross-linguistically

The assessor assigns this system an overall score of 2, i.e. "somewhat implausible."

The critic is then given the same prompt as the assessor, along with the assessor's assessment of the system, and asked to critique the assessment. In the case at hand, the critic (Gemini-2.5-Pro) proposed refining assessor's characterization of the suppletion, noting:

> The assessor's argument could have been made even stronger by analyzing not just the **presence** of suppletion, but its completely chaotic **distribution** within the paradigms.
>
> - In Paradigm 2, the suppletive stem appears in the 3sg and 1pl.

- In Paradigm 6, it's in the 1sg and 2sg.
- In Paradigm 7, it's in every form **except** the 1sg.
- In Paradigm 8, it appears **only** in the 1pl.

> This lack of a coherent pattern across the suppletive classes is the biggest red flag. Historical processes that lead to suppletion (like the merger of defective verbs) rarely result in such a diverse and arbitrary set of syncretic patterns. This distribution looks computationally generated rather than historically evolved.

The critic in this case ended up agreeing with the assessor's score of 2.

Given that the two scores are the same (or close) to one another, the iteration stops, with an overall score of 2, but if the scores were further apart, the assessor would be asked to consider revising their assessment, and the process would iterate, up to some predetermined number of steps (8 in our experiments). Usually, the assessor and critic come to an agreement within no more than 3 steps.

Clearly the point of assessing evolved artificial languages is to see how they compare in terms of plausibility with actual languages, and so it is necessary to also run the AIHL on real systems. To this end, we considered verb paradigms in Italic languages and noun paradigms in Uralic languages. Specifically, we considered the present active indicative and subjunctive in Latin, and the present indicative and subjunctive in French and Italian; and the nominal paradigms of Finnish, Estonian and Hungarian. As noted previously, all data were from Paralex [208–213].

However these data present two problems. First, apart from Finnish and Estonian, the Paralex data does not provide a uniform decomposition of morphologically complex forms into stems and affixes, and while we were able to compute the correct decomposition in most cases, there were residual errors leading to an inflated number of paradigms and stem alternation patterns in several cases. Not surprisingly, the AIHL noticed these and often rated the systems lower in plausibility. The solution to this was to add the caveat noted above about morphological misanalysis: the system could note the existence of a morphological misanalysis, but insofar as this was the fault of the 'linguist' rather than the language itself, the language should not be penalized for this.

A more serious concern is that, also not surprisingly, the LLMs frequently identified the language in question, presumably because they had seen Paralex or similar data. (Note that the data are in phonetic transcription, so the identification of the language is not a trivial match to standard orthography.) To address this issue, we developed a system to phonologically disguise the input, so that the language could not be (trivially) identified. This system performs such operations as rotating the place of articulation, e.g. labials becoming velars, and changing vowels. Crucially, these changes are deterministic and one-to-one, since it is important not to eliminate any distinctions found in the source language, nor add distinctions not found there. The code can be found in `https://github.com/SakanaAI/LanguageEvolution/blob/main/NewAIHistoricalLinguist/tools/disguise_2.py`. This is similar in spirit to work reported in [224], who perform careful disguising of language data for Linguistic Olympiad problems.

The disguised languages tend to be rated lower than the undisguised languages which, as noted, the systems generally recognize. Part of the reason for this is that peculiar features of actual languages are cited as problems more readily if the language cannot be readily recognized. For example, Estonian has an unusual three-way length distinction in consonants. As long as the AIHL recognizes that the language is Estonian, this poses no problem, but when the language is disguised, this feature is marked as unusual. On the flip side, however, the AIHL is often able to make plausible decisions noting, for example, that disguised Finnish or Estonian are reminiscent of Uralic languages, or that disguised Italian is reminiscent of Romance languages.

When using the AIHL with synthetically evolved languages, since the agents in the simulations may not have converged on an identical grammar, we pick an agent that is close to the median of the distribution in terms of string edit distance from the other agents and is therefore most representative of the morphological patterns evolved by the system. Across ten runs of the system with Gemini-2.5-Pro as assessor and Claude-Sonnet-4-20250514 as the critic (Gemini/Claude), all real languages as well as disguised Italian had a rating of 3.0 or above for all runs. The other disguised languages had a rating of 2.0 or above. Three of the synthetic evolved languages had a rating of 2.0 or above and the others had ratings lower than 2.0. See Fig 8, top panel. With the critic-assessor roles reversed (Claude/Gemini), all real languages had ratings of 3.0 or above, except for Latin which was 2.0 or above. For the disguised languages, Finnish, French and Italian had ratings of 3.0 and above, Hungarian and Latin were 2.0 or above, and disguised Estonian fared the worst. In this case none of the artificial languages were consistently above 2.0. See Fig 8, bottom panel. For the disguised languages, problems noted included the seemingly too complex phonology, such as the three-way length distinction in the case of disguised Estonian, and apparent over complexity of the morphology—issues that are not problematic when the language is known to the LLM. For the synthetic languages one of the recurring themes is the sheer amount of suppletion, whose root cause is the number of alternate forms for lexemes assumed for a subset of the initial population, as well as the likelihood of those forms being borrowed by other agents into parts of their paradigms. While the results are mixed, it is broadly the case that real languages (for obvious reasons) are rated highly, disguised languages generally less so. The synthetic evolved languages that are the topic of this work are generally considered less natural still, though there is variation, with some of them reaching a mean level of 'somewhat plausible'. Indeed, mean values for some of the synthetic languages rate on a par with some of the disguised languages, in line with the fact that the undisguised real languages automatically benefit, most of the time, by being languages that the systems already recognize. We explore the variation in assessments of the synthetic languages in more depth in the next subsection.

We note in passing that the cost to evaluate one language is approximately US 15 cents for Claude, and probably around 10 cents for Gemini, based on the published cost differences between Claude Sonnet 4 and Gemini 2.5 Pro, so approximately 25 cents in total for a complete evaluation by assessor and critic for one language.

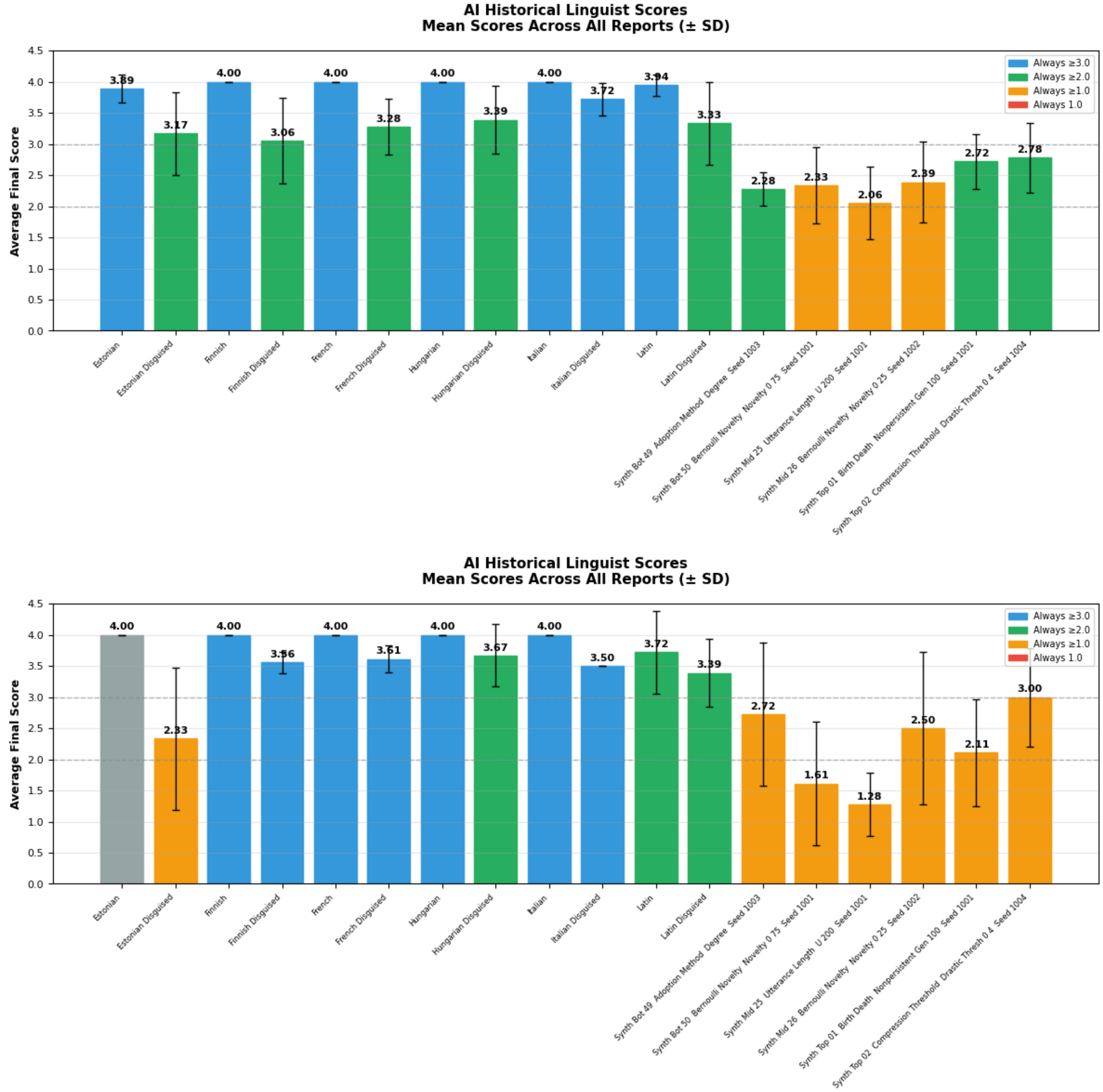

**Figure 8. Top panel**: Bar plot showing the mean scores assigned by the assessor and critic over ten runs of the AI Historical Linguist. Here the assessor is Gemini-2.5-Pro and the critic is Claude-Sonnet-4-20250514. The morphologies evaluated are for Estonian, Finnish, French, Hungarian, Italian and Latin in both real and disguised versions; and six sample artificial languages from the set evaluated more fully in the section "Evaluating generated languages with the AIHL". Blue bars are for those languages where the mean assessed score was 3.0 or above *for all of 5 iterations*, green bars are for cases that are 2.0 or above for all 5 iterations, orange bars are those that are 1.0 or above in all iterations. In this case, the problematic 'real' languages are the disguised versions of Estonian, Finnish, French, Hungarian and Latin. Four of the artificial languages also show scores consistently greater than or equal to 2.0.

**Bottom panel**: showing the results with the assessor-critic roles reversed. The Estonian bar in this plot is gray since one evaluation failed for this language.

### Evaluating generated languages with the AIHL

Continuing the exploratory work as described in the Experiments section we applied the AIHL to the resulting artificial morphological systems. Across the 10 families of experiments there are 33 distinct experimental 'conditions' in total, each simulated under 5 different seeds. For each of these conditions we sampled from the results by independently drawing 3 seeds at random per condition. For each seed we then extract the lexicon of the centroid agent. The centroid agent is defined to be the agent whose lexicon has the lowest Levenshtein distance to all other agents, and thus represents the most 'average' speaker. All evaluations used Gemini-2.5-Flash as the assessor and Claude-Sonnet-4-20250514 as the critic, with 5 lexemes sampled per paradigm class and at most 20 paradigm classes presented to the dialogue.

The mean plausibility scores for each experiment are reported in Table 7, ranked from most to least plausible. Table 8 shows the same results grouped by change family. See also Figs 10 and 11.

**Table 7.** All experimental conditions ranked by mean AIHL plausibility score (1–4 scale), averaged over three independently drawn simulation seeds. SD is the standard deviation across the three runs. Horizontal rules separate three broad bands: conditions at or above 3.0 (top), conditions between 2.5 and 3.0 (middle), and conditions below 2.5 (bottom). Baseline condition is in boldface.

| Rank | Family | Condition | Mean | SD |
|---|---|---|---|---|
| 1 | Utterance length | $u = 10$ | 3.50 | 0.00 |
| 2 | Bernoulli probability | $p = 0.01$ | 3.17 | 0.85 |
| 3 | Compression | Global drift ($r = 0.2$, $\theta = 0.7$) | 3.17 | 0.47 |
| 4 | Form targeting | Affix target | 3.17 | 0.47 |
| 5 | Population size | $n = 10$ | 3.17 | 0.47 |
| 6 | Bernoulli probability | $p = 0.2$ | 3.00 | 0.41 |
| 7 | Birth/death | Nonpersistent ($g = 100$) | 3.00 | 0.71 |
| 8 | Compression threshold | Class consensus ($\theta = 0.4$) | 3.00 | 0.71 |
| 9 | Dialect variation | 10 differing stems | 3.00 | 0.41 |
| 10 | Utterance length | $u = 50$ | 3.00 | 0.71 |
| 11 | Affix inventory | Alternate 6 affixes | 2.83 | 0.47 |
| 12 | Bernoulli probability | $p = 0.001$ | 2.83 | 0.47 |
| 13 | Compression threshold | Drastic ($\theta = 0.5$) | 2.83 | 0.47 |
| 14 | Dialect variation | 25 differing stems | 2.83 | 0.47 |
| 15 | Form targeting | Stem target | 2.83 | 0.94 |
| 16 | Bernoulli novelty | $\nu = 0.25$ | 2.67 | 0.24 |
| 17 | Bernoulli novelty | $\nu = 0.5$ | 2.67 | 0.62 |
| 18 | Compression threshold | Class consensus ($\theta = 0.7$) | 2.67 | 0.62 |
| 19 | Dialect variation | 5 differing stems | 2.67 | 0.62 |
| 20 | Population size | $n = 40$ | 2.67 | 0.62 |
| 21 | Population size | $n = 80$ | 2.67 | 0.62 |
| **22** | **Baseline** | **Default simulator** | **2.62** | **0.61** |
| 23 | Bernoulli probability | $p = 0.5$ | 2.50 | 1.41 |
| 24 | Birth/death | Persistent ($g = 100$) | 2.50 | 0.82 |
| 25 | Compression | None | 2.50 | 0.82 |
| 26 | Utterance length | $u = 200$ | 2.50 | 0.82 |
| 27 | Topology | Complete | 2.33 | 1.03 |
| 28 | Topology | Erdős–Rényi ($p \approx 0.21$) | 2.33 | 0.85 |
| 29 | Topology | Log-normal ($\mu = 1.2$, $\sigma = 0.35$) | 2.33 | 0.24 |
| 30 | Adoption method | Cumulative | 2.17 | 0.24 |
| 31 | Birth/death | Persistent ($g = 25$) | 2.17 | 0.94 |
| 32 | Adoption method | Degree | 2.00 | 0.41 |
| 33 | Bernoulli novelty | $\nu = 0.75$ | 1.67 | 0.47 |

**Table 8.** All experimental conditions, sorted by family, and within each family by mean AIHL plausibility score (1–4 scale), averaged over three independently drawn simulation seeds. Other details as in Table 7, except that here the bolded baseline is repeated for each family with the settings for that family.

| Rank | Family | Condition | Mean | SD |
|---|---|---|---|---|
| 1 | Utterance length | $u = 10$ | 3.50 | 0.00 |
| 10 | Utterance length | $u = 50$ | 3.00 | 0.71 |
| **22** | **Utterance length** | $u = 100$ | **2.62** | **0.61** |
| 26 | Utterance length | $u = 200$ | 2.50 | 0.82 |
| 2 | Bernoulli probability | $p = 0.01$ | 3.17 | 0.85 |
| 6 | Bernoulli probability | $p = 0.2$ | 3.00 | 0.41 |
| 12 | Bernoulli probability | $p = 0.001$ | 2.83 | 0.47 |
| **22** | **Bernoulli probability** | $p = 0.05$ | **2.62** | **0.61** |
| 23 | Bernoulli probability | $p = 0.5$ | 2.50 | 1.41 |
| 16 | Bernoulli novelty | $\nu = 0.25$ | 2.67 | 0.24 |
| 17 | Bernoulli novelty | $\nu = 0.5$ | 2.67 | 0.62 |
| **22** | **Bernoulli novelty** | $\nu = 0.0$ | **2.62** | **0.61** |
| 33 | Bernoulli novelty | $\nu = 0.75$ | 1.67 | 0.47 |
| 3 | Compression | Global drift ($r = 0.2$, $\theta = 0.7$) | 3.17 | 0.47 |
| 8 | Compression threshold | Class consensus ($\theta = 0.4$) | 3.00 | 0.71 |
| 13 | Compression threshold | Class consensus ($\theta = 0.5$) | 2.83 | 0.47 |
| 18 | Compression threshold | Class consensus ($\theta = 0.7$) | 2.67 | 0.62 |
| **22** | **Compression threshold** | **Class consensus (**$\theta = 0.6$**)** | **2.62** | **0.61** |
| 25 | Compression | None | 2.50 | 0.82 |
| 4 | Form targeting | Affix target | 3.17 | 0.47 |
| 15 | Form targeting | Stem target | 2.83 | 0.94 |
| **22** | **Form targeting** | **None** | **2.62** | **0.61** |
| 5 | Population size | $n = 10$ | 3.17 | 0.47 |
| 20 | Population size | $n = 40$ | 2.67 | 0.62 |
| **22** | **Population size** | $n = 20$ | **2.62** | **0.61** |
| 26 | Population size | $n = 80$ | 2.67 | 0.62 |
| 7 | Birth/death | Nonpersistent ($g = 100$) | 3.00 | 0.71 |
| **17** | **Birth/death** | **None** | **2.62** | **0.61** |
| 24 | Birth/death | Persistent ($g = 100$) | 2.50 | 0.82 |
| 31 | Birth/death | Persistent ($g = 25$) | 2.17 | 0.94 |
| **17** | **Adoption method** | **Bernoulli** | **2.62** | **0.61** |
| 30 | Adoption method | Cumulative | 2.17 | 0.24 |
| 32 | Adoption method | Degree | 2.00 | 0.41 |
| **22** | **Topology** | **Scale-free** | **2.62** | **0.61** |
| 27 | Topology | Complete | 2.33 | 1.03 |
| 28 | Topology | Erdős–Rényi ($p \approx 0.21$) | 2.33 | 0.85 |
| 29 | Topology | Log-normal ($\mu = 1.2$, $\sigma = 0.35$) | 2.33 | 0.24 |
| 11 | Affix inventory | Random 6 affixes | 2.83 | 0.47 |
| **22** | **Affixes** | **Latin-based affixes** | **2.62** | **0.61** |
| 9 | Dialect variation | 10 differing stems | 3.00 | 0.41 |
| 14 | Dialect variation | 25 differing stems | 2.83 | 0.47 |
| **22** | **Dialect variation** | **50 differing stems** | **2.62** | **0.61** |
| 19 | Dialect variation | 5 differing stems | 2.67 | 0.62 |

Across all experiments the scores ranged from 1.67 to 3.50, with the baseline experiment obtaining a mean rating of 2.62. All runs show relatively high variance, except for the experiments with a population size of 80 and utterance length of 10. We speculate that these examples represent rather extreme cases relative to the baseline, and may not result in a well-mixed population. Many of the parameters in the simulator are intrinsically coupled, having a larger population means it takes much longer to reach equilibrium, and fewer utterances transmitted mean agents do not adopt new forms so readily. Thus one should take those results with a grain of salt.

While the results are noisy, there are apparent trends, and one can make the following tentative observations on the factors that lead to more plausible morphologies:

- scale-free graph social networks,
- the random Bernoulli form adoption,
- a relatively *low* probability of adoption,
- a lower bias towards adopting novel forms.

To connect the AIHL judgements with the empirical signature-distribution analysis, we also compare each AIHL-scored synthetic condition with the fitted rank-frequency slope of its terminal stem-alternation signature distribution in Fig 9. Recall that we fit distributions of the form $p(r) \propto r^{-\alpha}$, where $r$ is the rank of a signature and $p(r)$ is its relative frequency. Larger values of $\alpha$ therefore correspond to steeper, more concentrated distributions: a small number of signatures account for more of the lexicon, while the tail of rare signatures is correspondingly lighter. The positive descriptive trend in Fig 9 suggests that the AIHL tends to assign higher plausibility scores to systems whose stem-alternation signatures are less diffuse. This is consistent with the qualitative AIHL assessments, where synthetic systems are often penalized for excessive or irregular suppletion. At the same time, most simulated conditions fall within or near the empirical range derived in the previous section, suggesting that the simulator often produces broadly naturalistic signature-distribution shapes even when particular systems are judged less plausible for phonological, affixal, or suppletive reasons.

The population-splitting and cascade-targeting interventions were evaluated separately, since these manipulations alter the global structure of the simulation rather than varying a parameter within the same well-mixed interaction setting.

For the population-split simulations, the component-level scores are mostly in the same range as the main experiments. The unsplit condition obtains a mean score of 2.83, as do both components in the two-component split and one of the three components in the three-component split. The three-component case, however, shows substantial heterogeneity across components: one component is rated relatively low, with a mean score of 2.17, while another receives a consistent score of 3.50 across the evaluated runs. This suggests that splitting the population does not simply raise or lower plausibility uniformly. Rather, it can produce distinct local evolutionary trajectories, some of which appear more naturalistic to the AIHL than others.

The cascade-targeting results are similarly close to the baseline range. Applying the cascade to the whole population gives a mean score of 2.50, while targeting a random quarter of the population gives a slightly higher mean of 2.67. Targeting by connectivity degree gives the highest scores in this family: both the degree-high and degree-low quarter-population conditions obtain a mean of 2.83, though the degree-low condition is more variable across runs. We therefore do not take these results to show a strong effect. At most, they suggest that restricting a cascade to a subset of the social network can produce morphologies judged at least as plausible as those obtained when the cascade is applied globally. As above, however, the number of evaluated runs is small, and these results should be read as diagnostic rather than conclusive.

Exact scores for these structural interventions are given in Tables 9 and 10, with the corresponding visual summaries in Figs. 12 and 13.

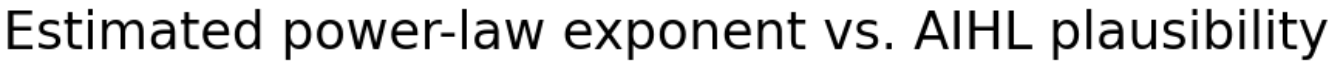
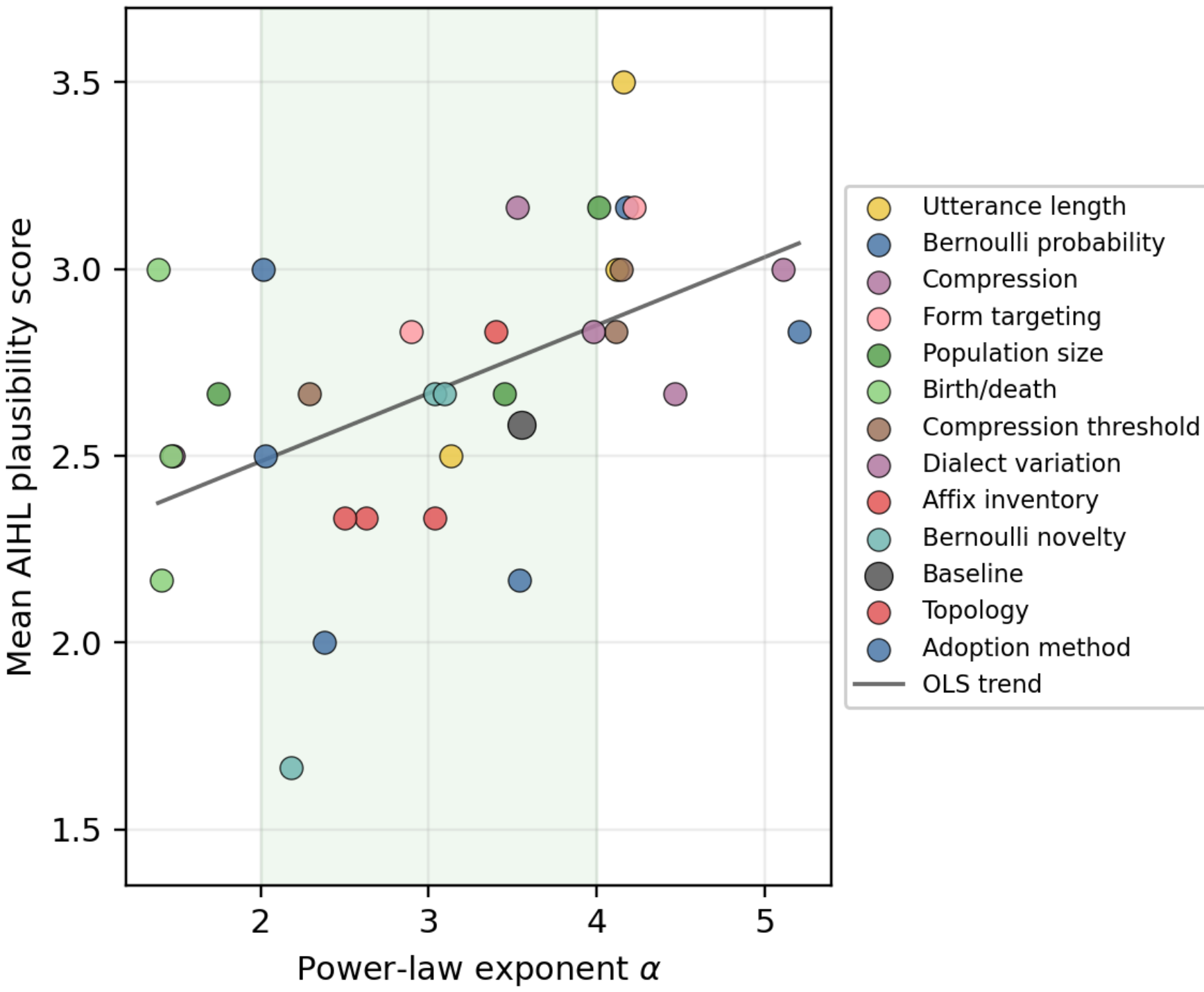

**Figure 9.** Relationship between fitted rank-frequency slope and AIHL plausibility for the main AIHL-scored experiment sweep. Each point represents an experimental condition, colored by the parameter family varied relative to the baseline. The horizontal axis gives the fitted power-law exponent for the terminal stem-alternation signature distribution, while the vertical axis gives the mean AIHL plausibility score. The shaded interval marks the approximate empirical range of exponent values observed in the natural-language stem-allomorphy analysis. The OLS line is shown only as a descriptive trend.

| Rank | Condition | Mean | SD |
|---|---|---|---|
| 1 | Degree-high (frac. 0.25) | 2.83 | 0.47 |
| 2 | Degree-low (frac. 0.25) | 2.83 | 0.94 |
| 3 | Random (frac. 0.25) | 2.67 | 0.62 |
| 4 | All | 2.50 | 0.71 |
| 5 | Targeted subset | 2.50 | 0.00 |

**Table 9.** Ranked AIHL plausibility scores for cascade targeting conditions. Scores are on a 1–4 scale and are averaged over the AIHL-scored simulation seeds. SD is the population standard deviation across scored runs. The Baseline is represented by the ‘All’ Condition.

| Count | Sub-population | Mean | SD |
|---|---|---|---|
| 1 | 0 | 2.83 | 0.85 |
| 2 | 0 | 2.83 | 0.47 |
| 2 | 1 | 2.83 | 0.47 |
| 3 | 0 | 2.17 | 0.47 |
| 3 | 1 | 3.50 | 0.00 |
| 3 | 2 | 2.83 | 0.47 |

**Table 10.** AIHL plausibility scores for individual sub-populations in the population-split simulations. The first column indicates how many sub-populations the initial population is split into, the second column provides an index for the sub-population for which the scores are calculated. Scores are on a 1–4 scale and are averaged over the AIHL-scored simulation seeds. SD is the population standard deviation across scored runs. The Baseline is represented by Count 1, i.e. no population split.

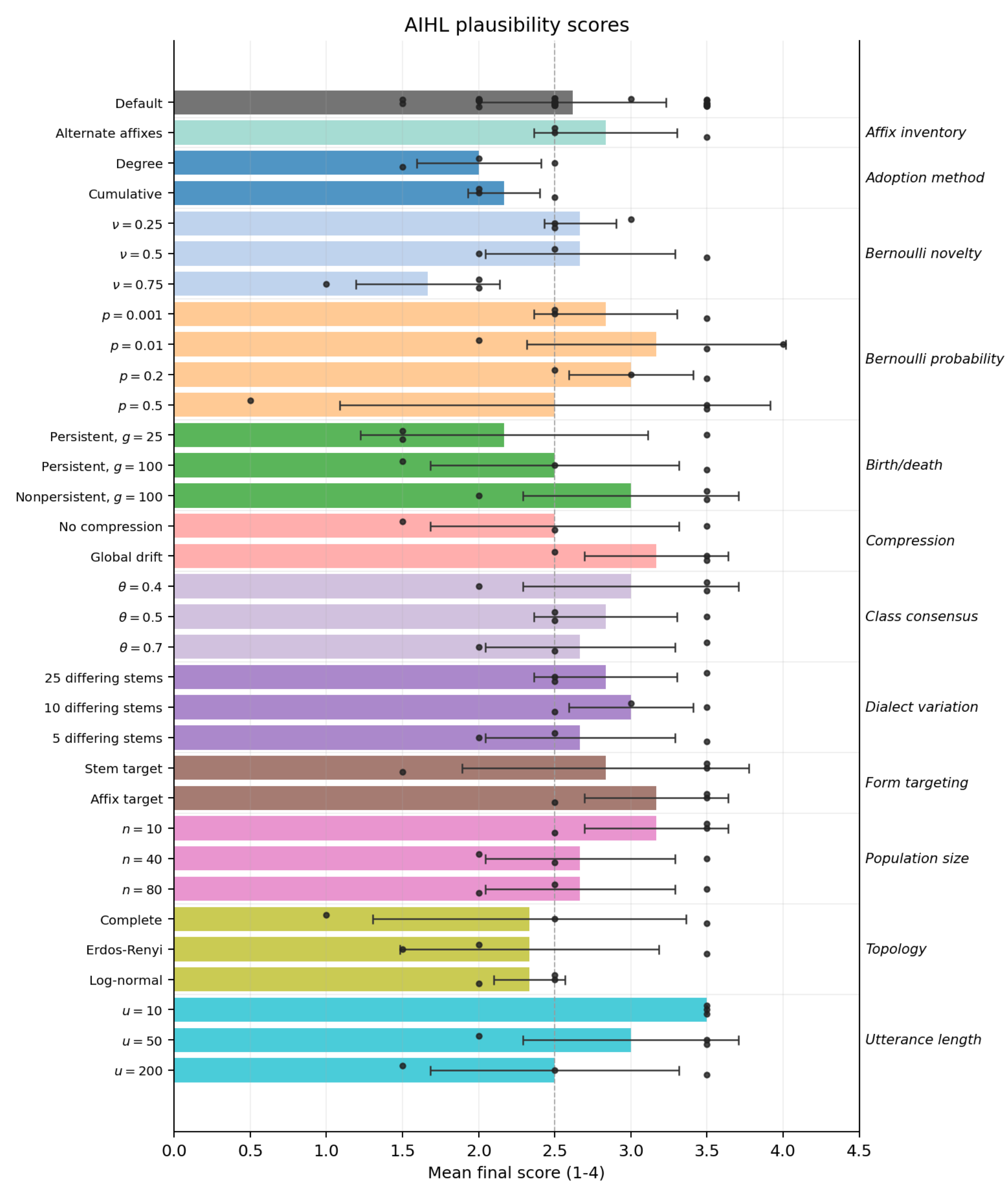

**Figure 10.** AIHL plausibility scores by experimental condition. Bars show mean final score across AIHL evaluations, error bars show standard deviation, and points show individual runs. Higher scores indicate greater judged morphological plausibility.

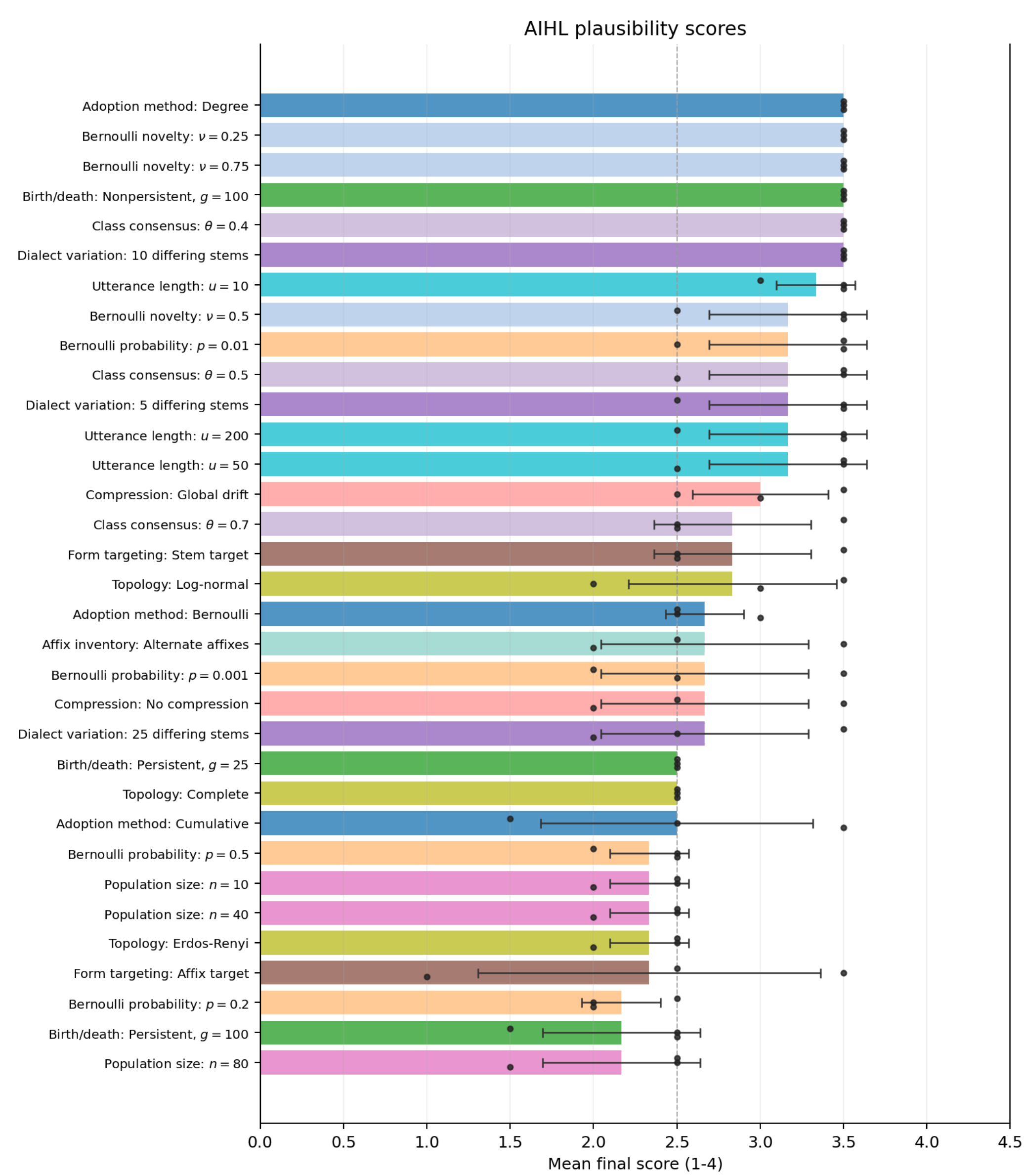

**Figure 11.** AIHL plausibility scores by experimental condition, ordered by decreasing AIHL score. Bars show mean final score across AIHL evaluations, error bars show standard deviation, and points show individual runs. Higher scores indicate greater judged morphological plausibility.

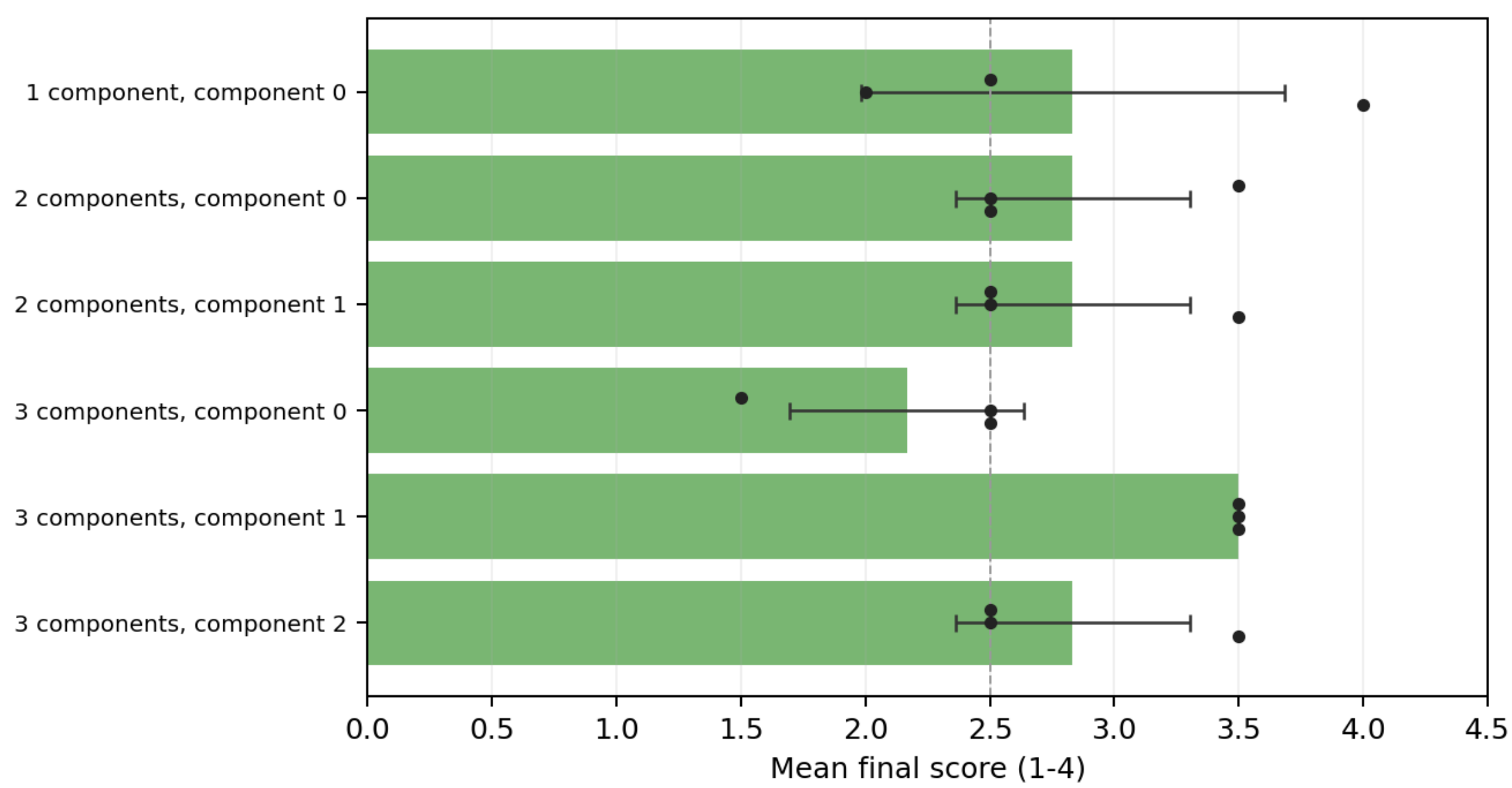

**Figure 12.** AIHL plausibility scores for individual components in the population-split simulations. Bars show mean final score across AIHL evaluations, error bars show standard deviation, and points show individual runs.

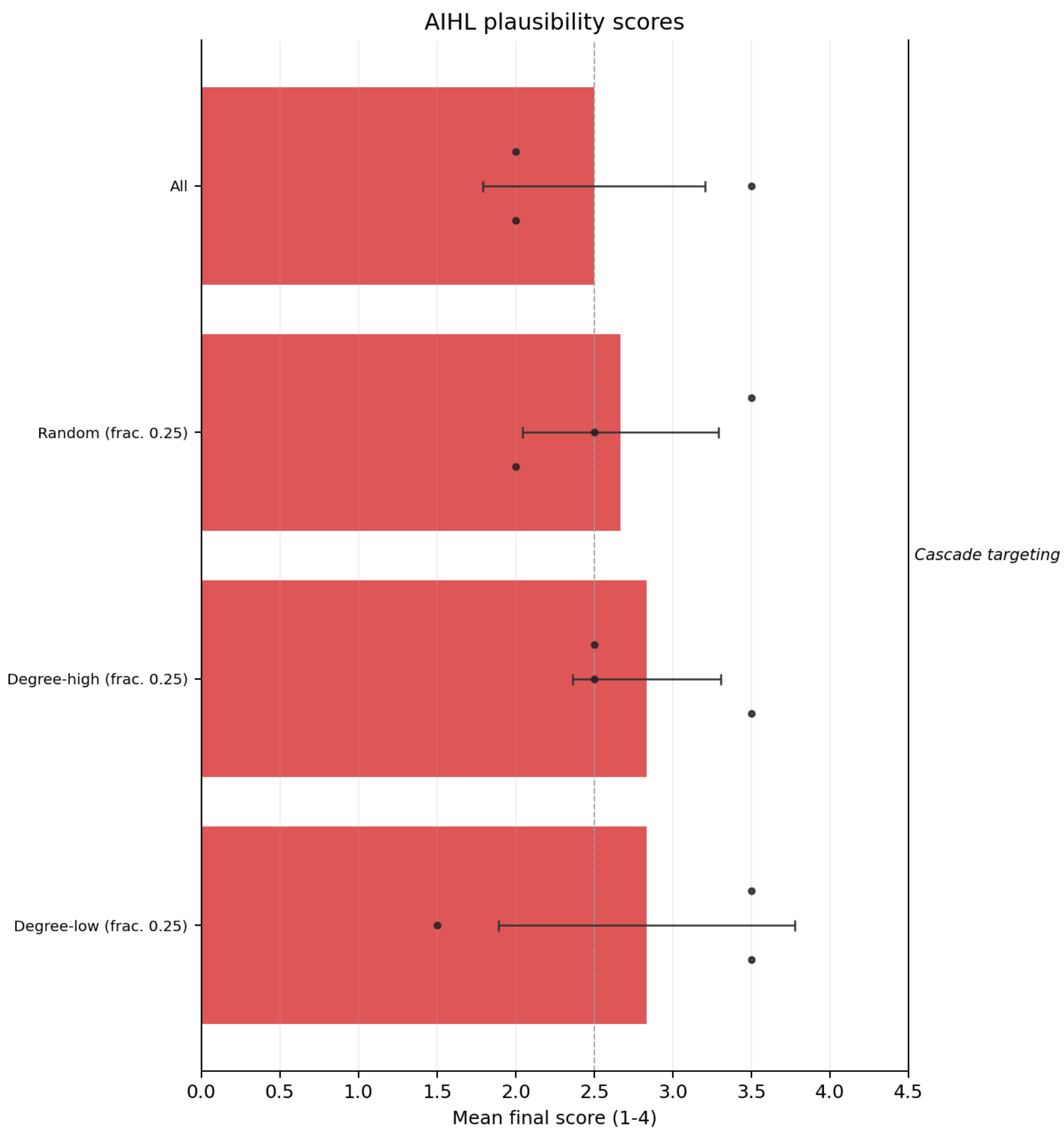

**Figure 13.** AIHL plausibility scores for cascade-targeting conditions. Bars show mean final score across AIHL evaluations, error bars show standard deviation, and points show individual runs.

# Discussion: Conclusions and Limitations

## Emergent morphological structure

Language is a complex system, and the mechanisms which underlie language change remain poorly understood. Motivated by the empirical phenomena of stem allomorphy and paradigmatic alternations, we have introduced a multi-agent system for computationally modeling language change. The model suggests that rich morphological dynamics can arise from the interaction of individually simple mechanisms: local communication, agent adoption behaviour, lexicon-level structure, and compression. We systematically explored the parameter space of our model in order to understand the key drivers of change in the system. Using what we learned we carried out a series of case studies where we applied this machinery to model several attested instances of language change ("Case studies" section in the Supplementary Materials).

## Model limitations

Our work has several limitations. Despite the number of elements incorporated into our model, it remains a substantial abstraction from the full complexity of natural language. Agent behaviour is simplified, and the representation of the lexicon is relatively rigid. These simplifications make the model tractable, but they also constrain how its outputs should be interpreted. A further challenge is the difficulty of empirically verifying the results of the model. Language change is historical, noisy, and only partially observed, making it difficult to verify computational simulations against real-world trajectories.

## Naturalness, signature distributions, and evaluation

We have nevertheless taken steps towards addressing these shortcomings. We empirically investigated and quantified the effect of stem allomorphy in several present-day languages. Through this analysis we observed that natural languages tend to show a small number of frequent stem-alternation signatures with a long tail of rarer patterns, providing a distribution target for our simulations. The AI Historical Linguist provides a complementary assessment of the final systems, allowing us to grade the behaviour of the simulations and to identify why some generated languages look more natural than others. One recurring diagnostic in those assessments is the incidence of suppletion, which we discuss next.

## The incidence of suppletion

One important feature of some of the simulations that the AIHL evaluators remarked on was the incidence of suppletive alternations in the evolved languages. High numbers of lexemes with one or more suppletive alternations was one reason given for rating a given language as being unnatural.

It is thus worth spending some space discussing what, after all, is a natural level of suppletion? The Surrey Suppletion Database [16, 225] provides for one estimate of this. This dataset consists of morphological instances of suppletion across various grammatical categories for 34 languages from 33 language families. (The one family with two, distantly related, languages included in the set was Uralic, with Hungarian and Komi.) Of this sample, 4 languages showed no instances of suppletion. For the remaining 30 languages, we extracted examples and subdivided them according to grammatical category. The resulting counts by language and category are shown in Fig 14. The bulk of cases, 90%, involve no more than 3 instances of suppletion for a given language/category pairing.

Japanese verbs turn out to be one of the cases with high incidence, with 9 verbs listed that have suppletive alternations. Most of the Japanese verbal stem alternations involve the honorific system, where different verb roots are used to express the 'polite' or 'humble' forms. Note that there are regular morphological operations to derive these forms for most verbs, so it is fair to count honorific inflection as part of the Japanese verbal paradigm, and therefore reasonable to consider the use of alternate stems to be instances of suppletion. In fact the Surrey data for Japanese is probably an undercount. Two verbs that have suppletive alternations for honorific forms that are not listed in the dataset are *shiru* 'know' and *morau*

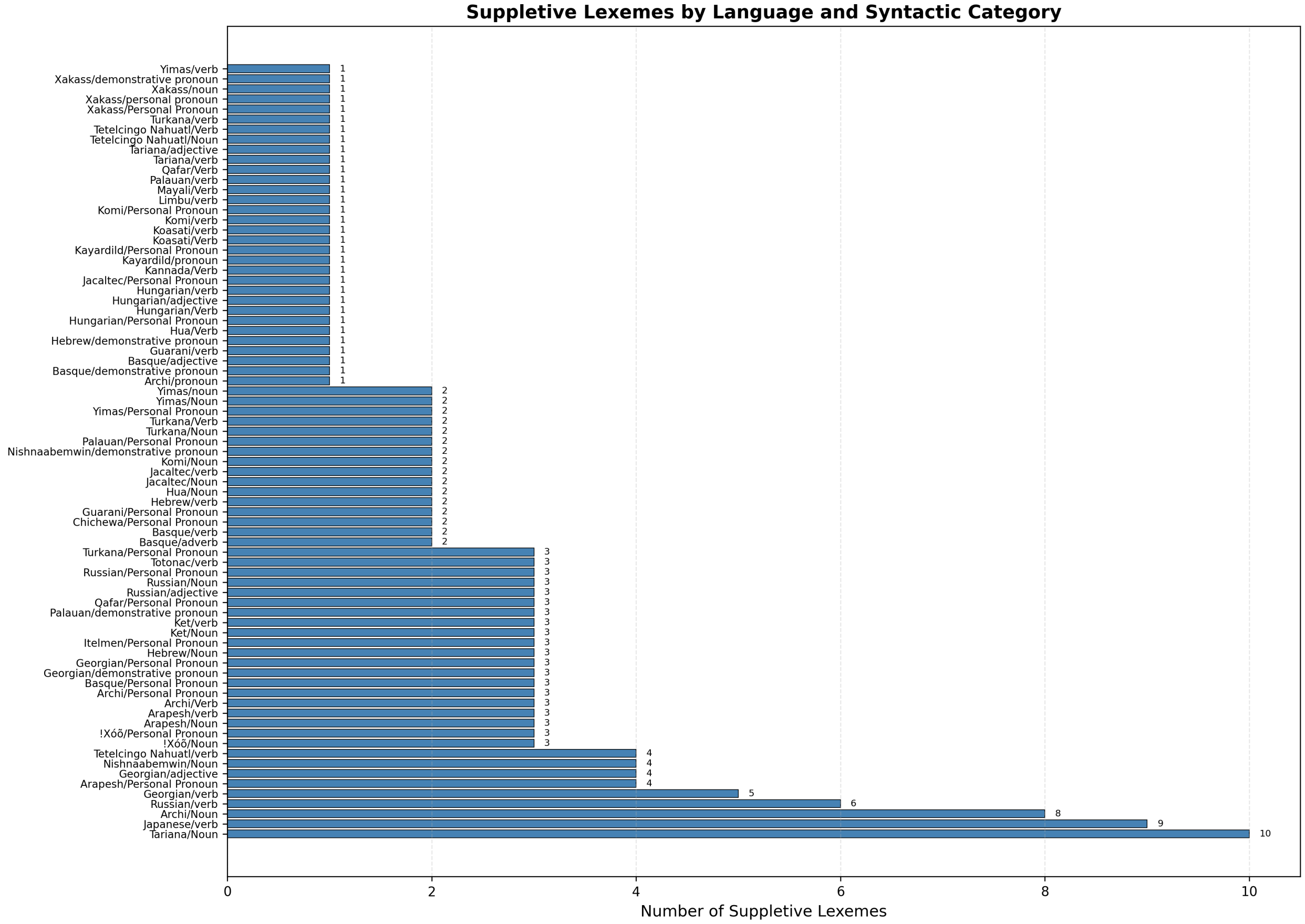

**Figure 14.** Numbers of lexemes in a given category and language that show suppletive stem alternations from the Surrey Suppletion Database [225].

'receive'. So systems with ten or more lexemes in a given category showing suppletion are not unheard of, though relatively rare.

Honorific systems are particularly ripe for lexical substitutions. Javanese, for example, is famous for its multiple speech levels. Alongside the plain speech register *ngoko*, there is a polite register *krama* that is used when showing respect to the addressee, and yet another register *krama inggil* that is used when referring to an honored individual. *Krama* consists of 850 words that are substituted for their *ngoko* equivalent when speaking in *krama* style, and *krama inggil* has 260 words [226]. Thus the *ngoko* word for 'house' is *omah*, but one uses the word *griya* when speaking in *krama* and *dalem* in *krama inggil*. Unlike the case of Japanese verbs, the Javanese cases do not involve morphological paradigms and thus it would not be fair to consider them to be instances of suppletion. However it does suggest that if an honorific system becomes grammaticalized, as it has in Japanese, that high rates of suppletion would not be unexpected. Note that the largest set of suppletive lexemes in the Surrey dataset is for Tariana (Arawak) nouns, all of which consist of terms for family relatives—another category where languages often have multiple vocabularies depending on situation. All of this suggests that how common suppletion is expected to be depends on how likely it is that alternative expressions for given referents will exist in the language.

The research paper describing the Surrey Suppletion Database [16] reviews three main properties of suppletion. These are not exceptionless, but hold broadly of suppletion systems across multiple languages. The first property is that suppletion tends to occur in high frequency lexemes, which means in turn that language users are more likely to encounter them and that they are more likely to be learned. Needless to say, frequency falls out as a consequence of the model we have described above since higher frequency lexemes are more likely to be communicated by the agents, and there is therefore more chance of a hearer adopting a novel form in higher frequency lexemes.

The second property is that suppletion tends to relate to *inherent* versus *contextual* inflection. Contextual inflection is determined by context, and includes well-known cases of agreement morphology, case marking on nouns and so forth. Inherent inflection includes such categories as number on nouns, tense, aspect or mood on verbs, and comparison of adjectives. Generally speaking, suppletion tends not to occur with contextual inflection. For example, [204] argues that true suppletion cannot be sensitive to case marking in nominal paradigms, though stem alternants involving modifications of the same etymon do occur. But there are of course many exceptions to this generalization as [16] note, comparison of adjectives (*good*/*better*), and suppletive stem alternants in Romance verbal inflection (Table 3, and see the "Case studies" section in the Supplementary Materials) being two such exceptions. Insofar as our model has no semantics associated with the morphological paradigms it produces, it is in any case agnostic about the inherent-contextual distinction.

The final property discussed by [16] is morphological systematicity: "the patterning of [suppletive] stems is not unconstrained but is determined by the morphological system in place" (p. 13). This morphological patterning has of course been the topic of much of the literature on stem alternations, and figures prominently in the book-length study in [198] as well as various studies of Romance verbal stem alternation [7–11, 227]. It also falls out from several of the mechanisms proposed in our model, including *stem-as-target* adoption and various compression methods, with the *stem-as-target* adoption protocol figuring prominently in our modeling of Romance verbal morphology in the "Case studies" section of the Supplementary Materials. As we noted in the "Model details" section, these protocols correspond to the assumption that the agent keeps track of the forms used within a paradigm (or in the case of *affix-as-target* slots with the same marking across paradigms), and substitutes all instances when encountering a new form. But exceptions do occur: [16] cite the verb 'come' in Georgian, which has a pattern of four distinct roots not seen with other verbs. This is of course expected in a complex dynamic system, where mechanisms are in place that favor certain outcomes, but at the same time do not guarantee them.

## Research questions revisited

We repeat, for convenience, our four research questions:

**RQ1.** Can stem-alternations and paradigms arise from lexical borrowing, and what assumptions do we need to make beyond neutrality in order for such alternations to arise?

**RQ2.** What properties of social networks and agent interaction best predict the emergence of stem-alternations and paradigms?

**RQ3.** Can we make use of Large Language Models in the evaluation of evolved synthetic systems?

**RQ4.** Can our system simulate actual historical changes?

In answer to **RQ1**, we have argued that natural stem-alternation patterns and paradigms can arise from the process of agents borrowing forms from other agents. In order for this to work, one must also assume either some form of compression mechanism whereby agents periodically regularize their systems and/or the option of borrowing a form not only into the slot in which it was heard being used, but into other slots that shared the same previous form (*stem-as-target* and *affix-as-target*). In the simulations, lower rates of adoption and lower bias toward adopting novel forms were also critical, with these in turn leading to a lower incidence of suppletion, which was deemed more natural. This latter point, however, needs to take into consideration the discussion in the previous section.

In answer to **RQ2**, scale-free networks seem to be associated with more natural results, as did random Bernoulli adoption.

For **RQ3**, we introduced the *AI Historical Linguist*, an automated system for evaluating the outputs of historical simulations. This system was instrumental in our assessments reported in the Methods section (and more particularly the section entitled "Evaluating generated languages with the AIHL"). While we certainly do not view the AIHL as final, we do expect that as Large Language Models continue to improve, so will the diagnostic strengths of the system. In any event, to our knowledge the AIHL is the first automated method proposed for evaluating the output of historical linguistic simulations.

Finally, for **RQ4**, we demonstrate in the Supplementary Materials the use of our system in simulating three rather diverse sets of historical changes in various language families.

## Future directions

Several avenues remain open for future work. The present study focuses on stem and inflectional marker allomorphy, but the same framework could be extended to other instances of attested language change. We have described a few examples in the Case Studies, but there are clearly many more.

For example, it should be possible to extend the model to forms displaying general agglutination, allowing one to model transitions between agglutinative and fusional languages. One question that relates to this is why agglutinative systems seem to be very persistent, so that for example the agglutinative morphology of Dravidian or Turkic languages seems to have retained its general form over thousands of years? Whereas the heavily fusional morphology of Indo-European eroded in many branches of the family over the same period?

Another case of interest is synchronized changes occurring in related but geographically separated languages. For example, one of the properties of the Celtic mutation system studied in the ”Case studies” section in the Supplementary Materials is that once the historical changes were initiated, they kept developing further in separate branches of the family. As noted there, other consonant mutation systems show similar historical trends, and more generally, one often finds that related languages undergo parallel, but apparently independent changes in the course of their history. For example, similar vowel shifts occurred in a variety of West Germanic languages at roughly the same time. In the Celtic simulations, these parallel developments were handled by exogenous mechanisms, injecting changes into the system at certain points in the evolution. A more realistic model would have such parallel processes arise endogenously from properties of the languages.

Finally, we would like to incorporate recent work on generating Constructed Languages [228, 229] into this framework. These systems allow one to specify a number of properties of the desired language, which could serve as an initial state for further evolution.

## Supporting information

**S1 Figures. Figures in attached figures.pdf.**

**S1 File. Supplementary materials in attached supplementary_materials.pdf.**

## Acknowledgments

TBD

## Author contributions

Both authors contributed equally to the conception, design, and writing. AK designed most of the agent modeling and ran the experiments reported in the "Materials and methods" section. RS developed the Pynini-based linguistic components and the AI Historical Linguist and ran the simulations in the "Case studies" section in the Supplementary Materials.

# Supplementary materials

## Adoption policies

In this section we provide further details on our model. As described in the "Model" section, we model communication between a pair of agents as the transmission of tokens sampled from the speaker's lexicon to the listener's lexicon. When a listener is provided with a list of utterances from their neighbors they must decide how to absorb these utterances. This is handled by an Adoption policy. In the experiments reported here we restrict attention to three simple adoption rules which we call Bernoulli, degree-scaled, and cumulative. Fig 15 illustrates how each of these rules updates a focal cell from a shared inbox of heard tokens.

**Bernoulli.** This is the baseline policy. Each received utterance is processed independently. If the heard form differs from the listener's current form in the relevant cell, it is adopted with a fixed probability. (Note that implicit in this construction is the assumption that each pair of communicating agents agree on the meaning (lexeme and slot) of what they are communicating about.)

**Degree-scaled.** This is a variant of the Bernoulli adoption policy where the probability of adoption is scaled by the degree, i.e. the number of connections, of the sender.

$$p_{\text{eff}}(j) = \min\big(1,\ p \cdot \max\{0, \text{deg}(j)\}\big),$$

Thus forms from more 'popular' speakers are more likely to be adopted.

**Cumulative.** The previous adoption policies operate on each utterance in the queue of heard utterances one-by-one. This adoption policy differs from the previous policies in that it considers all heard utterances in the queue at once. Here the agent aggregates all heard forms per lexeme and slot and selects the form that was heard most often during that time step.

We explore two further methods which can be seen as variants of the Bernoulli adoption policy: stem-as-target and affix-as-target. Fig 16 contrasts the scope of these two variants against the default cell-local update. These differ from the default behaviour of replacing the slot corresponding to the heard form alone. For the stem-as-target variant all forms within the paradigm of the lexeme have their stem altered to the stem of the adopted form. For the affix-as-target variant all forms sharing the same slot have their affix replaced by the affix of the adopted form.

**(a) Bernoulli**

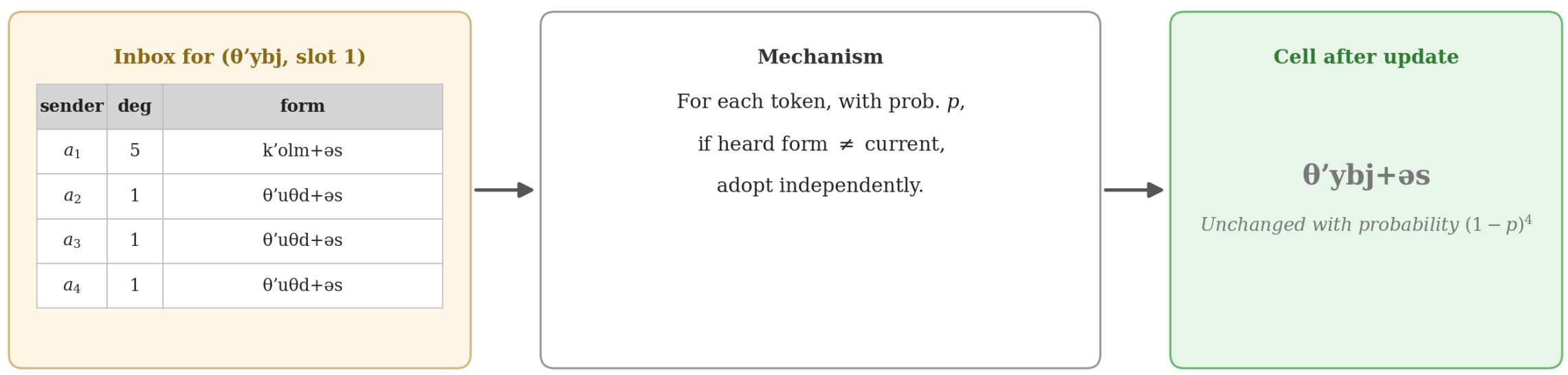

**(b) Degree-scaled**

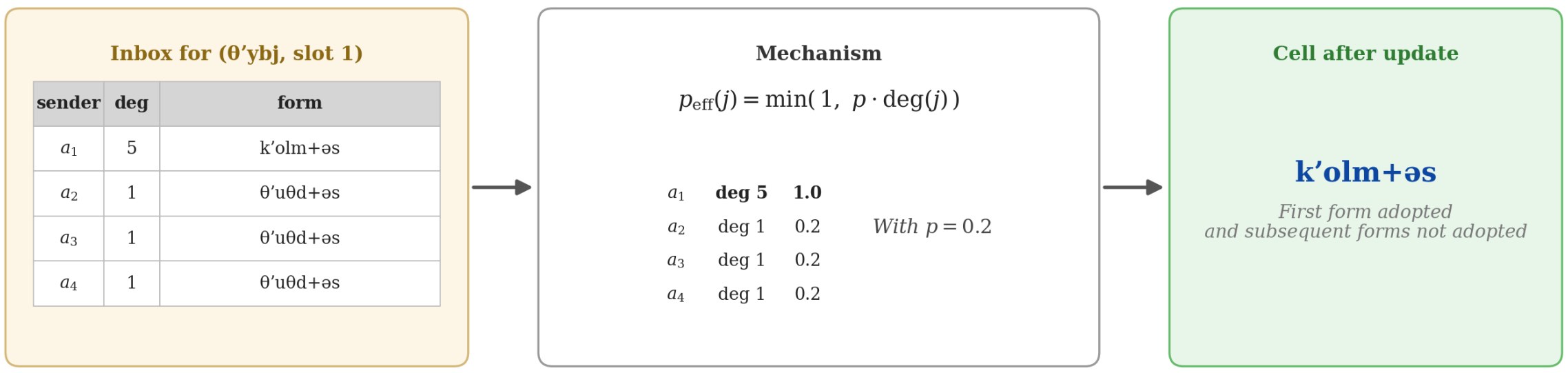

**(c) Cumulative**

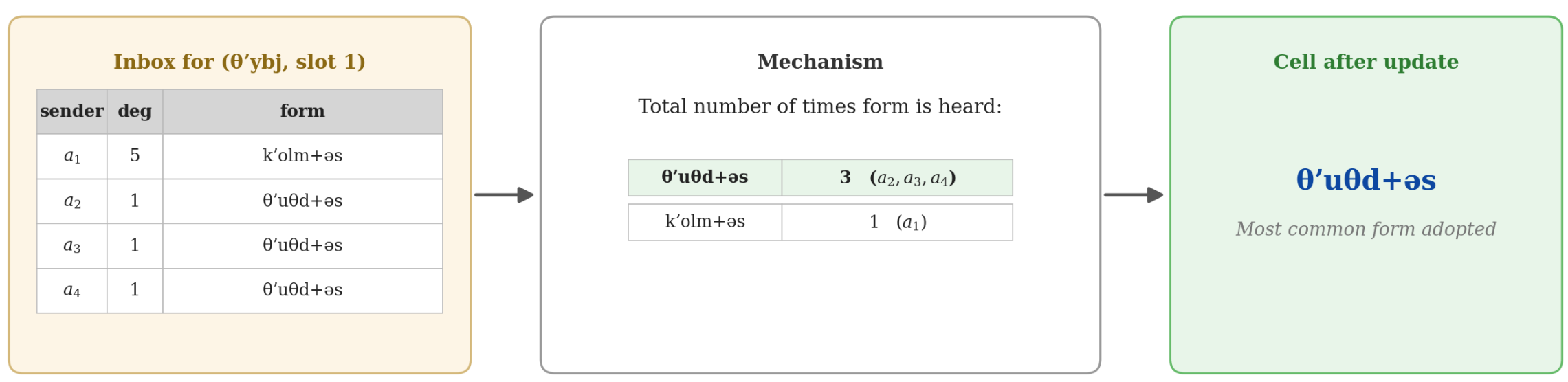

**Figure 15.** The three adoption methods applied to a common inbox for a single focal cell. Bernoulli treats each received token as an independent coin flip and may leave the cell unchanged; degree-scaled biases adoption toward high-degree senders; cumulative aggregates per-form counts across the inbox and deterministically adopts the most heard form.

**Initial Lexicon**

| | 0 | 1 | 2 |
|---|---|---|---|
| **p'epn** | p'epn+ə | p'epn+əs | p'epn+ət |
| **θ'ybj** | θ'ybj+ə | θ'ybj+əs | θ'ybj+ət |
| **w'uff** | w'uff+ə | w'uff+əs | w'uff+ət |
| **f'œfh** | f'œfh+ə | f'œfh+əs | f'œfh+ət |

**(a) Default**

| | 0 | 1 | 2 |
|---|---|---|---|
| **p'epn** | p'epn+ə | p'epn+əs | p'epn+ət |
| **θ'ybj** | θ'ybj+ə | **θ'uθd+is** | θ'ybj+ət |
| **w'uff** | w'uff+ə | w'uff+əs | w'uff+ət |
| **f'œfh** | f'œfh+ə | f'œfh+əs | f'œfh+ət |

**(b) Stem-as-target**

| | 0 | 1 | 2 |
|---|---|---|---|
| **p'epn** | p'epn+ə | p'epn+əs | p'epn+ət |
| **θ'ybj** | **θ'uθd+ə** | **θ'uθd+is** | **θ'uθd+ət** |
| **w'uff** | w'uff+ə | w'uff+əs | w'uff+ət |
| **f'œfh** | f'œfh+ə | f'œfh+əs | f'œfh+ət |

**(c) Affix-as-target**

| | 0 | 1 | 2 |
|---|---|---|---|
| **p'epn** | p'epn+ə | **p'epn+is** | p'epn+ət |
| **θ'ybj** | θ'ybj+ə | **θ'uθd+is** | θ'ybj+ət |
| **w'uff** | w'uff+ə | **w'uff+is** | w'uff+ət |
| **f'œfh** | f'œfh+ə | **f'œfh+is** | f'œfh+ət |

**Figure 16.** The three adoption-target modes applied to the same heard token. The default mode updates only the heard lexeme-slot cell; stem-as-target additionally propagates the heard stem across every slot of the target lexeme's paradigm; affix-as-target propagates the heard affix to every cell in the lexicon that previously carried the listener's old affix.

## Compression policies

Compression policies serve as our computational model of analogy. In our experiments these act to mutate an agent's morphology in order to reduce linguistic variation. In our case we focus mainly on stem alternation signatures, and our simple compression policies aim to model generalisation of dominant stem alternation signatures as well as erosion of more complicated paradigms. We briefly describe three simple compression policies, whose effect on a shared population of stem-alternation signatures is shown in Fig 17.

**No compression.** In this setting we apply no compression at all. The lexicon only changes through diffusion of heard utterances.

**Class consensus.** This is the default compression policy we use. In this case lexemes are grouped by the set of alternants that appear in their stem alternation signature. Thus a signature such as $(0, 0, 0, 1, 1, 0)$ and $(0, 1, 1, 1, 1, 1)$ will form part of the same group, as they have the same number of distinct stem alternations. If one signature accounts for at least a threshold, $\theta$, of the group that pattern is deemed a dominant pattern and all minority signatures are rewritten to share the same signature. This allows for a 'winner-takes-all' effect within groups, while preserving the broad diversity of stem alternation signatures within the lexicon.

**Global drift.** This policy is a softer alternative to class consensus. Instead of collapsing every signature within a group to the modal pattern, one compares each signature to a nearby signature as measured by Hamming distance. A signature is updated only if nearby competitors are sufficiently dominant within that local neighborhood. As with the class consensus method, dominance is dictated by a threshold parameter. The effect of this policy is to pull rare signatures toward nearby common patterns, as well as allowing stem alternants to vanish altogether.

**Decision-tree.** This policy frames compression as a learning problem rather than a direct rewrite over signatures, and unlike the above-described policies, is not limited specifically to stem alternants. For each cell in the agent's lexicon we extract a triple of the realized affix, the stem on which it occurs, and a small set of features describing that stem environment. A decision tree classifier is fit to these triples, learning to predict the appropriate form from the environmental features. Once trained, the classifier is applied back across the lexicon and any cell whose current form disagrees with the prediction is rewritten accordingly. The procedure is illustrated in Fig 18 for the Korean nominative case described in the "Case studies" section. In Modern Korean, vowel-final stems take *-ka* as the nominative marker, and consonant-final stems take *-i*: a tree fit on such forms learns this generalisation and corrects any outliers accordingly. The effect is to regularize allomorphy by replacing inconsistent forms with a rule conditioned on the local environment, rather than collapsing variation purely on signature similarity.

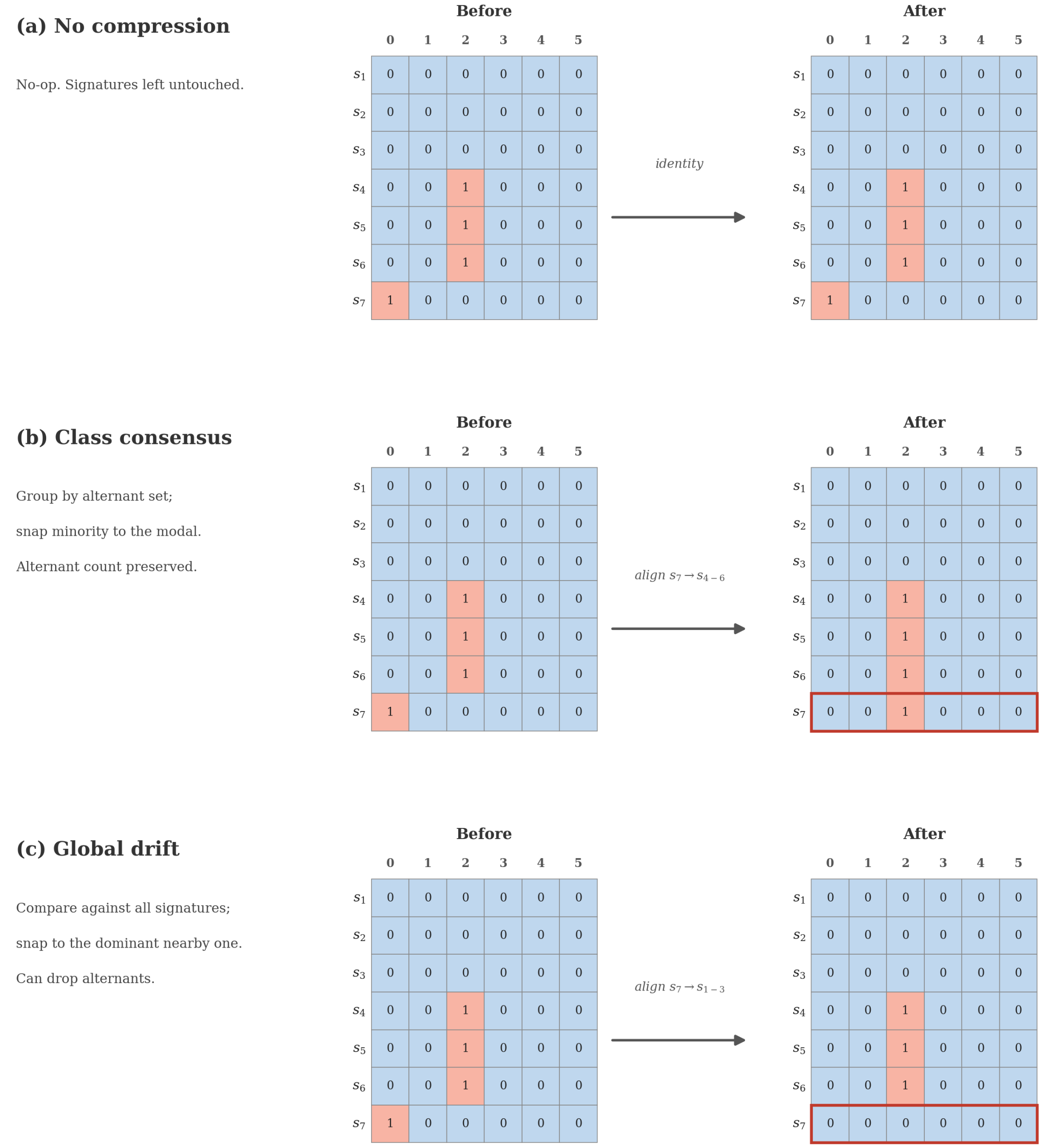

**Figure 17.** The three compression policies applied to a shared population of seven stem-alternation signatures. No compression leaves the signatures untouched; class consensus snaps minority signatures within an alternant-set class to the modal pattern, preserving the alternant count; global drift snaps each signature to a dominant nearby one and can drop alternants altogether.

## Decision-tree compression

Fit a classifier on (environment, affix) pairs in the agent's lexicon,

then apply it back to rewrite any cell whose form disagrees with the prediction.

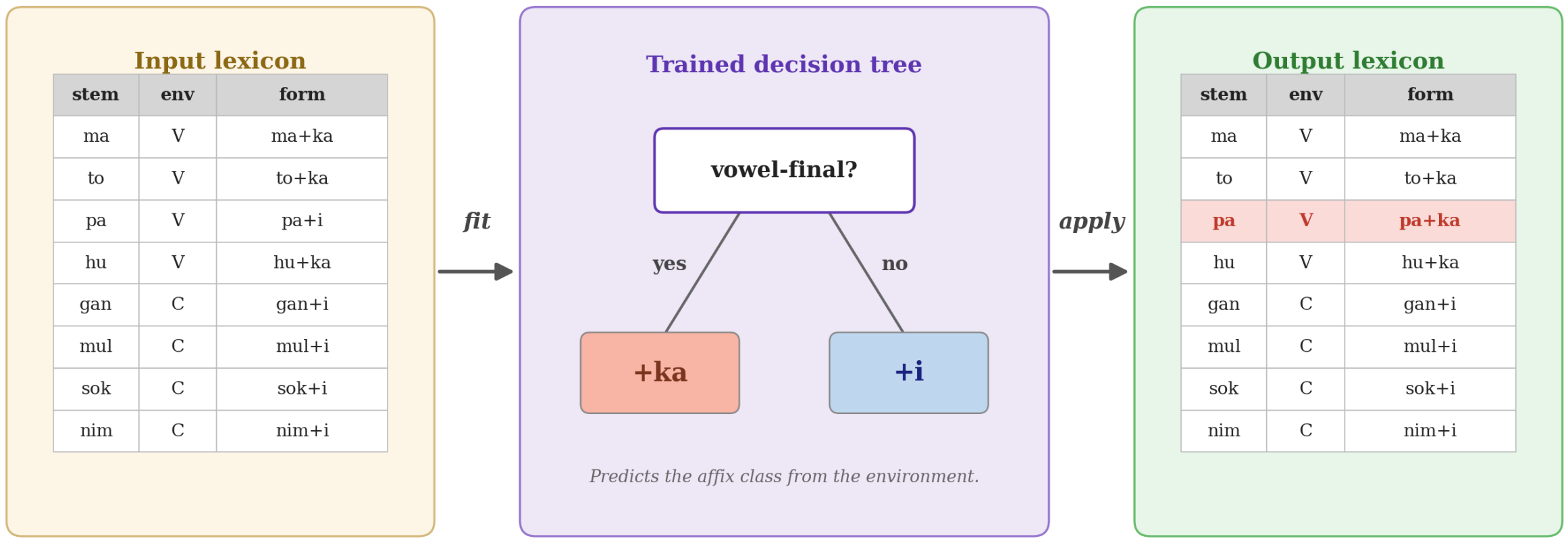

| stem | env | form |
|---|---|---|
| ma | V | ma+ka |
| to | V | to+ka |
| pa | V | pa+i |
| hu | V | hu+ka |
| gan | C | gan+i |
| mul | C | mul+i |
| sok | C | sok+i |
| nim | C | nim+i |

| stem | env | form |
|---|---|---|
| ma | V | ma+ka |
| to | V | to+ka |
| pa | V | pa+ka |
| hu | V | hu+ka |
| gan | C | gan+i |
| mul | C | mul+i |
| sok | C | sok+i |
| nim | C | nim+i |

*Example: Korean nominative /-i/ vs. /-ka/. The tree learns V-final → +ka, C-final → +i and rewrites pa+i → pa+ka.*

**Figure 18.** Decision-tree compression illustrated on a Korean nominative example. A classifier is fit to (environment, affix) pairs in the agent's lexicon and applied back, learning that vowel-final stems take *-ka* and consonant-final stems take *-i*, and rewriting the outlier accordingly.

## Phonological rules supplied by the library

We make use of the Pynini library [19, 20] to model phonological change, allowing the user to extend the system to cover any phonological changes desired. However a set of phonological rules is provided by default. These rules are inspired by rules that are commonly found in historical change and fall into the broad categories of accent patterning, phonetic reduction, deletion, assimilatory processes, lenition and fortition.

**Accent shift** Deletes any primary accent that has another primary accent to its right.
*Example:* pa'ka'na ⇒ paka'na.

**Deaccent** Any primary accent that precedes another primary becomes secondary; captures systems with a single primary plus earlier secondary prominences.
*Example:* pa'ka'na ⇒ pa"ka'na.

**Iambic reversal** Move a secondary accent from a syllable directly preceding a primary accent, to the preceding syllable.
*Example:* aba"la'ta ⇒ a"bala'ta.

**Reduce vowel** Reduces vowels in the domain preceding the primary accent; models stress□conditioned centralisation.
*Example:* pa'ka ⇒ pə'ka.

**Fully reduce vowel** All pretonic vowels reduce fully.
*Example:* tika'lu ⇒ təka'lu.

**Reduce vowel after non-accent** Broad reduction of unaccented vowels to a range of reduced vowels.
*Examples:* pa ⇒ pə; pi ⇒ pɨ.

**Fully reduce vowel after unaccented sound** Broad reduction of unaccented vowels to a centralized reduced vowel.
*Example:* pi ⇒ pə.

**Schwa deletion** Deletes reduced vowels; models syncope/cluster creation after reduction.
*Example:* pəta ⇒ pta.

**Apocope** Loss of final weak vowels.
*Example:* domə ⇒ dom.

**Vowel deletion** Deletes a vowel before a morpheme boundary when the next morpheme begins with a vowel.
*Example:* pa+i ⇒ p+i.

**Voicing assimilation** A consonant devoices before a following voiceless segment (even across a morpheme boundary or accent marks). Models general regressive assimilation.
*Example:* abta ⇒ apta.

**Intervocalic voicing** Intervocalic voicing/lenition (including across morpheme boundaries).
*Example:* ata ⇒ ada.

**Final devoicing** Devoicing of obstruents in final position.
*Example:* ad ⇒ at.

**Nasal assimilation** Anticipatory place assimilation (including across morpheme boundaries).
*Example:* anpa ⇒ ampa.

**Fronting** Fronts back vowels when an i follows (possibly with intervening consonants/boundaries); models i-umlaut.
*Example:* kuti ⇒ kyti.

**Vowel assimilation** The morpheme-final vowel spreads to the following morpheme-initial vowel. Models progressive vowel assimilation.
*Example:* pa+i ⇒ pa+a.

**Latin(-like) vowel assimilation** Models vowel changes like those of Latin endings in verbal paradigms.
*Examples:* i+o ⇒ +io; e+o ⇒ +eo.

**Fortition** Strengthens glide /j/ to stop /g/ before back vowels in the environment of /l/, /n/, mimicking some aspects of Italo-Romance hardening.
*Example:* al+jo ⇒ al+go.

**Delete morph boundary** Deletes morpheme boundary symbols.
*Example:* pa+ta ⇒ pata.

A sample chained outcome:

$$\text{al+jutiə} \xRightarrow{\text{Fortition}} \text{al+gutiə} \xRightarrow{\text{Fronting}} \text{al+gytiə} \xRightarrow{\text{Apocope}} \text{al+gyti.}$$

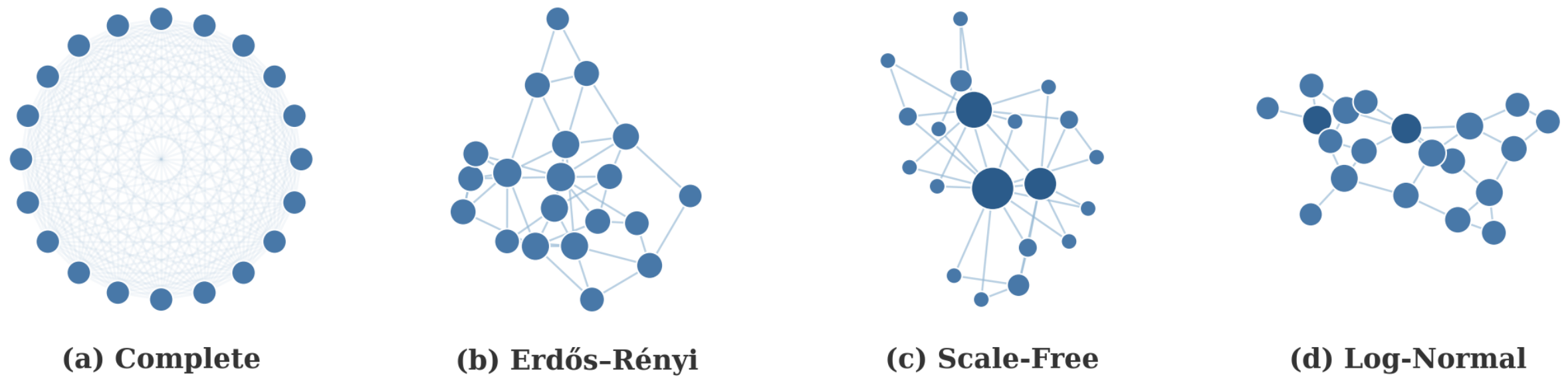

**Figure 19.** Representative social graph topologies used in the model: complete, Erdős–Rényi, scale-free, and log-normal. The complete graph maximizes connectivity; the Erdős–Rényi graph provides a homogeneous random baseline; the scale-free graph introduces hubs through preferential attachment; and the log-normal graph produces heterogeneous degree structure without the same hub dominance.

## Social graphs

The social structure of the population is represented by a graph $G = (V, E)$ with $|V| = n$ agents. In the implementation, graphs are constructed with the NetworkX library [21]. The model supports four primitive graph families: complete, Erdős–Rényi, scale-free, and log-normal. Fig 19 shows representative examples of the four topologies used in the experiments.

The complete graph is the fully connected baseline. The Erdős–Rényi family samples each edge independently with probability $p$, with a sparse default of approximately $2/(n-1)$ when $p$ is not specified. The scale-free family uses Barabási–Albert preferential attachment with parameter $m$, thereby introducing hubs. The log-normal family first samples a log-normal degree sequence with parameters $\mu$ and $\sigma$, then realizes a graph via the configuration model with self-loops removed.

A practical detail that matters for the simulations is that the implementation ensures connectivity for every generated component. If a sampled Erdős–Rényi or log-normal realisation is disconnected, or if any other generator produces multiple pieces, the code stitches those pieces together by adding a small number of random edges between components. As a result, the default single-graph case always yields one connected interaction network.

The generator can also create $k \geq 1$ disconnected components with total size $n$. In that case, each component is generated separately, each one is made internally connected by the same stitching procedure, and nodes are annotated with a component identifier. This option is mainly used when modeling social separation explicitly; later split and merge events rewire an existing graph rather than introducing a new graph family, as shown in Fig 20.

To model generational language change we introduce a birth/death process in which a subset of mentor agents is periodically replaced by a cohort of learners. Each learner is constructed from the morphology of one or more mentors in the current population and then takes over from them. We consider two regimes that differ in how the social graph reacts to this replacement, illustrated in Fig 21:

**Persistent.** Each learner inherits the social position of its mentor. The agent occupying a given node is replaced but the edges incident to that node are preserved. The network topology is therefore unchanged across generations.

**Non-persistent.** Learners replace mentors and the edges within each connected component are then re-sampled according to the same family used to build the original graph. The agent identities are renewed and the neighborhood structure changes, so the new generation interacts over a fresh social graph.

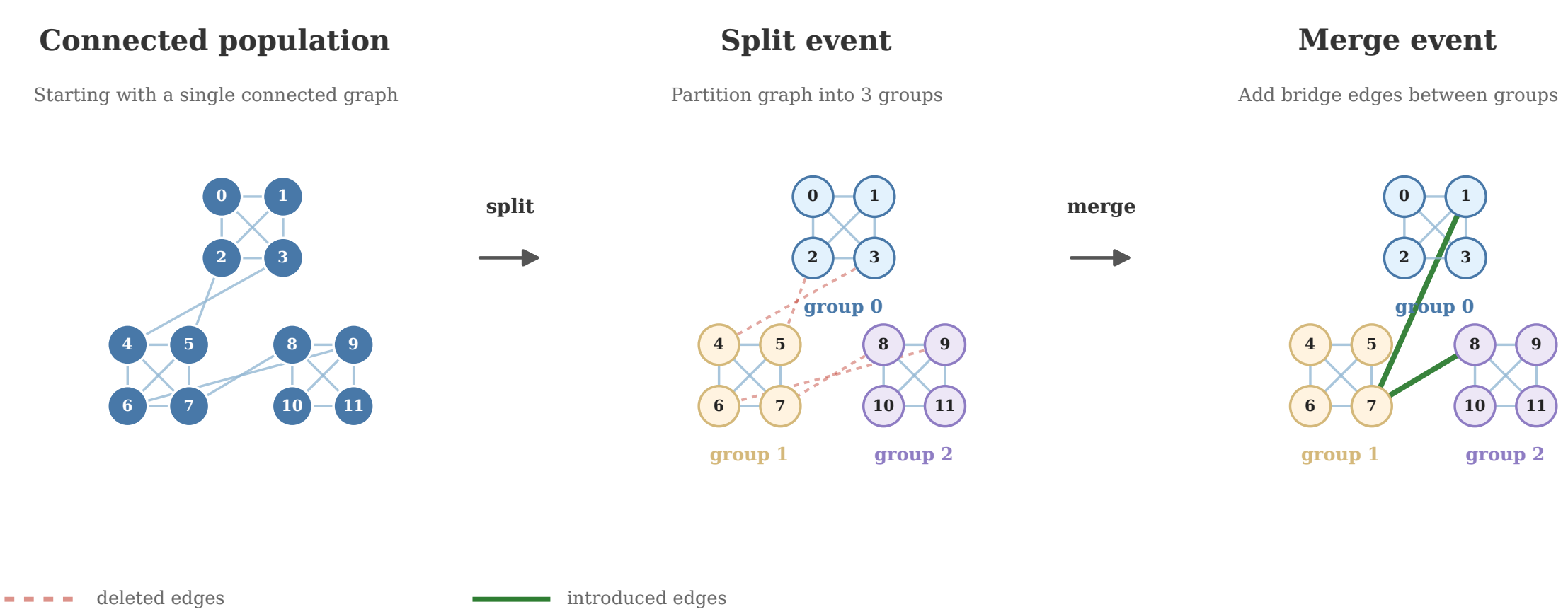

**Figure 20.** Example of a split and merge operation applied successively to a single population. Under a split instance the population is split into a set of connected components, the number of which is specified by the user. Under a merge operation disjoint components are merged by adding edges at random between the components.

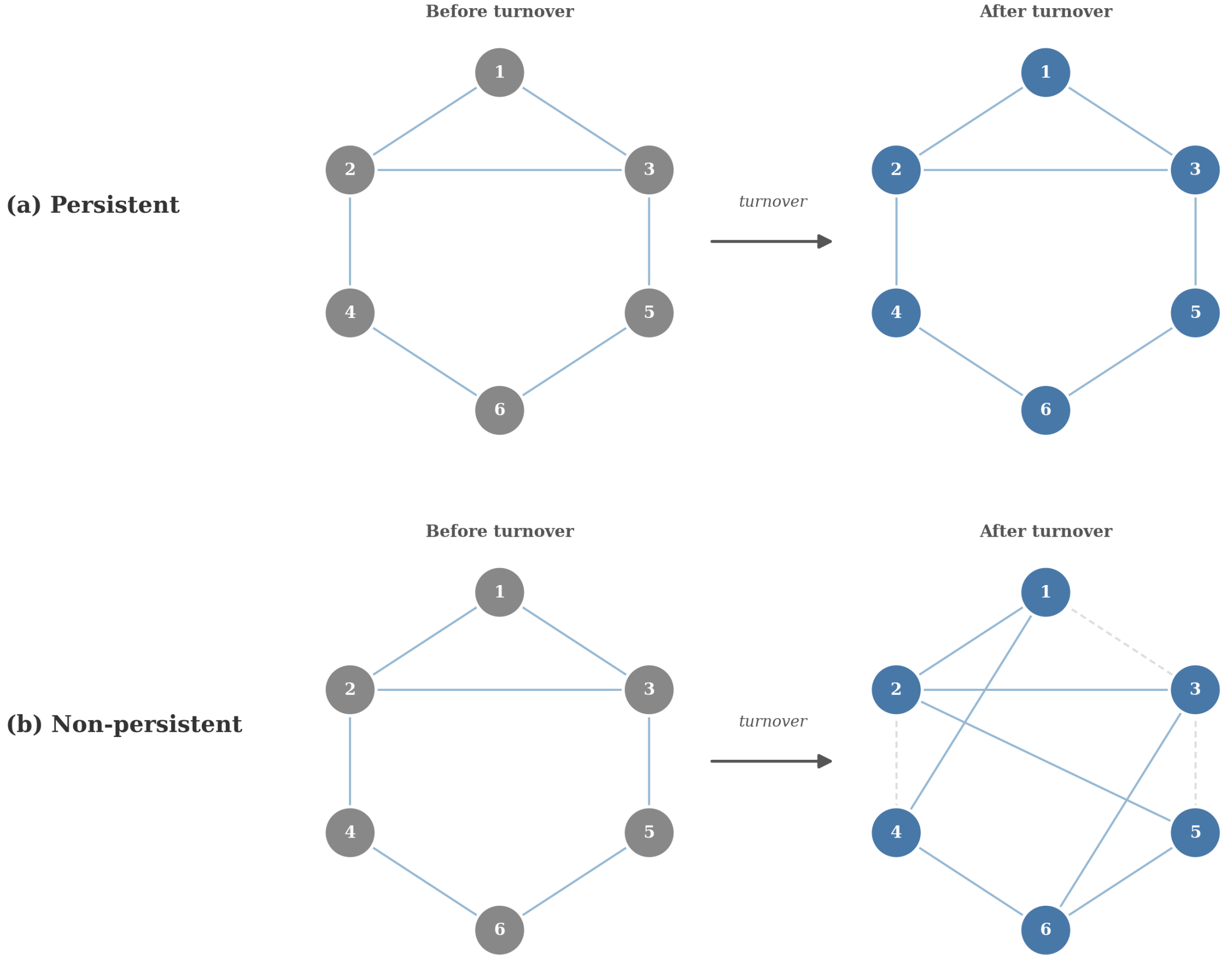

**Figure 21.** Birth/death turnover under the two persistence regimes. In the persistent regime each learner inherits its mentor's social position and the graph topology is preserved across generations; in the non-persistent regime the edges within each connected component are re-sampled, shown here by new solid edges in the post-turnover graph and the previous edges drawn as faded dashed lines.

## Proof of convergence in the analytic regime

As it stands our model of collective agent morphology evolution is highly non-linear and thus not immediately susceptible to analytic treatment. However, when we eliminate compression entirely and restrict ourselves to cell-local Bernoulli adoption policies, the model complexity reduces considerably. Compression effectively correlates distinct lexemes and slots in the lexicon. Removing compression results in a model where each cell can only change through adoption from a neighbor. Thus, for each fixed cell, the marginal dynamics reduce to a diffusive copying process. These models are effectively a set of voter-like models operating in parallel, one for each lexeme-slot cell. In this section we describe this analytic regime and demonstrate convergence of the stochastic process.

We assume a finite connected graph, $G = (V, E)$, a fixed population of $n = |V|$ agents, no exogenous events, no population turnover, no compression, and no stem-as-target or affix-as-target propagation. Adoption is Bernoulli with fixed probability $p \in (0, 1)$. Define the neighborhood of each node $v \in V$ to be $N(v) = \{u_1, \ldots, u_{d(v)}\}$, where $d(v)$ is the degree of node $v$. Restrict attention first to a single cell $c = (l, s)$ corresponding to a lexeme $l$ and slot $s$. Let $x_l$ be the probability of selecting lexeme $l$ and $y_s$ be the probability of selecting slot $s$. The probability of sampling the cell $c = (l, s)$ in a single utterance is therefore

$$q_c = x_l y_s. \tag{0.1}$$

In our simulations we impose independent Zipf-Mandelbrot distributions on the lexemes and slots, and thus $x_l$ and $y_s$. For the analytic argument below we assume $q_c \in (0, 1)$ and that each agent produces $r \geqslant 1$ utterances in a communication step.

The variable number of possible forms for a cell in the population is cumbersome if considered together. Although fixation probabilities can be computed one variant at a time, the convergence argument is clearest if we keep all variants explicit. Let $\mathcal{A}$ be the finite set of variants present for this cell, and let $X_v(t) \in \mathcal{A}$ be the random variable denoting the variant held by agent $v$ at time $t$. We bundle together the cell values for each agent into a state vector

$$\mathbf{X}(t) = (X_v(t))_{v \in V} \in \mathcal{A}^V.$$

During one time step agent $v$ receives a queue of utterances from its neighbors and decides whether to retain its current form or copy a form heard from one of its neighbours. We assume that, conditional on the current state, the utterance samples and Bernoulli adoption decisions have a joint law which gives positive probability to every finite combination of copy-source choices allowed by the one-step copy kernel defined below. This is the case, for example, if utterance samples and Bernoulli adoption coins are independent across ordered neighbor pairs. This assumption is only needed to make the positive-probability path argument below explicit.

The result proved in this section is the following.

**Theorem 0.1.** *Fix a finite connected graph $G = (V, E)$ and a single lexeme-slot cell $c = (l, s)$. Assume the analytic regime described above: fixed population, no exogenous events, no population turnover, no compression, no form-target propagation, and Bernoulli adoption with probability $p \in (0, 1)$. Let $q_c = x_l y_s \in (0, 1)$ be the probability of sampling cell $c$ in a single utterance, and let $r \geqslant 1$. Then the cell process $\mathbf{X}(t) \in \mathcal{A}^V$ reaches consensus almost surely. That is, if*

$$\tau_c = \inf\{t \geq 0 : X_v(t) = X_w(t) \text{ for all } v, w \in V\}, \tag{0.2}$$

*then*

$$\Pr(\tau_c < \infty) = 1. \tag{0.3}$$

*Moreover, if $K$ is the one-step copy kernel defined below and $\pi$ is its normalized stationary distribution, then for any variant $a \in \mathcal{A}$,*

$$\Pr\left(\text{eventual consensus on variant } a\right) = \sum_{v : X_v(0) = a} \pi_v. \tag{0.4}$$

*Finally, if the lexicon contains finitely many cells, then all cells reach consensus almost surely under the same analytic assumptions.*

We now prove the theorem through a sequence of lemmas.

**Lemma 0.2.** *The probability that at least one accepted token for cell $c$ arrives from a single neighbor in one step is*

$$\eta_c = 1 - (1 - q_c p)^r. \tag{0.5}$$

*In particular, if $q_c \in (0,1)$, $p \in (0,1)$, and $r \geqslant 1$, then $0 < \eta_c < 1$.*

**Proof.** Recall that each agent samples cell $c$ with probability $q_c$ when generating an utterance. It does so $r$ times. The receiving agent accepts a heard token with Bernoulli probability $p$. Thus the probability that a single generated utterance is both cell-relevant and accepted is $q_c p$. Regard the events of successful transmission as independent. Then the number of successful transmissions from a single neighbor is a binomial random variable

$$Y \sim \text{Binomial}(r, q_c p).$$

Therefore we have

$$\text{Pr}(Y \geqslant 1) = 1 - \text{Pr}(Y = 0) = 1 - (1 - q_c p)^r.$$

Since $q_c p \in (0,1)$ and $r \geqslant 1$, the resulting value lies strictly between 0 and 1.

**Lemma 0.3.** *The probability that at least one neighbor succeeds in producing at least one accepted token for $v$ during one step is*

$$\alpha_v = 1 - (1 - \eta_c)^{d(v)}. \tag{0.6}$$

**Proof.** This follows by the same argument as the previous lemma, now applied across the $d(v)$ neighbors of $v$. Each neighbor independently succeeds in producing at least one accepted token for cell $c$ with probability $\eta_c$. Thus the probability that no neighbor succeeds is $(1 - \eta_c)^{d(v)}$, and so the probability that at least one neighbour succeeds is

$$1 - (1 - \eta_c)^{d(v)}.$$

Fix $v$ and a neighbor order

$$u_1 \prec u_2 \prec \cdots \prec u_{d(v)}.$$

Introduce the indicator function $\delta_{x,y}$, which is 1 when $x = y$ and 0 otherwise.

**Lemma 0.4.** *The probability that node $v$ at time step $t+1$ copies the value held by node $u$ at time $t$ is given by*

$$\textit{Pr}\,(v \textit{ copies } u \textit{ in one step}) = K_{v \to u}, \tag{0.7}$$

*where $K_{v \to u}$ is the one-step copy kernel at node $v$ given by*

$$K_{v \to u} := \begin{cases} (1 - \eta_c)^{d(v)}, & u = v, \\ \eta_c (1 - \eta_c)^{d(v) - j}, & u = u_j \in N(v), \\ 0, & \textit{otherwise.} \end{cases} \tag{0.8}$$

**Proof.** Define $B_j \sim \text{Bernoulli}(\eta_c)$ to be independent indicators that $u_j$ had at least one accepted token for cell $c$ delivered to $v$ in this step. If accepted tokens are processed in the fixed neighbor order, with later accepted tokens overwriting earlier ones, then $u_j$ is the final copied source exactly when $B_j = 1$ and $B_m = 0$ for all $m > j$. Thus

$$\text{Pr}(\text{copy } u_j) = \eta_c (1 - \eta_c)^{d(v) - j}, \quad j = 1, \ldots, d(v). \tag{0.9}$$

If we define $Z = \sum_{j=1}^{d(v)} B_j$, then $Z \sim \text{Binomial}(d(v), \eta_c)$ and we have

$$\text{Pr}(\text{no update}) = \text{Pr}(Z = 0) = (1 - \eta_c)^{d(v)}. \tag{0.10}$$

The no-update event is equivalently the event that $v$ copies its own previous value. Combining these cases gives the stated kernel.

This kernel has three useful properties. These are properties of the auxiliary copy-source chain on graph vertices, not of the full configuration chain on $\mathcal{A}^V$.

- Row-stochasticity. To see this observe that all entries are $\geq 0$ by construction, and that the row-sum is

$$\sum_{u \in V} K_{v \to u} = (1-\eta_c)^{d(v)} + \sum_{j=1}^{d(v)} \eta_c (1-\eta_c)^{d(v)-j} = (1-\eta_c)^{d(v)} + \alpha_v = 1. \tag{0.11}$$

- Irreducibility. On a connected graph and for any $v$ with $d(v) \geqslant 1$ we have

$$\alpha_v = 1 - (1-\eta_c)^{d(v)} > 0.$$

For any neighbor $u_j \in N(v)$ we have

$$K_{v \to u_j} = \eta_c (1-\eta_c)^{d(v)-j} > 0.$$

Therefore every graph edge has a positive transition probability in the auxiliary copy-source chain, and by connectivity every vertex communicates with every other vertex.

- Aperiodicity. A vertex has period

$$\text{per}(v) = \text{gcd}\{t \geqslant 1 : (K^t)_{v \to v} > 0\}.$$

An irreducible finite chain is aperiodic if one, hence all, vertices have period 1. To see this note that

$$K_{v \to v} = (1-\eta_c)^{d(v)}. \tag{0.12}$$

Since $0 < \eta_c < 1$, we have $K_{v \to v} > 0$. So every vertex has a self-loop, making the auxiliary chain aperiodic.

Then via standard results in finite-state Markov chain theory one can conclude that $K$ has a unique normalized stationary distribution $\pi$ satisfying

$$\pi^T K = \pi^T, \qquad \sum_{v \in V} \pi_v = 1, \qquad \pi_v \geq 0. \tag{0.13}$$

The full cell process on $\mathcal{A}^V$ should not be confused with this auxiliary copy-source chain. The full cell process is not irreducible, since consensus configurations are absorbing. The auxiliary chain $K$ is introduced only to describe the one-step source from which each node copies.

**Lemma 0.5.** *Assume $0 < \eta_c < 1$. For the finite connected graph $G = (V, E)$, the cell process reaches consensus almost surely. That is, if*

$$\tau_c = \text{inf}\{t \geq 0 : X_v(t) = X_w(t) \textit{ for all } v, w \in V\}, \tag{0.14}$$

*then*

$$\text{Pr}(\tau_c < \infty) = 1. \tag{0.15}$$

**Proof.** If $n = 1$, the claim is immediate. Assume therefore that $n \geqslant 2$.

Consider any non-consensus state. Choose a variant $a$ present in the population, and choose a root node $v_0$ with $X_{v_0} = a$. Since $G$ is connected, choose a spanning tree rooted at $v_0$, and order the remaining vertices

$$v_1, \ldots, v_{n-1}$$

so that each $v_i$ has a parent $p_i$ closer to the root and all parents precede their children in the ordering.

There is a positive-probability path to consensus on $a$. At step $i$, let $v_i$ copy its parent $p_i$, while every other vertex self-copies. This event has positive probability because $K_{v_i \to p_i} > 0$ for graph edges and $K_{w \to w} > 0$ for all vertices when $0 < \eta_c < 1$. The joint scheduler assumption stated above ensures that this finite combination of allowed copy and self-copy choices has positive probability. After at most $n-1$ steps, every vertex holds $a$.

Thus from every non-consensus state there is a positive-probability path to a consensus state. Since the state space $\mathcal{A}^V$ is finite, there is a uniform $\varepsilon > 0$ such that, from any non-consensus state, consensus is reached within at most $n-1$ steps with probability at least $\varepsilon$. Therefore, by the Markov property,

$$\Pr(\tau_c > m(n-1)) \leq (1-\varepsilon)^m \to 0 \tag{0.16}$$

as $m \to \infty$. Hence $\Pr(\tau_c < \infty) = 1$.

**Lemma 0.6.** *Define, for any variant $a \in \mathcal{A}$, the stationary weighted mass at time $t$ by*

$$M_a(t) = \sum_{v \in V} \pi_v \delta_{X_v(t),a}. \tag{0.17}$$

*Then $\{M_a(t)\}_{t \geq 0}$ is a martingale:*

$$\mathbb{E}[M_a(t+1)|\mathbf{X}(t)] = M_a(t). \tag{0.18}$$

**Proof.** Conditioning on $\mathbf{X}(t)$, the current states are fixed. Hence

$$\begin{aligned} \mathbb{E}[1\{X_v(t+1)=a\}|\mathbf{X}(t)] &= \Pr\left(X_v(t+1)=a|\mathbf{X}(t)\right) &(0.19)\\ &= \sum_{u \in V} K_{v \to u} \delta_{X_u(t),a}. &(0.20) \end{aligned}$$

This implies

$$\begin{aligned} \mathbb{E}[M_a(t+1)|\mathbf{X}(t)] &= \sum_{v \in V} \pi_v \sum_{u \in V} K_{v \to u} \delta_{X_u(t),a} &(0.21)\\ &= \sum_{u \in V} \left( \sum_{v \in V} \pi_v K_{v \to u} \right) \delta_{X_u(t),a} &(0.22)\\ &= \sum_{u \in V} \pi_u \delta_{X_u(t),a} &(0.23)\\ &= M_a(t), &(0.24) \end{aligned}$$

where the third equality uses $\pi^T K = \pi^T$. Therefore $\{M_a(t)\}_{t \geq 0}$ is a martingale.

By the convergence argument above, with probability one there is a finite time $\tau_c$ at which all agents have the same form for the cell. Since $\pi$ is normalized, at this time the weighted mass of variant $a$ is either zero or one:

$$M_a(\tau_c) \in \{0, 1\}.$$

The martingale is bounded, since $0 \leq M_a(t) \leq 1$ for all $t$. Thus the stopped process $M_a(t \wedge \tau_c)$ is bounded, and optional stopping together with bounded convergence gives

$$\mathbb{E}[M_a(\tau_c)] = M_a(0). \tag{0.25}$$

Therefore

$$\begin{aligned} \Pr\left(\text{consensus on variant } a\right) &= M_a(0) &(0.26)\\ &= \sum_{v: X_v(0)=a} \pi_v. &(0.27) \end{aligned}$$

It remains only to pass from a single cell to the finite lexicon. Let

$$\mathcal{C} = \Lambda \times \Sigma$$

be the finite set of lexeme-slot cells under consideration. For each cell $c \in \mathcal{C}$, let $\tau_c$ be the corresponding consensus time. The preceding argument gives

$$\Pr(\tau_c < \infty) = 1$$

for each fixed $c$. Since $\mathcal{C}$ is finite,

$$\Pr\left(\max_{c\in\mathcal{C}} \tau_c < \infty\right) = 1. \tag{0.28}$$

Thus, in the analytic regime described above, all cells reach consensus almost surely. This proves the theorem.

## Empirical stem allomorphy results

In this section we include several diagrams illustrating the output of our morphological segmentation pipeline described in the main body of the paper. These are presented in Fig 22. Note that, while we label the Estonian and Finnish plots as ”rule-based”, they are in fact not the result of our pipeline but the result of an independent segmentation by linguists already provided in the datasets used.

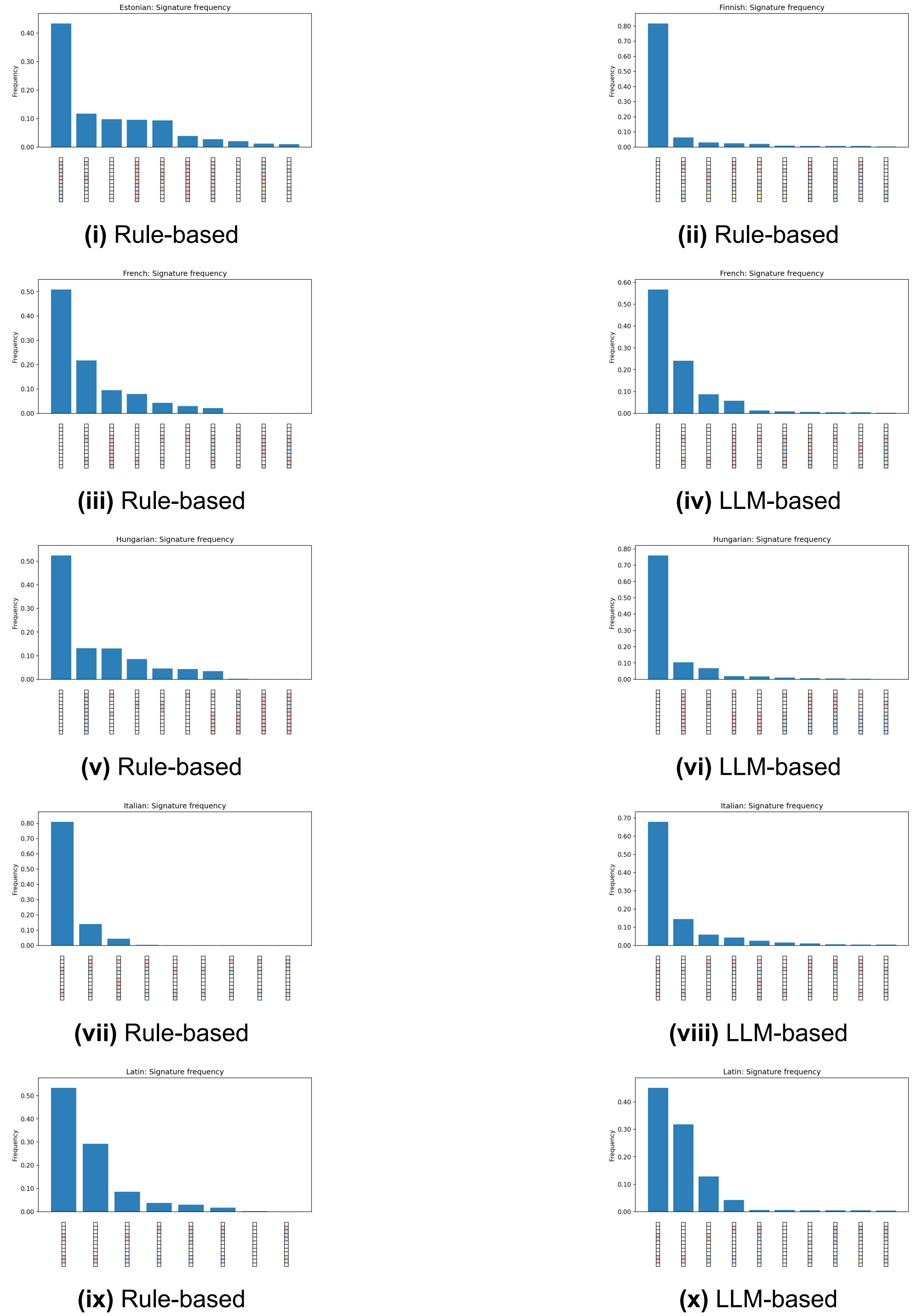

**Figure 22.** Signature histogram figures for the rule-based and LLM-based segmentation outputs.

## Case studies

In the following three sections we present simulations of three actual historical changes, using the model developed in this paper, with a few additional mechanisms for adoption. The first involves a simple case of affix allomorphy, namely the nominative marker in Korean. The second involves stem alternations in Romance languages, which we have already seen examples of in the main body of the paper. Finally, the third involves a change of a rather different kind, namely the morphophonological phenomenon of initial consonant mutation in Celtic languages.

In the latter two cases, namely Romance and Celtic, we also show how we can model the differential development of the system in two branches of the family after population split.

### The Korean nominative marker

In Korean, nominal case is indicated by suffixes attached to the noun phrase head. In the case of the subject marker, there are two allomorphs, *-i* and *-ka*, the former used to mark nouns that end in consonants, and the latter which mark nouns that end in vowels.

In Old and Middle Korean, however, there was only one marker, namely *-i*, which attached to nouns of any shape, including those that ended in the vowel *-i*. In the case of nouns ending in vowels, the /i/ was reduced to a glide /y/ and in case the noun ended in /i/, the /i/ was assimilated. Middle Korean was a pitch accent language and the *-i* suffix caused a low-tone final vowel to become rising in pitch. See [230, p. 187] for details, and [231] for discussion of Old Korean.

The earliest attestation of the *-ka* marker has been claimed to be from 1572 (see again [230]), but this single example is disputed [232], and in any case no other instance of *-ka* appears in the written record until nearly a century later in the Early Modern Korean period [232, p. 16]. As [232] notes, the first attestations of *-ka* occur after nouns ending in /i/, suggesting that the original motivation for *-ka* was to disambiguate cases where the *-i* marker would have been assimilated to the preceding vowel. [232] gives the following example from a letter from Queen Inseon (1619–1674) to her daughter concerning the health of her grandchildren. Note that while the initial segment of *-ka* is underlyingly voiceless /k/, in Revised Romanization it is transcribed as ⟨g⟩, and is in fact phonetically voiced between vowels, hence the transcription as *-ga* in the example below:

| 원샹이도 | 올라오고 | … | 원샹이<u>가</u> | 코 | 흘리고 |
|---|---|---|---|---|---|
| wonsyangi-do | ollaogo | … | wonsyangi-<u>ga</u> | ko | heulligo |
| Wonsyang-i-also | coming-up | … | Wonsyang-i-subj | nose | running |

‘Little Wonsyang also came over …He had a runny nose’

There are various theories of the origin of the *-ka* marker, including it being a native formation, or borrowed from the Japanese nominative marker *-ka*—there were at least some Japanese speakers in the Korean peninsula after the Imjin War (1592–1598). The issues around the latter theory, and in particular the question of whether grammatical particles are likely to be borrowed, are discussed in [232]. In any case, whatever the source, it seems clear at least that *-ka* was introduced initially as a way of more overtly marking nominative case in nouns ending in /i/, and then spread to attach to all nouns ending in vowels.

We simulate the development of the *-ka* marker as follows. We use a corpus of 399 Middle Korean noun forms extracted from Wiktionary. While we do not have frequency information for these nouns, we used the proxy of ordering them in descending order by the frequency of their English translation in the British National Corpus [233]. (See `https://www.kilgarriff.co.uk/bnc-readme.html` for the frequency lists.) To this was applied a Zipf-Mandelbrot frequency curve, which we have used in all our simulations. The system is initialized with nominative forms ending solely in /-i/, which reduces to /y/ after vowels, and assimilates entirely to the preceding stem if it ends in /i/, as discussed above. For a subset of the speakers, /-ka/ is introduced to mark the nominative of nouns ending in /-i/. The agents then adopt a decision-tree compression scheme whereby the decision categories are /-i/ or /-ka/, and the features of the environment are /i/-Final, Vowel-Final and Consonant-Final. Later on, these features are relaxed to just Vowel-Final and Consonant-Final. Fig 23 shows one run of the simulation, with a novelty parameter setting of 0.6. Fig 24 shows decision trees at two stages, the first from shortly after the decision-tree compression is first introduced, and the second from shortly after the features are relaxed.

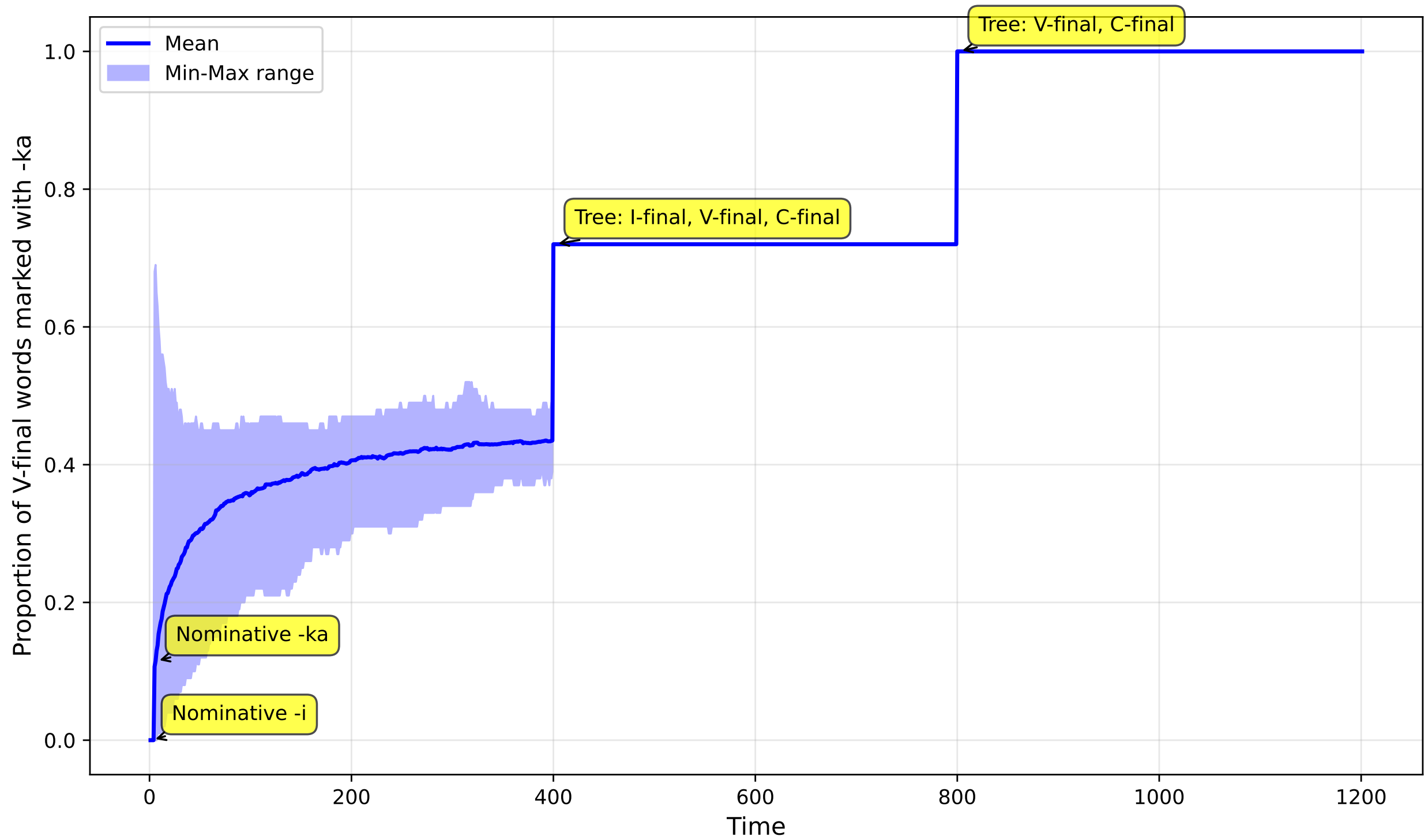

**Figure 23.** A run of the simulation of the change from the nominative marker *-i* to *-ka* after vowels in early Modern Korean. The population has 50 agents, exchanging 400 utterances at each tick. The probability of adoption is set to 0.1, and the probability of the initial change from *-i* to *-ka* after nouns ending in /i/ is also 0.1. The novelty parameter is 0.6.

In Fig 25 we show the simulation with a lower novelty setting of 0.2, where speakers are much less inclined to adopt new forms. This presents another possible history, one where the *-ka* innovation was introduced, but quickly lost favor and died out.

01_high_front

HIGH_FRONT_FINAL <= 0.5
log_loss = 0.875
samples = 397
value = [280, 117]
class = +I

log_loss = 0.0
samples = 280
value = [280, 0]
class = +I

log_loss = 0.0
samples = 117
value = [0, 117]
class = +KA

02_any_vowel

CONSONANT_FINAL <= 0.5
log_loss = 0.977
samples = 397
value = [234, 163]
class = +I

log_loss = 0.0
samples = 163
value = [0, 163]
class = +KA

log_loss = 0.0
samples = 234
value = [234, 0]
class = +I

**Figure 24.** Decision trees for an agent at step 420 (left) and 820 (right) for the simulation presented in Fig 23.

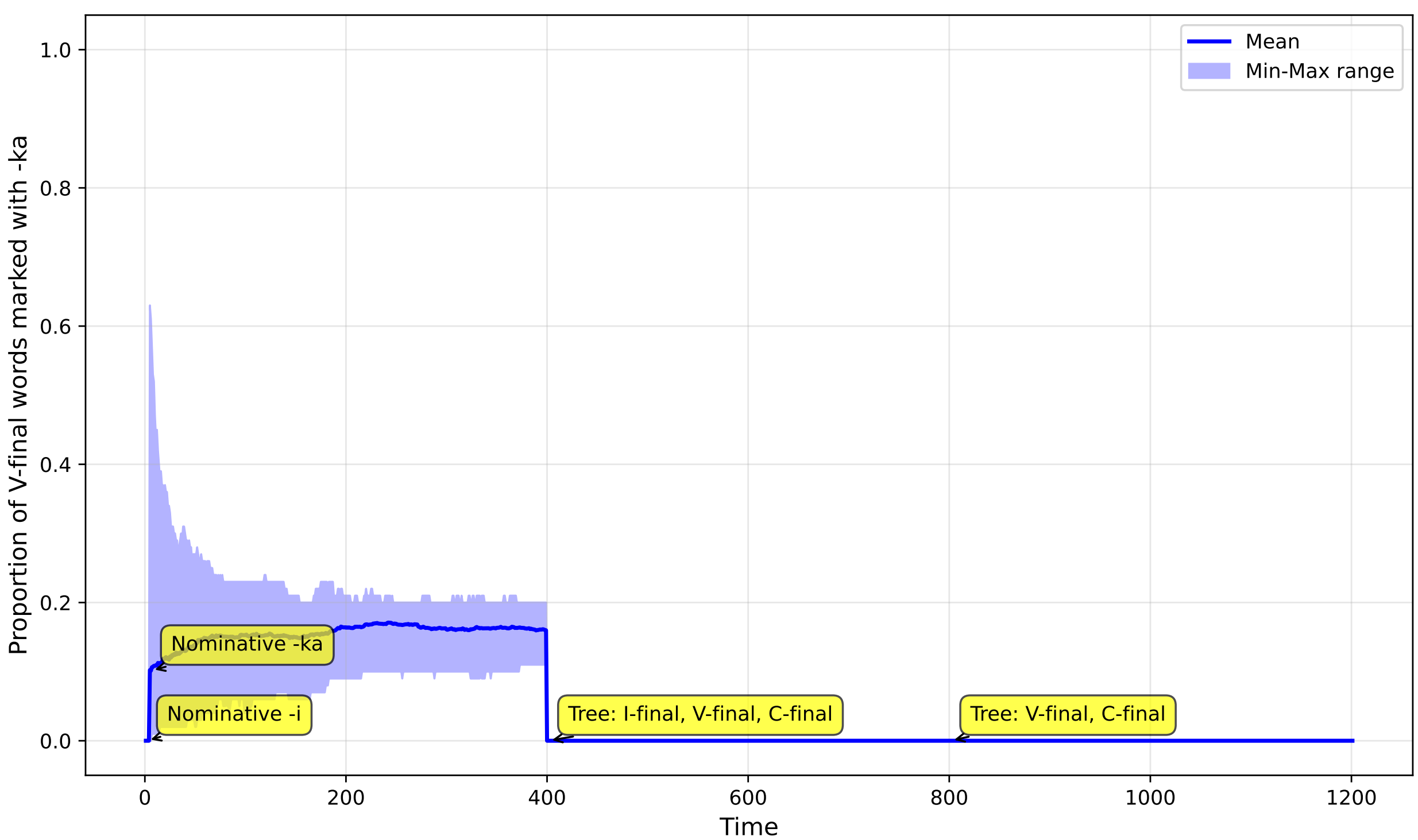

**Figure 25.** A run of the simulation of the change from the nominative marker *-i* to *-ka* after vowels in early Modern Korean. The population has 50 agents, exchanging 400 utterances at each tick. The probability of adoption is set to 0.1, and the probability of the initial change from *-i* to *-ka* after nouns ending in /i/ is also 0.1. Here the novelty parameter is 0.2, and rather than spreading the system collapses.

## Romance verbal stem alternants

Stem alternations in Romance languages have been widely studied. They figure prominently in Carstairs-McCarthy's treatment of the evolution of morphology [198], and have been studied in depth in [7–11, 227]. There are various patterns of stem alternation attested in a large proportion of Romance languages, but in this discussion we will consider one of the more prominent ones, namely the so-called 'N-pattern', illustrated Table 11 for the Occitan verb *portar* 'carry'. The pattern was initially instigated by stress alternation in Vulgar Latin, so that the gray cells in the table have stress on the stem, and the remainder show stress on the suffix. In Occitan this resulted in vowel alternations, in the case at hand /ɔ/ when the stress is on the stem, and /o/ otherwise.

**Table 11.** The N-pattern as exemplified in Occitan, for the verb *portar* 'carry', from [9], Table 1. The vowel written ⟨o⟩ is /o/, whereas the one written ⟨ò⟩ is /ɔ/. Abbreviations: PRS = 'present'; IND = 'indicative'; IPFV = 'imperfective'; SBJV = 'subjunctive'; PFV = 'perfective'; PST = 'past'; FUT = 'future'; COND = 'conditional'.

| | | PRS.IND | IPFV.IND | PRS.SBJV | PFV.PST.IND | IPFV.SBJV | FUT | COND |
|---|---|---|---|---|---|---|---|---|
| | 1 | pòrti | portavi | pòrti | portèri | portèssi | portarai | portariá |
| SG | 2 | pòrtas | portavas | pòrtes | portères | portèsses | portaràs | portariás |
| | 3 | pòrta | portava | pòrte | portèt | portèsse | portarà | portariá |
| | 1 | portam | portàvem | portem | portèrem | portèssem | portarem | portariam |
| PL | 2 | portatz | portàvetz | portetz | portèretz | portèssetz | portaretz | portariatz |
| | 3 | pòrtan | portavan | pòrten | portèron | portèsson | portaràn | portarián |

As can be seen in the table, the slots that exhibit the so-called N-pattern are the minority of slots in the table, corresponding to the fact that in early Romance most of the verb forms had stress on the suffix.

A similar pattern is found for the verb *contar* in Spanish (Table 12), where here and henceforth we just present the present and imperfective indicative forms.

**Table 12.** Present and imperfective indicative forms of the Spanish verb *contar* 'speak'.

| | | Present | Imperfect |
|---|---|---|---|
| Sg. | 1 | cuento | contaba |
| | 2 | cuentas | contabas |
| | 3 | cuenta | contaba |
| Pl. | 1 | contamos | contábamos |
| | 2 | contáis | contabais |
| | 3 | cuentan | contaban |

The N-pattern not only engendered changes in the stem form, but was also the attractor for partial and full suppletion. Partial suppletion is exemplified by the distribution of the *-isc* (originally an inchoative suffix in Latin) augment in various verbs in the *-ir(e)* inflection class. Full suppletion is exemplified by the *ambulāre/vadere* alternation for the verb 'go', which manifests itself in various ways in multiple Romance languages—though Spanish and Portuguese notably do not show these alternations. These are illustrated for the present and imperfective indicative for Italian in Table 13 for the verbs *andare* 'go' and *finire* 'finish'. Here one sees the same pattern in the present and imperfective indicative as in Occitan, but rather than mere stress and vowel quality distinctions, partially or completely different stems occupy the shaded cells.

**Table 13.** Present and imperfective indicative forms of the Italian verbs *andare* 'go' and *finire* 'finish'.

| | | Present | Imperfect | Present | Imperfect |
|---|---|---|---|---|---|
| Sg. | 1 | vado | andavo | finisco | finivo |
| | 2 | vai | andavi | finisci | finivi |
| | 3 | va | andava | finisce | finiva |
| Pl. | 1 | andiamo | andavamo | finiamo | finivamo |
| | 2 | andate | andavate | finite | finivate |
| | 3 | vanno | andavano | finiscono | finivano |

As [9] documents, the loss of lexical stress in French, which occurred prior to the Old French period (approximately mid 8th Century), decimated this pattern in French, though many residues of the earlier alternation remained, in particular the *ambulāre/vadere* alternation, manifest in the verb *aller* 'go'. In Table 14, one sees the N-pattern retained in *aler* 'go', but lost entirely in *fenir* 'finish', where the *-isc* augment forms, manifest with an *-s* stem augment in French, have taken over the entire paradigm. This suggests in turn that the *ambulāre/vadere* must have stabilized fairly early, whereas the *-isc* augment was later, and had not stabilized when French started losing its lexical stress.

**Table 14.** Present and imperfective indicative forms of the Medieval French verbs *aler* 'go' and *fenir* 'finish'.

| | | Present | Imperfect | Present | Imperfect |
|---|---|---|---|---|---|
| Sg. | 1 | vays | alois | fenis | fenissoie |
| | 2 | vas | alois | fenis | fenissoies |
| | 3 | va | aloit | fenist | fenissoit |
| Pl. | 1 | alons | alions | fenissons | fenissiions |
| | 2 | alez | aliez | fenissiez | fenissiiez |
| | 3 | vont | aloient | fenissent | fenissoient |

In our simulation we focus on modeling the history of two consequences of the N-pattern, namely the *ambulāre/vadere* alternation, and the *-isc* augmentation alternation. We model the introduction of the N-pattern as a consequence of the stress alternation pattern introduced in Vulgar Latin discussed above. We assume a set of 278 Latin verbs, inflected in the present and imperfective indicative (12 slots in total). The stress alternation is introduced by rule. At a later step, a subset of the agents replace the verb *ambulāre* with forms of *vadere* as the verb 'go' for all paradigm slots. Later on, 10 third conjugation (*i*-stem) verbs are augmented with *-isk* in all paradigm slots, again for a subset of the verbs. (Both of these changes are implemented as rewrite rules in the simulation.)

We assume the *stem-is-target* setting so that when a form is replaced by borrowing, all slots that share the same stem form get replaced by the new form. We also assume that agents will be biased towards borrowing a form if it is minimally disruptive to their current morphology in that it would involve replacing a minority of the stem forms, and similarly biased against borrowing where a majority would be replaced. This is modeled by scaling the probability of an agent adopting a form by

$$\sigma = \frac{1}{1 + e^{-p'}}$$

where $p' = \lambda(p - 0.5)$, with scaling factor $\lambda$, and $p$ is the proportion of slots in a lexeme's inflection that would be changed if the form were adopted. In the experiments below, $\lambda = 5.0$.

We also assume that speakers can be aware of lexical classes, so that for example verbs that are augmented with *-isc* form a lexical class. If a form is adopted from a member of a class, then for each of the other members of the class, with the same probability, the comparable form may be adopted. So for example if the agent decides to adopt the *-isc* augment for the stressed-stem forms of *impartire*, with the same probability the agent also adopts the *-isc* augment for the comparable slots for *farcire*, for *florire*, etc.

The steps in the simulation are as follows:

- 278 Latin verb lemmata are arranged in descending relative frequency order. As with the Korean simulation, the frequencies are estimated on the basis of the frequency of the English translation in the British National Corpus.
- Verbs are inflected for the six person-number combinations in the present and imperfective indicative.
- Endings are mapped by rule to a more 'Romance-like' set of endings, and stress assigned to the endings in all except the 1sg, 2sg, 3sg, 3pl present, where stress is on the stem.
- Simulations involve 40 agents, each agent exchanging 100 messages on each tick, with the simulation run for 500 ticks. The network topology is scale-free. (Note that a complete network topology, unrealistic anyway for a population, resulted in almost immediate elimination of the verbal variants introduced below.) Adoption is Bernoulli, with probability of adoption 0.6 and novelty weighting 0.6. The *stem-as-target* setting is used, and no compression.

- Agents may use lexical classes in their adoption, as described above.
- Agents die after $t$ ticks, with the mentorship mode being set to *random*.
- The initial verb for 'go' is based on *ambulāre*, and the ten *īre* verbs that will undergo *-isc* augmentation start out without the *-isc* augment.
- For a proportion $p$ of the agents, forms of *vadere* replace all forms of *ambulāre* at step 6.
- A random split into two dialects—'French' and 'Italian'—with 20 agents each occurs at tick 150.
- Destressing occurs in the 'French' branch at some point after the split.
- At some point either before or after the split, *-isc* augment is introduced for all forms of ten verbs for a proportion $p$ of agents.

The desired outcome of the simulation is one that mimics what happened in real Italian and French, namely:

- 'go' shows the N-pattern for both languages.
- *-isc* augment verbs show the N-pattern for Italian; for French all forms in the paradigm end up with the *-isc* augment.

The configuration that came closest to this pattern is as in Table 15. This resulted in the evolution patterns shown in Figures 26–28. Here, both branches have the N-pattern for 'go', and correct patterns—N-pattern for Italian, and all-*isc* for French, in 6/10 verbs. The remaining four verbs show either:

- Elimination of the *-isc* ending entirely in both branches. Two cases: *impartīre* and *largīre*, Fig 27.
- Correct augmentation in French, but elimination of *-isc* in Italian: *ērudīre*, Fig 27.
- Correct augmentation in French, and in Italian equal numbers of total elimination of *-isc*, or an 'anti N-pattern' where the *-isc* augment is found in the first and second person plural, and in the imperfective: *grunnīre*, Fig 27.

**Table 15.** Best settings for the Romance simulation.

| Seed | 831 |
|---|---|
| Probability of change $p$ | 0.08 |
| Use of lexical classes | True |
| Introduction of *-isc* | Tick 140 |
| Destressing in French | 50 ticks after split |
| Ticks per generation | 10 |

What the above simulation shows is that it is possible for a small number of agents (about 8%) to introduce a change to all forms of the verb, and have that borrowed into one stem alternant for other agents, with the final suppletive paradigm eventually spreading to the entire population. The elimination of lexical stress in the 'French' branch eliminates the possibility of the Italian-style *-isc* alternation. However, under the assumption that the *ambulāre*/*vadere* alternation was introduced earlier than the *-isc* alternation, the former was already entrenched by the time the dialect split happened and before the loss of stress could remove the target pattern for the N-pattern alternations. In this simulation, the 'death' of agents is critical: For example, the older speakers who have *vadere* forms in all positions need to die out so that the stem-alternate pattern can take over.

How sensitive is the simulation to the particular settings chosen? With the settings in Table 15, 16 out of 22 of the righthand two panels showed the historically correct outcome. Changing the probability with which the change was introduced yielded the following variation:

| Probability | # Correct panels |
|---|---|
| 0.06 | 4 |
| 0.08 | 16 |
| 0.1 | 8 |
| 0.12 | 6 |

Disabling use of lexical classes reduced the count to 7. Introducing *-isc* too early before the dialect split resulted in lower numbers of expected patterns, as did introducing it after the split (below the mid rule line in the table below):

| *-isc* introduction | # Correct panels |
|---|---|
| 120 | 7 |
| 130 | 7 |
| 140 | 16 |
| 160 | 8 |
| 170 | 6 |

Where destressing occurs in the French branch also matters; here we show the point at which destressing occurs in terms of the number of ticks after the dialect split:

| Where destressing occurs | # Correct panels |
|---|---|
| 10 | 11 |
| 20 | 12 |
| 30 | 13 |
| 50 | 16 |
| 60 | 12 |
| 70 | 13 |
| 80 | 14 |

Finally, the number of ticks per generation also matters:

| Ticks per generation | # Correct panels |
|---|---|
| 5 | 4 |
| 10 | 16 |
| 15 | 5 |
| 20 | 9 |

The outcome is highly dependent on the initial seed meaning that random events in the history can yield very different outcomes. Out of a sample of 114 seeds, 31 (27%) of the seeds resulted in the correct *ambulāre*/*vadere* alternation appearing in both branches of the family, and 37 (32%) in one or more branches. For the *isc*-augment verbs, the following table shows the numbers of verbs that show the correct alternation patterns—all *-isc* for French, N-pattern for Italian—for both branches:

| # Verbs with correct patterns | # Seeds | Percentage |
|---|---|---|
| 1 or more | 70 | 61% |
| 2 or more | 37 | 32% |
| 3 or more | 11 | 9.6% |
| 4 or more | 4 | 4.4% |
| 5 or more | 1 | 0.8% |

In summary, what do we need to assume in order to derive the pattern observed in French and Italian? First of all, we need to assume a phonological environment that introduces stem alternations: this is provided by the stress alternation, as argued in previous work [7–9, 227]. Second, we need to assume that a small percentage of agents introduce a novel stem pattern in all slots for the target verb lexemes. Third, we need to assume that when an agent borrows a form, it replaces that stem in all slots where it occurs (*stem-as-target*), and that borrowing is favored when it replaces fewer slots in the paradigm: note that without the *stem-as-target* setting, the system does not converge on the historically correct patterns. Finally, we need to assume agent death, so that the patterns where the novel form is used in all slots—e.g. all forms of ‘go’ have the *vadere* stem—will be likely to die out. In addition, we needed to assume that the stem alternation for ‘go’ evolved early compared to the *-isc* augment, since the latter is affected by the lexical stress loss innovation in French, but the former is not. Over and above these considerations, the system is sensitive to initial conditions—as expected since had history been different, the innovations that were fossilized in many Romance languages might instead have died out.

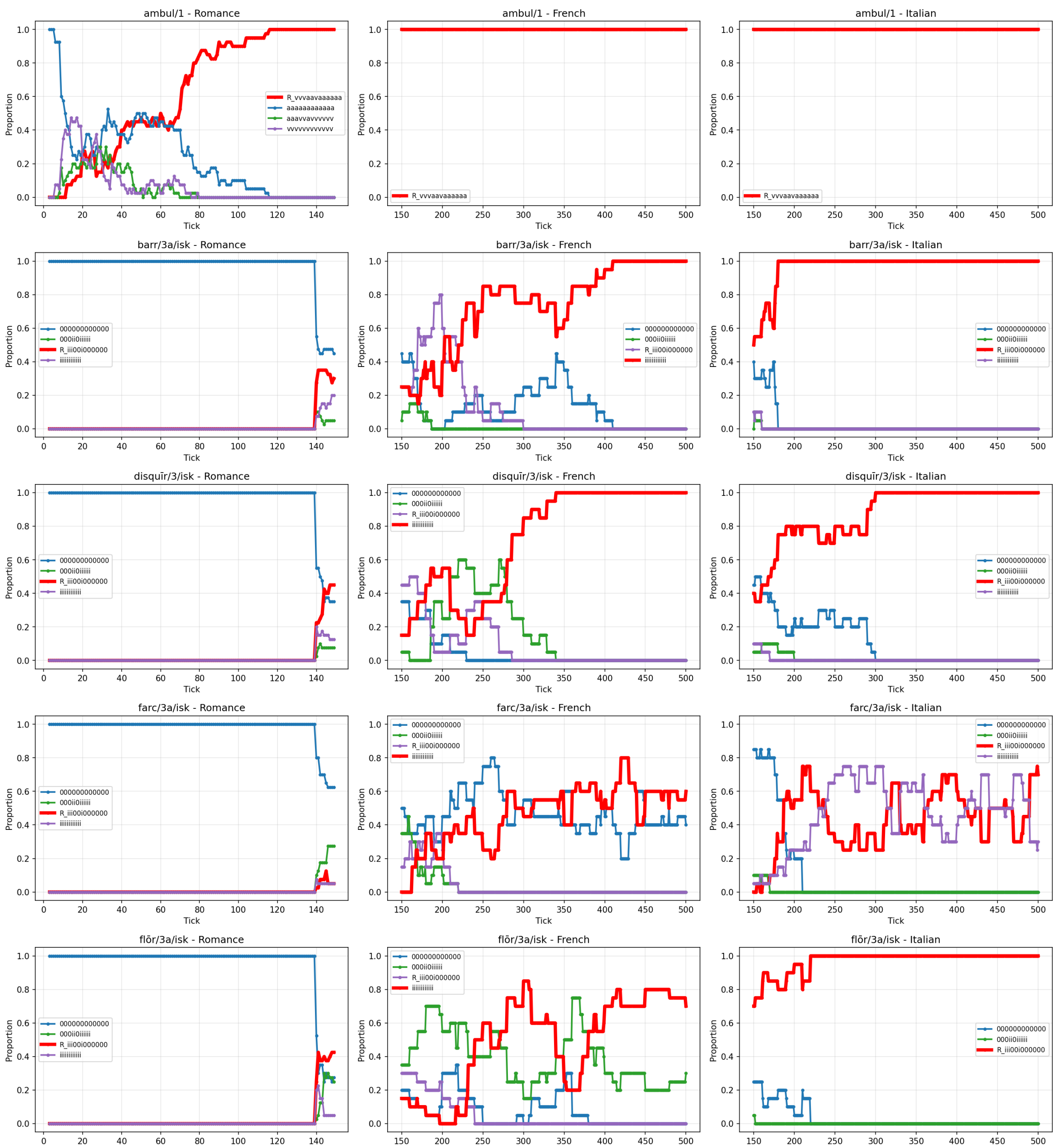

**Figure 26.** Evolution of forms of *ambulāre* ‘go’, *barrıre* ‘trumpet’, *disquırıre* ‘discourse’, *farcıre* ‘force’, *flōrıre* ‘flower’. Each row corresponds to the evolution of the prevalence of a particular pattern for the verb. The left panel represents the period prior to the dialect split, and the right two panels represent the ‘French’ and ‘Italian’ branches respectively. The curve corresponding to the historically correct form is shown in red, so correct plots always have the red winning out. The labels for the curves represent the distributions of the stem alternants. Thus for *ambulāre* `vvvaavaaaaaa` represents the distribution of *vadere* and *ambulāre* forms for the historically attested outcome; `aaaaaaaaaaaa` is all *ambulāre* forms, `vvvvvvvvvvvv` all *vadere* forms, as in, e.g., Spanish. For *-isc* augment forms the letter `i` represents a stem with an *-isc* augment, and `0` the bare stem. The prefix `R_` tags the N-pattern alternation common to many Romance languages and is the expected pattern for all the verbs in the Italian branch. In French, for *-isc* verbs, the expected outcome is `iiiiiiiiiiii`. Settings are as in Table 15.

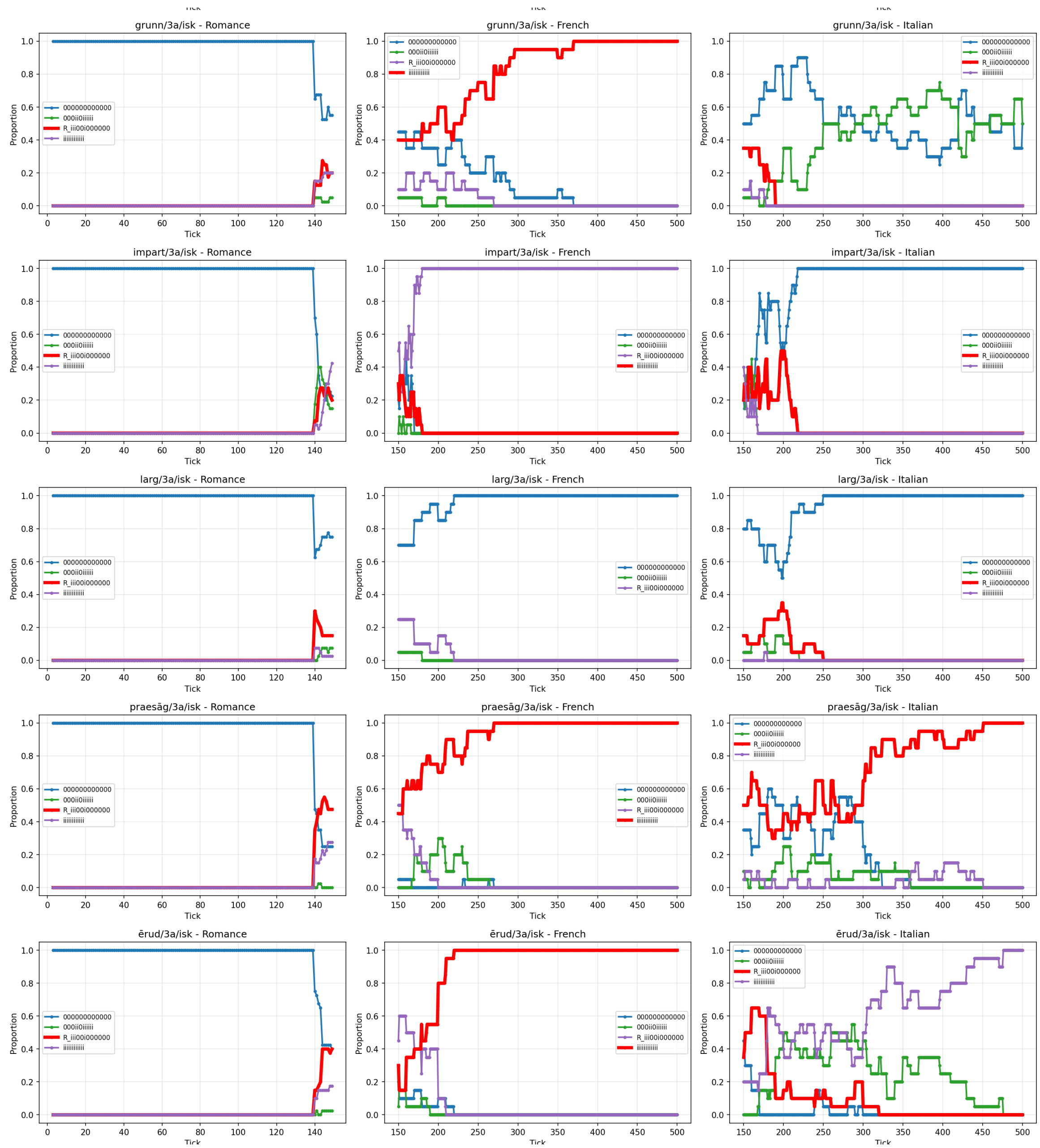

**Figure 27.** Continuation of Fig 26 for verbs *grunnıre* ‘grunt’, *impartıre* ‘give’, *largıre* ‘give generously’, *praesāgıre* ‘presage’, *ērudıre* ‘educate’.

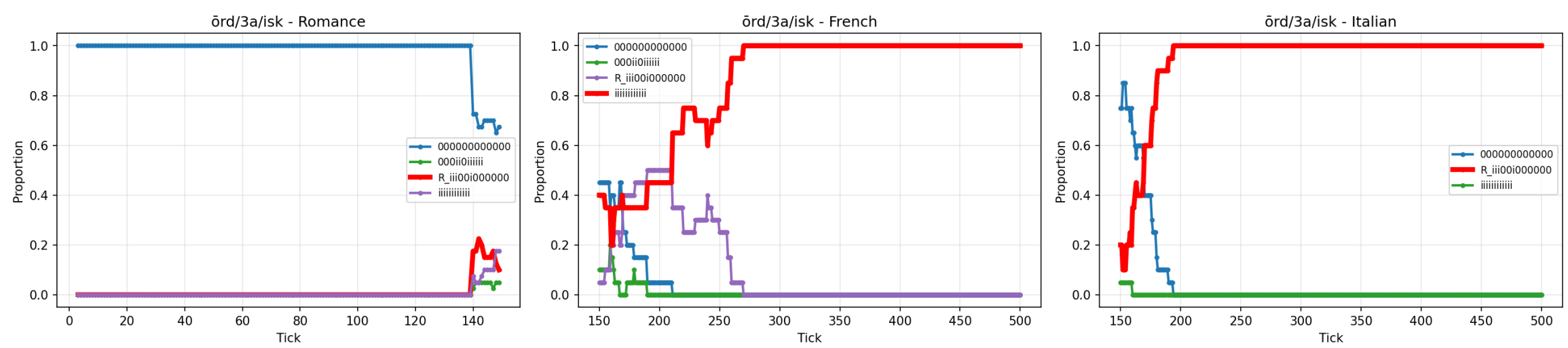

**Figure 28.** Continuation of Fig 27 for the verb *ōrdıre* ‘plot, hatch’.

## Lenition in Celtic

Consonant mutation is a morphophonological change whereby (usually) initial consonants are changed (‘mutated’) in various morphological or syntactic environments. Archetypical instances of consonant mutation are the systems found in all modern Celtic languages, but similar phenomena are found in many language families scattered across the world. See [234] and in particular [235] for a survey of mutation systems in the world’s languages.

Some examples of consonant mutation in Welsh and Irish, two modern Insular Celtic languages, can be found in Table 16. Illustrated are lenition (in Welsh termed the ‘soft mutation’), nasalization (in Irish termed

**Table 16.** Examples of consonant mutation in Welsh and Irish. Shown are various forms of the word ‘cat’, in its original unmutated form in the two languages; lenited after the word for ‘his’; nasalized after the word for ‘my’ in Welsh and ‘their’ in Irish; and spirantized after the word for ‘her’ in Welsh. Note that lenition is referred to as ‘soft mutation’ in Welsh, and nasalization as ‘eclipsis’ in Irish. The forms are shown in the standard Welsh and Irish orthography. The phonetic value of ⟨ch⟩ in both languages is /x/. ⟨gc⟩ in Irish is /g/. ⟨ngh⟩ in Welsh is /ŋ̥/.

| | **Welsh** | **Irish** |
|---|---|---|
| Unmutated | *cath*<br>‘cat’ | *cat*<br>‘cat’ |
| Lenition | *ei gath*<br>‘his cat’ | *a chat*<br>‘his cat’ |
| Nasalization | *fy nghath*<br>‘my cat’ | *a gcat*<br>‘their cat’ |
| Spirantization | *ei chath*<br>‘her cat’ | —<br>— |

‘eclipsis’), and in the case of Welsh the spirant (or ‘aspirate’) mutation. The examples given all involve mutation after various possessive pronouns, but the mutations are used in a wide variety of constructions in the language. Furthermore, the details of where mutations have diverged significantly across the languages, though there are some commonalities. For example, in both Irish and Welsh the third-person masculine possessive pronoun induces lenition in the following word. (On the other hand, while the first person singular possessive induces nasalization in Welsh, the equivalent pronoun in Irish induces lenition.) As an example of the diversity of the phenomenon, a partial list of mutation environments in Welsh is given below:

| | |
|---|---|
| Lenition | feminine singular nouns after the definite article; adjectives after feminine singular nouns; after *ei* ‘his’; after prepositions including *i* ‘to’, *o* ‘from’; at the beginning of a noun phrase after a major constituent |
| Nasalization | after *fy* ‘my’; after the preposition *yn* ‘in’ |
| Spirantization | after *ei* ‘her’; after the conjunction *â* ‘and’; after the numeral *tri* ‘three’ |

While some of these details—notably the lenition of feminine singular nouns after the definite article, and the lenition of adjectives after feminine singular nouns—are common to all Celtic languages, many of the above details are particular to Welsh. The final listed environment for lenition, namely the lenition of the beginning of a noun phrase that follows a major constituent, is a particular peculiarity of Welsh, though see [236] for discussion of a similar use of lenition in earlier forms of Irish. Each of the other modern Celtic languages has their own particular patterns.

The consonants that are subject to mutation, and the changes they undergo also differ. For Welsh, the following changes are found, in this case shown with phonetic symbols:

| | p | t | k | b | d | g | m | ɬ | r̥ |
|---|---|---|---|---|---|---|---|---|---|
| Lenition | b | d | g | v | ð | Ø | v | l | r |
| Nasalization | m̥ | n̥ | ŋ̥ | m | n | ŋ | | | |
| Spirantization | f | θ | x | | | | | | |

In addition, vowel-initial words in various Celtic languages will also undergo some changes. We will not discuss these changes here. Note that the Ø for the lenition of /g/ indicates that in this case the /g/ deletes: in Early Brythonic (the Insular Celtic branch that includes Welsh, Cornish and Breton) the lenited form of /g/ was /ɣ/. Blanks indicate that the consonant in question does not undergo the given change.

For Irish, the changes are as follows:

| | p | t | k | b | d | g | m | s | f |
|---|---|---|---|---|---|---|---|---|---|
| Lenition | f | h | x | v | ɣ | ɣ | v | h | Ø |
| Nasalization | b | d | g | m | n | ŋ | | | v |

In addition to those listed /l/ and /n/ participate in lenition in that there is an alternation between fortis and lenis pronunciations of /n/. This is not indicated in the orthography. In Old Irish /t/ became /θ/, but this changed into /h/ by the Middle Irish period [237]. Similarly, in Old Irish the lenited form of /d/ was /ð/, which became /ɣ/ by Middle Irish.

For lenition, the most significant difference between Welsh (and other Brythonic languages) and Irish (and the other Goidelic languages, Manx and Scots Gaelic) is with the voiceless stops /p, t, k/. In Brythonic languages these become voiced under lenition, whereas in Goidelic they become voiceless fricatives.

In what follows we will focus on lenition. We note in passing that spirantization in Welsh (and other Brythonic languages) is sometimes characterized as lenition [237–239], but in any case its origin is different from, and later than, the main lenition process.

The latter was originally a process in Proto-Celtic whereby stops, and in particular voiced stops, became spirantized when following a vowel and followed by another vowel or sonorant; in this and the ensuing discussion we follow [238, 240]. By the Proto-Insular Celtic period, the domain of application had broadened so that word-initial consonants in the appropriate phonetic environment also underwent the change. The upshot of this is that, for example, adjectives following a noun ending in a vowel that started with a voiced stop, would be subject to this change. Since many (though not all) feminine singular nouns in Early Celtic ended in a vowel (typically *-a*), adjectives after such nouns would undergo lenition. Similarly, many masculine plural nouns ended in a vowel (typically *-oi*), and thus induced a similar change on following adjectives. Thus far, lenition was phonologically motivated and thus was no different from the very similar lenition of voiced stops after vowels, including across words, in Spanish.

At some point after the split between the Brythonic and Goidelic branches of Insular Celtic, lenition broadened further, this time including voiceless stops. But the mode of change was different in that in Goidelic the voiceless stops spirantized, whereas in Brythonic they became voiced.

Later, around 500 AD, final short vowels were eliminated via apocope [238], thus removing the phonological environment for lenition. However, lenition remained, having been reanalyzed as being triggered by morphosyntactic features, such as Feminine-Singular or Masculine-Plural, or by particular lexical items. At this point it had become a true consonant mutation process, where a phonological change is induced in the beginning of a word due to grammatical, rather than phonological, conditioning environments. Due to this change, lenition spread to cases that would not have been subject to it in earlier forms of Celtic, for example adjectives after feminine nouns that originally ended in consonants. In particular [241] notes (p. 141) that:

> Die Kasus, hinter denen sich Lenierung findet, sind:
>
> 1. …
> 2. der Nsg aller Feminina …
> 3. …, der Npl der maskulinen *o-* und *io*-Stämme …

Anticipating our discussion below, in our simulation, the system mirrors the development in feminine singular nouns, but overgenerates in the case of masculine plurals in that it extends lenition to all masculines. In Modern Irish, lenition of adjectives after masculine plural nominatives is limited to masculine nominative plurals that end in 'slender'—i.e. palatalized—consonants, which correspond to the vowel-stem masculines Thurneysen refers to above. This suggests that a more accurate model would consider both morphosyntactic features as well as phonological features (i.e. the palatalization of the final consonant) in determining whether the noun in question induces mutation.

In the case of the Brythonic branch, feminine singular nouns induced lenition on the following adjective as in Irish. Note that unlike Irish, which retains a nominal case system, case distinctions were lost in Brythonic languages from the medieval period onwards. Lenition was lost in masculine plurals in Welsh, though it was retained in plurals that were semantically dual; cf. [242] who notes for Middle Welsh (p. 34) that "[t]races of old duals with vocalic ending also probably survive in the lenition of the adjective in M[iddle]W[elsh] after nouns preceded by the numeral 'two'," and see also [243]. However, the Southwestern Brythonic languages Cornish [244] and Breton [245] both retain mutation of adjectives following masculine plurals (in Breton, restricted to nouns denoting persons), suggesting that Brythonic paralleled the development in Goidelic despite the loss of masculine plural-induced mutation in Welsh.

While the details just sketched are particular to Celtic, the history parallels the development of mutation systems in other language families. Thus as [235] shows in his detailed account of consonant mutation in African Atlantic languages, mutation systems start out as phonologically motivated changes and develop their morphosyntactic triggers only when the phonological motivation is lost due to sound changes. After the phonological motivation is lost and the trigger becomes morphosyntactic features, the mutation may spread to other environments that would not historically have had the right phonological properties. Furthermore, the mutation processes may continue to evolve in parallel branches of the family, but with differences of detail in terms of the phonological properties of the change, and the conditioning environments.

We now turn to a description of our simulation of the evolution of lenition in Celtic. We start with a set of nouns and adjectives from Proto-Celtic reconstructions retrieved from Wiktionary. For the adjectives, we select ten adjective stems that start with mutable consonants.

For nouns, we construct two sets, one set, with 226 entries, which we will call *balanced*, which contains singular and plural noun forms that contain vowel- and consonant- stems in proportion to the occurrence of those stems in the Wiktionary data. Note that feminine-singular (nominative) vowel-stem nouns predominantly end in *-ā*, and masculine-plural (nominative) vowel-stem nouns predominantly end in *-oi* (the masculine-singular ending in *-os*, thus in a consonant). This set is thus intended to reflect roughly the distribution of vowel- and consonant- stem nouns that would have existed as the lenition system was evolving. This results in 48 consonant-stem forms (singular and plural) and 202 vowel stem forms. The second set of nouns, with 228 entries, which we will call *biased* contains half consonant-stems and half vowel-stems. The purpose of this set is to see how the system might have evolved if Celtic nouns had a higher proportion of consonant stems. One reasonable hypothesis at least when it comes to noun-adjective constructions is that the lenition system would not have survived had it not been for the strong prevalence of vowel-stem nouns in the system. As with the Korean nouns described previously, we ordered the nouns in descending order by the frequency of their English translation in the British National Corpus.

Nouns, along with their grammatical tags—fem-sg, fem-pl, mas-sg, mas-pl—are combined with the adjective stems and an intervening word boundary '#'. Some examples with the feminine noun *tanā* 'time' are given in Table 17.

**Table 17.** Some examples of noun-adjective initial forms with the feminine noun *tanā* 'time'.

| |
|---|
| tan+ā/fem-sg#k$^{w}$els+ tan+ā/fem-sg#men+ tan+ā/fem-sg#garw+ tan+ā/fem-sg#wind+ tan+ā/fem-sg#derw+ tan+ā/fem-sg#sām+ tan+ā/fem-sg#tlāt+ tan+ā/fem-sg#blāw+ tan+ā/fem-sg#koil+ |
| tan+ās/fem-pl tan+ās/fem-pl#k$^{w}$els+ tan+ās/fem-pl#men+ tan+ās/fem-pl#garw+ tan+ās/fem-pl#wind+ tan+ās/fem-pl#derw+ tan+ās/fem-pl#sām+ tan+ās/fem-pl#tlāt+ tan+ās/fem-pl#blāw+ tan+ās/fem-pl#koil+ |

We assume the following changes, which get introduced at successive stages of the simulation. The details of the rules can be found at `https://github.com/SakanaAI/LanguageEvolution/blob/main/celtic/mutation_2.py`. Note that the split into the Brythonic and Goidelic branches occurs at tick 10. Splits are randomly assigned, and of the 50 agents in our simulations 10 were assigned to the Goidelic branch and the remaining 40 to the Brythonic branch. By coincidence this may correspond roughly to the ratio of the population of Britain around 400 AD at the end of the Roman period—between 3 and 4 million [246]—to that of Ireland at the same time—under 1 million: On the latter see `https://forum.paradoxplaza.com/forum/threads/irish-population-history-my-estimates.1182719/`.

Note that the spread of lenition to voiceless consonants and their grammaticalization due to loss of endings occurred around the 5th Century [237, 238]:

**Tick 5.** Proto-Celtic word internal lenition of voiced stops to fricatives.

**Tick 10.** Split into Goidelic and Brythonic branches.

**Tick 11.** Various sound changes to the Goidelic and Brythonic branches: e.g. /$k^w$/→/k/ in Goidelic and /$k^w$/→/p/ in Brythonic.

**Tick 12.** Insular Celtic extension of the rule to apply across words.

**Tick 15.** Extension of lenition to voiceless stops, changing them to fricatives (Goidelic branch)

**Tick 15.** Extension of lenition to voiceless stops, changing them to voiced stops (Brythonic branch)

**Tick 900.** Introduce decision tree compression to learn mutation in the two branches.

**Tick 1000.** Apocope, removing final morphological affixes.

The decision tree compression uses the environmental features vowel_final, consonant_final, FemSg, MasSg, FemPl and MasPl, predicting two classes, lenited and unlenited.

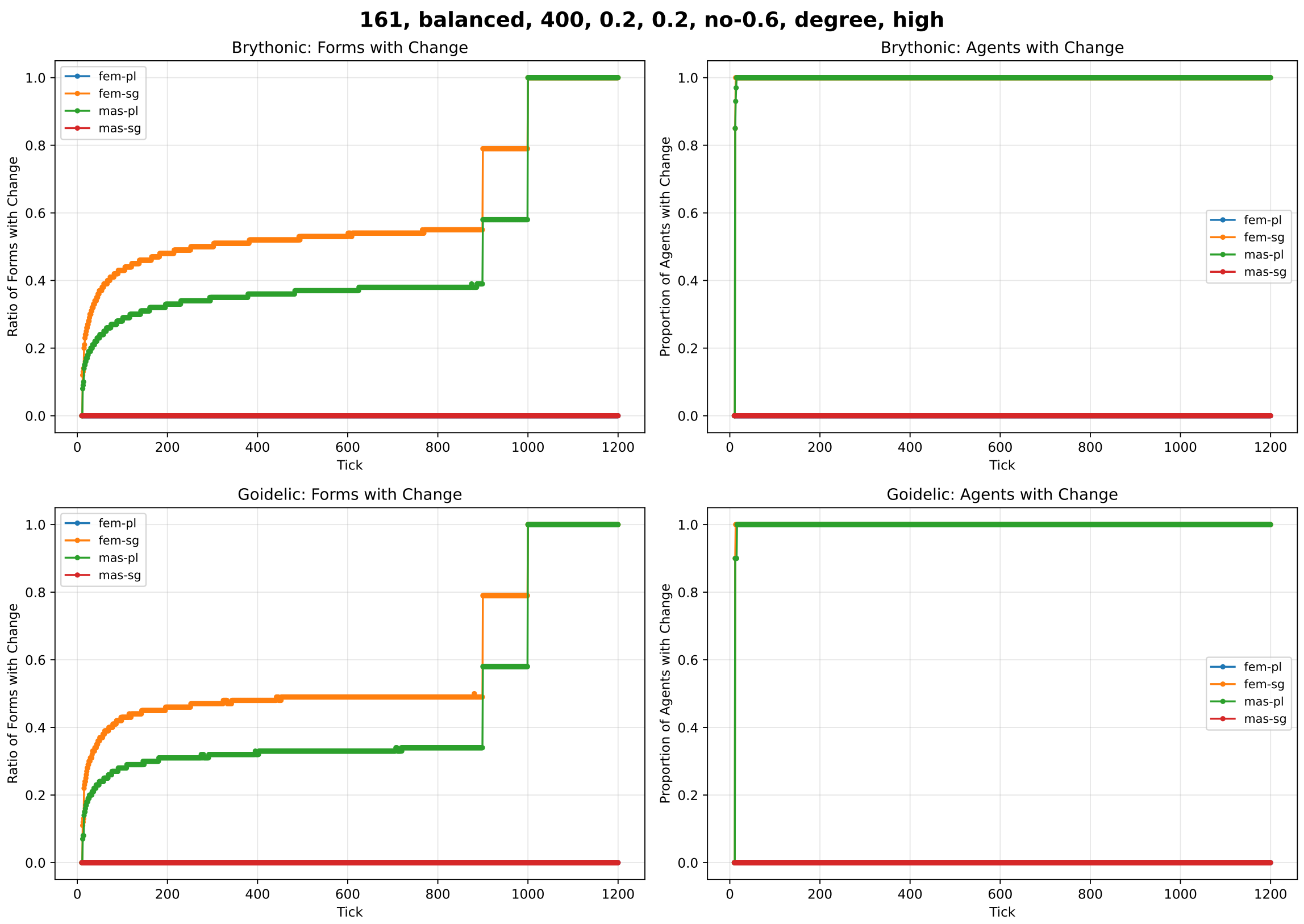

**Figure 29.** Evolution of Lenition from Proto-Celtic to the two Insular Celtic branches for the case where the examples are balanced to roughly represent the proportion of vowel and consonant stems in Proto-Celtic. The number of agents is 50. Probability of adoption is 0.2, novelty is 0.6, and probability of mutation is 0.2. Mutation rules were introduced to the 20% of agents with the highest degrees of connectivity. The seed used is 161. Two plots are shown for each of the Brythonic and Goidelic branches. The lefthand panel shows the proportion of lenited forms for adjectives following nouns (for all forms for all agents) in each of the four morphosyntactic feature combinations. The righthand panel shows the proportion of agents that show at least one instance of mutation for each of the classes.

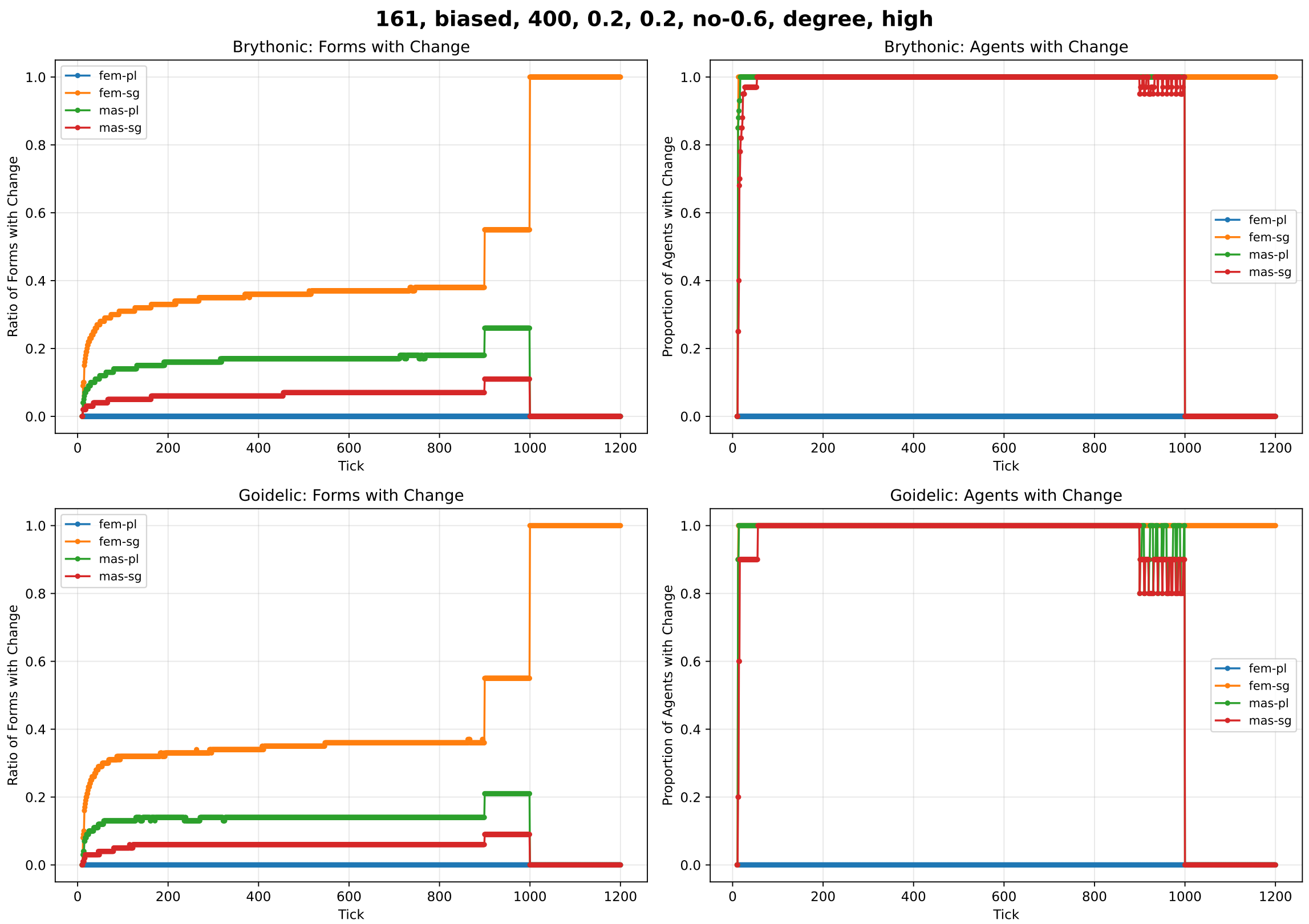

**Figure 30.** Evolution of Lenition from Proto-Celtic to the two Insular Celtic branches for the case where the examples are biased towards having as many consonant stems as vowel stems. The same settings were used as in Fig 29. See the caption for Fig 29 for an explanation of the different panels.

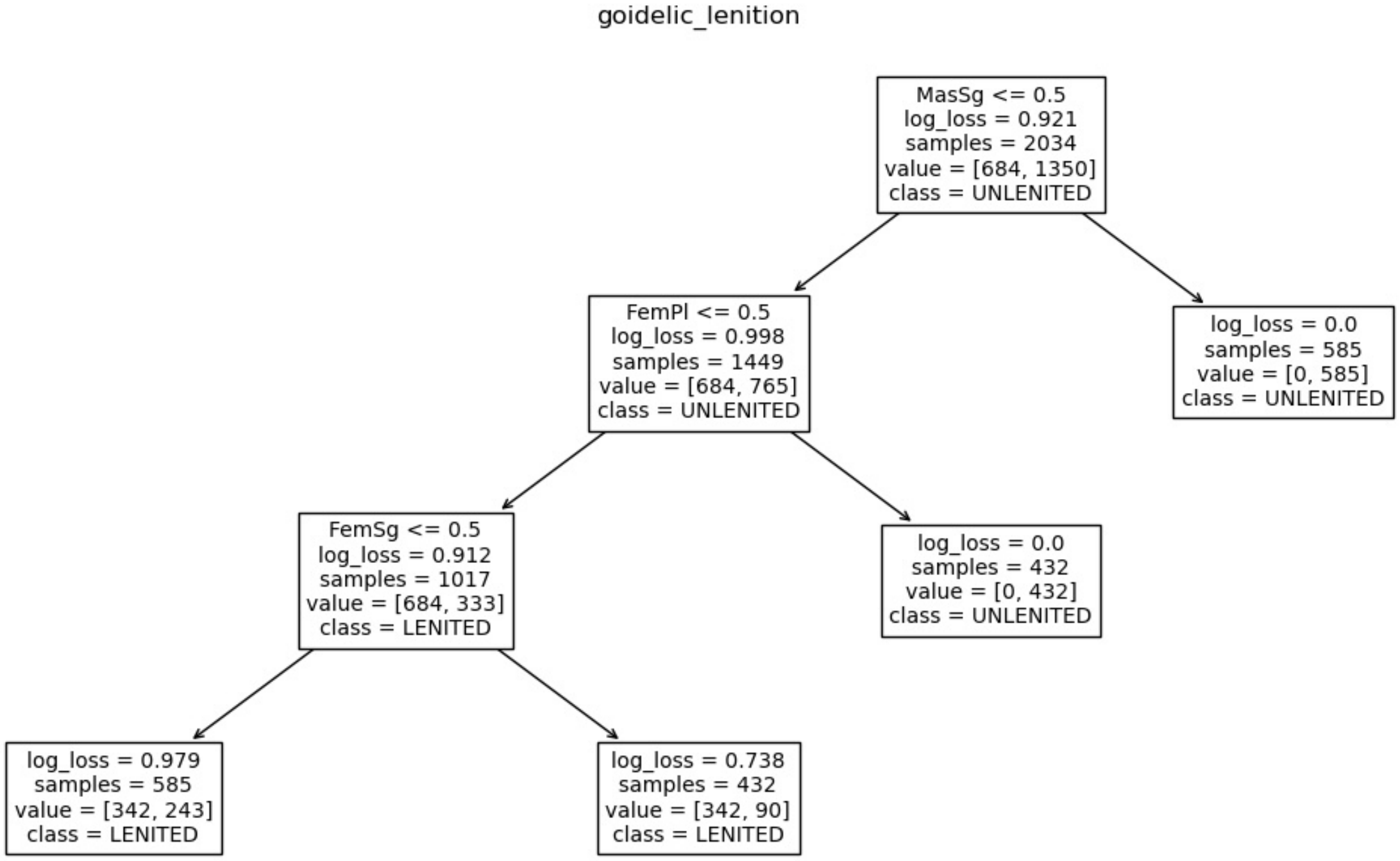

**Figure 31.** Decision tree for one of the Goidelic-speaking agents at tick 1000 for the balanced high-degree case. For masculine-singular and feminine-plural nouns, the tree predicts no lenition, whereas the other two classes induce lenition.

Results for the balanced and biased runs are shown in Figures 29 and 30. For the balanced run after the apocope rule at step 1000, all feminine-singular nouns induce lenition in the following adjectives for both branches. In both branches masculine-plural nouns also induce lenition.

In the case of the biased run, only the feminine singular nouns end up being lenited, and the masculine-plural nouns cease to be triggers for lenition. The reason for this is simply the preponderance of vowel-final forms in the initial data. In the case of feminine-singular, the majority of stems, 55%, are still vowel final. In the case of masculine-plural, only 26% are vowel-final. This contrasts with the situation for the historically more balanced set, where 58% are vowel final. Had the initial distribution of vowel versus consonant stems in Proto-Celtic been different, lenition would not have become morphosyntactically active for some of the classes where it is now found.

Fig 31 shows a decision tree for one of the Goidelic-speaking agents at tick 1000 for the balanced high-degree case. For masculine-singular and feminine-plural nouns, the tree predicts no lenition, whereas the other two classes induce lenition.

Across both runs, the feminine singular (nominative) is a robust environment for lenition of the following adjective, due to the preponderance of *-ā* stems. This corresponds to the situation in real Celtic languages, where in all cases feminine-singular (nominative) nouns induce lenition. The situation for masculine plurals is less stable: in the biased case, lenition completely collapses in that environment. This corresponds to the situation in the real languages where masculine plural (nominatives) as a trigger are less stable across languages: in Irish there is an additional phonological restriction that lenition only occurs after masculine plural nominatives ending in palatalized consonants—basically the residue of the old *-o* stems. In Breton it only applies to masculine plural nouns denoting persons. In Welsh masculine plurals do not induce lenition at all (though forms derived from historical duals do). Only in Cornish do masculine plurals induce lenition in the following adjective as a general rule.